\pdfoutput=1
\documentclass[11pt,letterpaper]{article}

\usepackage[margin=1in]{geometry}
\usepackage[hyphens]{url}
\usepackage{graphicx}
\usepackage{amsmath}
\usepackage{amsfonts}
\usepackage[numbers,sort&compress]{natbib}
\usepackage{caption}
\usepackage{algorithm}
\usepackage{algorithmic}
\usepackage{booktabs}
\usepackage{longtable}
\usepackage{array}
\usepackage{placeins}
\usepackage{float}
\usepackage{microtype}
\usepackage[hidelinks]{hyperref}
\hypersetup{
  pdftitle={MOSAIC-SR: Transformer-Guided Symbolic Regression for Scientific Equation Recovery},
  pdfauthor={Peiyi Zheng, Yanming Kang, Giang Tran}
}

\graphicspath{{AuthorKit27/}}
\def\UrlFont{\rm}
\title{MOSAIC-SR: Transformer-Guided Symbolic Regression for Scientific Equation Recovery}
\author{
Peiyi Zheng$^{*}$ \quad Yanming Kang$^{*}$ \quad Hans De Sterck \quad Giang Tran\\
University of Waterloo\\[0.25em]
\small $^{*}$Co-first authors with equal contribution
}
\date{}

\begin{document}

\maketitle

\begin{abstract}
Symbolic regression aims to recover closed-form equations from observations, providing interpretable models for scientific discovery. Existing approaches struggle to combine flexible structural search with efficient inference. Search-based methods can refine expression structure, but often rely on costly combinatorial optimization with random initialization. Pretrained neural models generate formulas almost instantly, but their predictions often contain symbolic errors. We introduce MOSAIC-SR, which uses a pretrained Transformer to propose multiple initial sketches. These sketches initialize searches in several promising regions, avoiding random starts in the vast expression space. Each search jointly recovers structure and constants through scale-aware constant optimization and local symbolic repair. We evaluate MOSAIC-SR on the SRSD-Feynman dataset with and without dummy variables and on six additional benchmarks. MOSAIC-SR obtains the highest symbolic solution rate on every dataset while ranking among the top two methods in predictive accuracy. This advantage persists in the presence of irrelevant dummy inputs. The results show that learned priors can focus search on promising equation structures, and that numerical optimization and symbolic repair are important for recovery.
\end{abstract}

\section{Introduction}
\label{sec:introduction}

Symbolic regression is a form of model discovery that seeks to infer a closed form expression from observed data.
Given observations $D=\{(\mathbf{x}_i,y_i)\}_{i=1}^{n}\subset\mathbb{R}^{d}\times\mathbb{R}$, the goal is to find an expression \(e\in\mathcal{E}\), with induced function \(f_e\), such that \(y_i\approx f_e(\mathbf{x}_i)\).
Unlike standard regression, the model structure is not fixed in advance. The algorithm must jointly infer a symbolic structure and fit its numerical constants.
A common formulation is
\begin{equation}
e^\star = \displaystyle\arg\min_{e\in\mathcal{E}}
\left[
\sum_{i=1}^{n}\ell\!\left(f_e(\mathbf{x}_i),y_i\right)
+\lambda\,\Omega(e)
\right],
\end{equation}
where \(\Omega(e)\) penalizes expression complexity.
This objective combines discrete structure search with continuous parameter optimization.
The resulting space is large and irregular, and exact symbolic regression is computationally hard in general \citep{virgolin2022symbolic}.
Some systems use heuristic search \citep{cranmer2023pysr}, whereas others exploit domain-specific structure \citep{udrescu2020ai} or learned equation priors \citep{biggio2021neural}.

Search-based systems treat each dataset as a new combinatorial problem.
Classical methods such as Eureqa and age fitness Pareto optimization evolve expression trees from data \citep{schmidt2009distilling,schmidt2010age}.
Modern variants improve different parts of this pipeline. PySR uses a multi-population evolutionary algorithm with an evolve-simplify-optimize loop \citep{cranmer2023pysr}. Operon emphasizes efficient tree representation and high performance implementation \citep{burlacu2020operon}. GP-GOMEA learns dependencies between tree positions to guide variation \citep{virgolin2021improving}. AI Feynman reduces search through recursive decomposition based on mathematical and physical structure \citep{udrescu2020ai}.
These methods are flexible because they can modify both structure and constants after seeing data.
However, they do not amortize structure discovery across problems as useful subexpressions must be rediscovered for every new input table.

Therefore, the main bottleneck is not only fitting coefficients but also locating a small region of the expression space that contains a recoverable equation.
This cost becomes especially visible in scientific formulas, where constants may encode physical scales and small structural errors can be hidden by local numerical fit.

Recent neural methods change this cost profile by learning priors over equations from many training tasks.
NeSymReS and end-to-end Transformer symbolic regression train sequence models on synthetic functions and query them on new input-output observations at inference time \citep{biggio2021neural,kamienny2022end}.
Other work studies links between symbolic formulas and numeric observations through unified or contrastive pretraining \citep{meidani2024snip}.
These approaches can propose candidate expressions much faster than a search method initialized from random trees.
However, fast proposal is not the same as equation recovery.
Many models are trained or selected by pointwise prediction error on sampled domains, but an expression with high test \(R^2\) may still use the wrong variables, constants, or algebraic form.
For example, a polynomial approximation may achieve near perfect \(R^2\) over a narrow domain but fail to recover the underlying exponential law.
This distinction is central for scientific discovery, where the output is not merely a predictor but an equation.

This paper focuses on the gap between numerical fit and symbolic recovery.
For a target expression \(f^\star\) and a prediction \(\hat{f}\), high predictive accuracy does not imply algebraic equivalence. We therefore report both predictive accuracy, defined by test \(R^2>1-\varepsilon\) for some small $\varepsilon$, and symbolic solution rate under a fixed symbolic verification protocol.
We use the same symbolic verification protocol as SRBench to determine whether a predicted expression is equivalent to the known ground-truth equation \citep{lacava2021contemporary}.
Accordingly, our experiments emphasize methods that can move from a good numerical fit to an algebraically correct formula.

We introduce MOSAIC-SR, a hybrid framework that uses neural generation to guide symbolic search. A pretrained Transformer maps a small subsample of observations to multiple symbolic sketches, concentrating search in promising regions of expression space. Monte Carlo tree search (MCTS) maps the decoded variable tokens to observed inputs and produces several complete equations. These equations initialize independent local search paths. Within each path, structural edits alternate with constant refitting to correct errors in variables, operators, subexpressions, and numerical scales after decoding. Together, the pretrained Transformer provides fast neural proposals, while explicit search retains the flexibility needed for exact equation recovery. We use log-spaced initializations for constant fitting to cover scientific constants that may span many orders of magnitude.

\paragraph{Contributions.}
Our main contributions are summarized as follows. First, we introduce a recovery-oriented pretraining scheme that learns normalized structural skeletons while deliberately deferring dataset-specific variable assignment and constant estimation to inference, where the observed data support more accurate choices. We sample equations under varied input distributions, domains, and point clouds. We also omit affine transformations from the pretrained skeletons. Although such transformations can improve numerical fit, their additional scales and offsets may introduce terms absent from the ground-truth expression and thereby hinder exact symbolic matching. Our pretraining objective instead encourages sketches that preserve the target algebraic structure.

Second, we introduce a Transformer-guided inference framework that assigns decoded variable tokens to observed inputs and transforms structural sketches into complete equations through parallel local search, tree-based repair, and scale-aware constant fitting. Performing variable assignment and constant estimation at inference allows these choices to be guided directly by the observed data. Structural repair and constant fitting recover high predictive accuracy, while the cleaner initial structures remain compatible with exact symbolic equivalence. The resulting framework therefore targets both strong \(R^2\) and accurate symbolic recovery rather than predictive fit alone.
Like constructing a mosaic from selected pieces, MOSAIC-SR identifies and assembles symbolic components that explain the observed numerical data.

Third, we compare MOSAIC-SR with search-based symbolic regression methods, direct neural generation methods, and neural-guided search methods. Across SRSD-Feynman with and without dummy variables and six additional benchmarks, MOSAIC-SR achieves the highest symbolic solution rate on every dataset while ranking among the top two methods in predictive accuracy.
% and retains comparable symbolic recovery performance when irrelevant dummy variables are introduced.

\section{Related Work}
\label{sec:related_work}

\subsection{Symbolic Regression as Equation Recovery}

Symbolic regression has been used as a tool for scientific discovery because its output is an explicit equation rather than a black box predictor.
Early work showed that evolutionary search can recover compact analytic relations from data and can produce models that are interpretable to scientists \citep{schmidt2009distilling}.
Unlike ordinary regression, symbolic regression aims not only to predict unseen observations well but also to recover the underlying generating mechanism. However, predictive accuracy alone does not guarantee recovery of the target equation. Indeed, many algebraically different expressions can agree on a finite set of sampled points.

Recent SR benchmarks therefore evaluate symbolic recovery separately from predictive accuracy. SRBench evaluates contemporary SR methods under a shared protocol and reports symbolic solution rate in addition to predictive accuracy \citep{lacava2021contemporary}. SRSD-Feynman extends this view for scientific discovery.
The original Feynman benchmark samples variables from predefined narrow ranges \citep{udrescu2020ai}. SRSD-Feynman uses equation specific value ranges based on the physical meaning and SI unit of each variable. It also samples variables on a logarithmic scale to cover various orders of magnitude \citep{matsubara2022rethinking}. 

This sampling strategy aims to mitigate a known limitation. An incorrect expression may fit finite observations well over a narrow domain. Evaluation across physically meaningful scales can help distinguish such local approximations from the true underlying equation. We therefore report $R^2$ based accuracy and symbolic solution rate separately. Predictive accuracy measures numerical fit and the symbolic solution verifier tests whether the output expression is symbolically equivalent to the ground truth.

\subsection{Search-Based Symbolic Regression}

Search-based symbolic regression represents candidate equations as expression trees and optimizes their structure directly. Genetic programming maintains a population of trees and generates new candidates through selection, crossover, mutation and constant fitting \citep{schmidt2009distilling,schmidt2010age}. Without imposing a fixed parametric form, these methods can revise operators and subtrees in response to the observed data. 

Modern systems improve different components of this search. Operon uses a compact tree representation to evaluate expressions efficiently \citep{burlacu2020operon}. GP-GOMEA learns dependencies between tree positions to guide structural variation \citep{virgolin2021improving}. PySR uses a multi-population evolutionary algorithm with an evolve-simplify-optimize loop \citep{cranmer2023pysr}. AI Feynman takes a more scientific route. It recursively reduces a problem by exploiting structure such as separability, symmetry, and dimensional analysis \citep{udrescu2020ai}. The main strength of search-based SR is adaptive refinement after observing the data. Constants can be optimized again after each structural revision. Explicit mathematical and physical constraints can further guide the search. The limitation is that most search-based methods solve each new problem from scratch.
Locating the true underlying equation in the vast search space from random initialization is often difficult. Repeated candidate generation and evaluation therefore incur substantial computational cost.

\subsection{Neural Symbolic Regression and Guided Search}

Neural symbolic regression methods differ in how the learned model interacts with the discrete equation space. Direct generation methods use a neural model to define a distribution over candidate expressions. Candidate selection may occur through sequence decoding or optimization in a learned latent space. These procedures operate within a learned output space rather than explicitly editing symbolic trees. Neural-guided methods use learned models to guide a separate search process that explores and revises symbolic structures. The distinction is whether the neural model is treated as the primary solver or as a prior for structural search.

\paragraph{Direct Neural Generation.}
Early neural generators formulate expression construction as a sequential decision process. DSR trains an RNN policy with a risk-seeking policy gradient that concentrates probability on high-reward expressions \citep{petersen2021deep}. Pretrained Transformer models transfer expression patterns learned from large synthetic data to new problems. NeSymReS decodes symbolic skeletons with beam search and fits placeholder constants with BFGS \citep{biggio2021neural}. End-to-end Transformer SR predicts complete expressions with numerical constants. An optional optimization step refines these constants after decoding \citep{kamienny2022end}. SNIP-SR learns a shared symbolic-numeric representation and searches its continuous latent space. SNIP-SR refines decoded constants with BFGS \citep{meidani2024snip}. GenSR constructs a continuous equation-generative space with a conditional variational autoencoder. CMA-ES then searches this space for candidate equations \citep{li2026gensr}. In SNIP-SR and GenSR, optimization occurs in a learned continuous space rather than through direct edits to expression trees.

Pretrained generators can reduce per-problem inference cost, but direct generation remains structurally fragile. An early token error can alter the remaining expression. Finite observations over a compact domain also do not uniquely determine the underlying symbolic structure. Constant fitting and other numerical optimization can make an expression with incorrect structure closely approximate the target on that domain. The resulting model may achieve a high \(R^2\) score without symbolic recovery.

\paragraph{Neural-Guided Search.}
Neural-guided search combines learned proposals with explicit structural exploration. NGGP uses a recurrent policy to seed genetic programming populations \citep{mundhenk2021symbolic}. uDSR combines recurrent generation with pretraining and evolutionary search \citep{landajuela2022unified}. TPSR integrates MCTS into Transformer decoding so that prediction error and expression complexity can guide token selection \citep{shojaee2023transformer}. RSRM uses Double Q-learning to restrict the actions explored by MCTS during expression tree construction \citep{xu2024rsrm}. SR4MDL uses a description length estimate from MDLformer as the objective for symbolic search \citep{yu2025sr4mdl}. In these methods, explicit structural search remains responsible for constructing the final equation. The learned model can narrow the candidate space or define the search objective. Unlike methods that use the learned model to generate the final expression or guide individual search decisions, MOSAIC-SR uses the Transformer to propose diverse initial sketches. Local search then completes and revises these sketches while refitting constants. Our objective is not predictive fit alone, but recovery of an expression that is symbolically equivalent to the target equation.

\clearpage
\begin{figure}[!htbp]
\centering
\includegraphics[page=1,trim=230 60 230 200,clip,width=\textwidth]{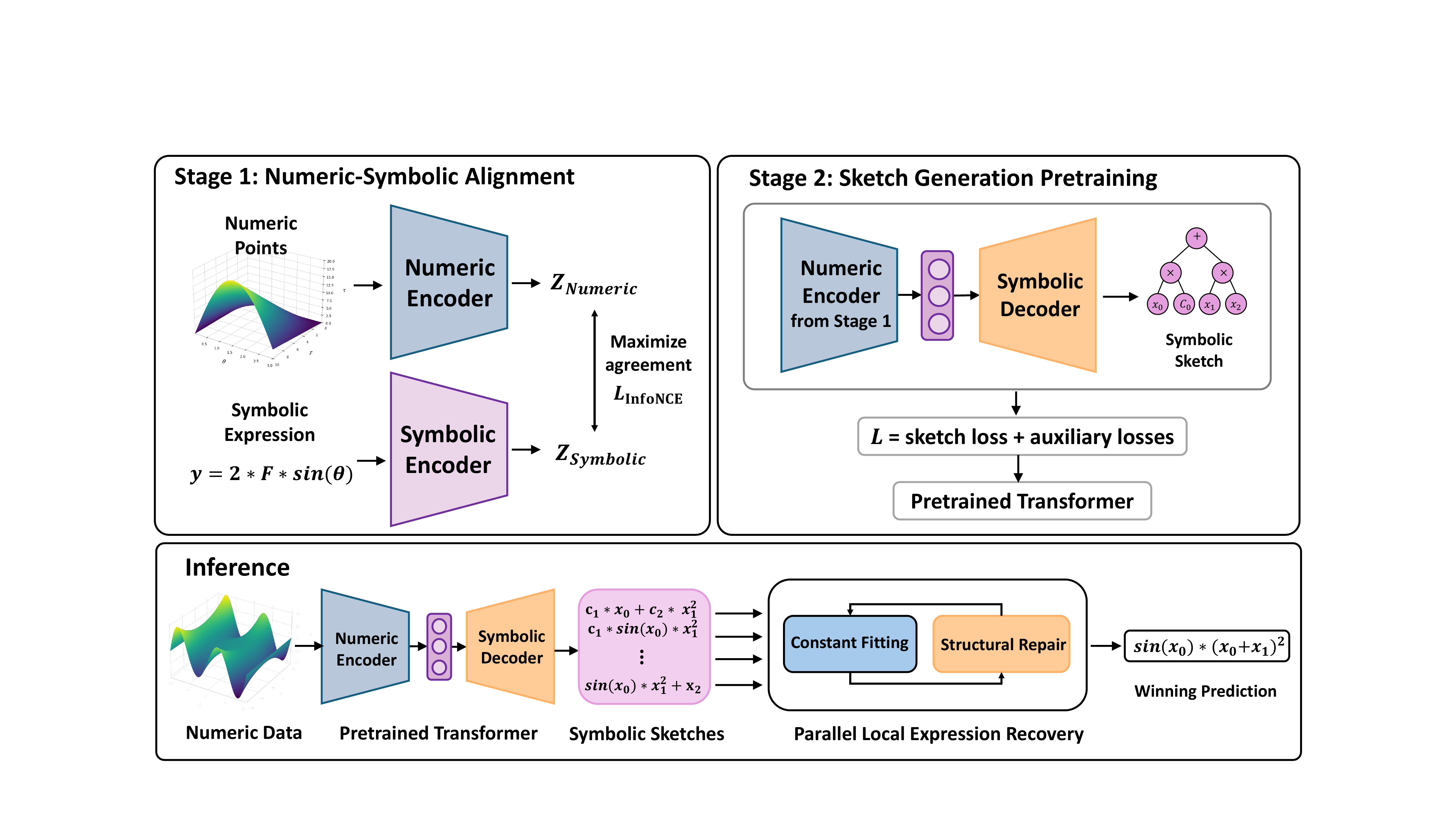}
\caption{Overview of MOSAIC-SR. Stage 1 aligns numerical observations with symbolic expressions through contrastive pretraining, and Stage 2 trains the decoder to predict operator structures with variable and constant placeholders. At inference time, MCTS maps these variable tokens to observed inputs, and the resulting equations initialize parallel search trajectories that alternate constant fitting with structural repair. The final expression is selected according to validation NMSE and complexity.}
\label{fig:mosaic_overview}
\end{figure}

\section{Method}
\label{sec:method}
MOSAIC-SR is motivated by the observation that a pretrained model can often identify the broad form of an equation, while a nearly correct prediction may still contain the wrong variable, operator, or constant. To address this limitation, we use the pretrained Transformer to propose equation sketches and then apply search-based refinement to correct structural errors.

Figure~\ref{fig:mosaic_overview} summarizes the three phases of MOSAIC-SR. First, the numeric and symbolic encoders are jointly pretrained with a contrastive objective that aligns numerical observations with their corresponding expressions. The pretrained numeric encoder is then connected to an autoregressive decoder and further trained to generate symbolic sketches. Second, MCTS uses the decoded sketches and learned structural priors to assign variables and fit constants. This produces a diverse set of promising initial equations. Third, parallel local searches refine these equations through structural edits and constant optimization. The final expression is selected according to validation NMSE and complexity.

\subsection{Pretraining: Learning the Symbolic Sketch Prior}
\label{subsec:sketch_prior}

\paragraph{Synthetic Numeric-Symbolic pairs.}
The sketch prior is trained entirely with synthetic equations, following the pretraining strategy used by E2E-SR \citep{kamienny2022end}. We sample expressions with one to ten variables from a predefined symbolic vocabulary comprising binary operators \(\{+,-,\times,\div\}\) and unary operators $\{(\cdot)^2,(\cdot)^3,\sin,\cos,\exp,\log,\sqrt{\cdot},\tan,\tanh,\arctan\}$.
Each expression is represented in prefix Polish notation, where every operator appears before its operands. For example, \(x_1+\sin(x_2)\) is written as \texttt{(ADD X\_1 SIN X\_2)}. 
% The symbolic encoder and sketch decoder share a deterministic vocabulary of 45 tokens including special tokens, operators, fixed constants, free-constant placeholder, and variable tokens.
In contrast to the data pipelines described for E2E-SR, we use SymPy \citep{meurer2017sympy} to normalize each sampled expression under a fixed set of simplification rules. For example,
\begin{equation}
3+x+2x+\sin(y\cdot x)-1
\;\mapsto\;
2+3x+\sin(xy).
\end{equation}
This step reduces the number of algebraically equivalent forms treated as distinct symbolic targets. We use the normalized expression to create two symbolic sequences: the full sequence and the sketch sequence. The full sequence contains the exact operators, variable identities, and constants and is used by the symbolic encoder during contrastive pretraining. The sketch sequence preserves the operator tree and leaf positions while replacing each variable and constant leaf with a corresponding special token. This sequence serves as the decoder target. We then randomly choose an input distribution and domain from predefined sets, sample 200 input points, and evaluate the expression to generate the corresponding output values. Consequently, the same symbolic expression can be observed through different input ranges rather than a single fixed set of observations.

\paragraph{Numeric encoder.}
Let $D=\{(\mathbf{x}_i,y_i)\}_{i=1}^{n}$ denote the numeric observations, with \(n=200\) during pretraining. We pad all inputs to the maximum variable dimension to maintain a consistent input width across tasks. We then concatenate each padded input with its output and apply the following signed logarithmic transform elementwise:
\begin{equation}
\phi(a)=\mathrm{sign}(a)\log(1+|a|).
\end{equation}
Since the order of observations is arbitrary, the transformed rows are encoded as a set rather than a sequence. A Set Transformer \citep{lee2019set} contextualizes each row using information from the other observations, and attention pooling summarizes the resulting features:
\begin{equation}
\mathbf{M}_D=\mathrm{SetEnc}\!\left(\{\phi(\mathbf{r}_i)\}_{i=1}^{n}\right),
\qquad
\mathbf{z}_D=\mathrm{Pool}(\mathbf{M}_D).
\end{equation}
The set representation \(\mathbf{M}_D\) is retained for decoder cross-attention, while the pooled representation \(\mathbf{z}_D\) is used only for numeric-symbolic contrastive learning \citep{meidani2024snip}.

\paragraph{Symbolic encoder.}
The symbolic encoder processes the normalized prefix sequence of an expression. In prefix notation, each operator appears before its operands, and its fixed arity determines how many subsequent subexpressions belong to it. The expression tree can therefore be parsed unambiguously without parentheses. Each token in the prefix sequence is mapped to an embedding. The resulting sequence is passed through a Transformer encoder with learned positional embeddings \citep{vaswani2017attention}. Masked mean pooling over the output tokens produces a symbolic embedding $\mathbf{z}_e$ for the complete expression. This encoder is used during the contrastive learning stage to teach the numeric encoder which equation generated the observations.

\paragraph{Stage 1: Numeric-Symbolic alignment.}
We train the numeric and symbolic encoders with the bidirectional InfoNCE objective used by CLIP \citep{oord2018representation,radford2021learning}. Following SNIP, we apply this objective to align each set of numerical observations with its corresponding symbolic expression \citep{meidani2024snip}. For a minibatch of $B$ matched observation sets and expressions, define the temperature scaled similarity
\begin{equation}
s_{ij}=\frac{\cos(\mathbf{z}_{D_i},\mathbf{z}_{e_j})}{\tau}.
\end{equation}
Here \(\mathbf{z}_{D_i}\) is the pooled numeric embedding of the \(i\)-th observation set, \(\mathbf{z}_{e_j}\) is the symbolic embedding of the \(j\)-th expression, and \(\tau>0\) is the temperature parameter that controls the sharpness of the similarity distribution. A smaller \(\tau\) places greater emphasis on the most similar pairs.
The bidirectional InfoNCE loss is
\begin{equation}
\mathcal{L}_{\mathrm{con}}=-\frac{1}{2B}\sum_{i=1}^{B}\left[
\log\frac{e^{s_{ii}}}{\sum_{j=1}^{B}e^{s_{ij}}}
+\log\frac{e^{s_{ii}}}{\sum_{j=1}^{B}e^{s_{ji}}}
\right].
\end{equation}
This loss increases the similarity of each matched pair relative to all mismatched pairs in both directions, aligning the observations with their corresponding symbolic expressions in a shared embedding space.

\paragraph{Stage 2: sketch generation.}
After contrastive pretraining, an autoregressive Transformer decoder is attached to the numeric encoder. Training uses teacher forcing against the ground-truth sketch sequence of operators, variable tokens, and constant placeholders. Let \(u_{i,t}\) denote the target token at position \(t\) in the \(i\)-th sketch, let \(T_i\) be the number of non-padding target tokens, and let \(N_{\mathrm{tok}}=\sum_{i=1}^{B}T_i\). The token-level cross-entropy is
\begin{equation*}
\mathcal{L}_{\mathrm{CE}}
=-\frac{1}{N_{\mathrm{tok}}}
\sum_{i=1}^{B}\sum_{t=1}^{T_i}
\log p_{\theta}\!\left(u_{i,t}\mid u_{i,<t},\mathbf{M}_{D_i}\right),
\end{equation*}
where \(p_{\theta}\) is the decoder distribution conditioned on the preceding ground-truth tokens and numeric representation \(\mathbf{M}_{D_i}\). Several lightweight prediction heads are trained jointly to estimate coarse structural properties such as the root arity and root operator class. The second-stage sketch-generation objective is summarized as
\begin{equation}
\mathcal{L}_{\mathrm{sketch}}
=\mathcal{L}_{\mathrm{CE}}
+\lambda_{\mathrm{aux}}\mathcal{L}_{\mathrm{aux}},
\end{equation}
where \(\mathcal{L}_{\mathrm{aux}}\) collects the structural prediction losses and \(\lambda_{\mathrm{aux}}\) controls their contribution. These auxiliary objectives encode structural information in the numeric representation, which then guides the construction of initial equations.

\subsection{Inference}
\label{subsec:inference}

\subsubsection{From Sketches to Initial Equations}

The numeric encoder maps the numerical observations to latent representations. Stochastic decoding then samples diverse high probability sketches from these representations \citep{holtzman2020curious}. We discard duplicate sketches and those that violate the prefix grammar. Each retained sketch defines an expression skeleton with unresolved variable and constant tokens. MCTS assigns observed inputs to the variable tokens, and numerical optimization fits the constant values on the training set \citep{shojaee2023transformer}. We rank the resulting complete expressions by training NMSE and use the top 24 to initialize 24 independent local search trajectories.

\subsubsection{Parallel Local Equation Recovery}
\label{subsec:local_repair}
Each search path maintains one current expression and attempts to improve it through a sequence of small tree edits. Different decoded sketches and MCTS assignments seed the search paths in distinct regions.

\paragraph{Scale-aware constant fitting.}
Every candidate tree may contain several free constants.
For a fixed structure \(e(\mathbf{x};\mathbf{c})\), constants are estimated by minimizing squared error on the training set,
\begin{equation}
\hat{\mathbf{c}}
=
\arg\min_{\mathbf{c}}
\sum_{(\mathbf{x}_i,y_i)\in D_{\mathrm{train}}}
\left(e(\mathbf{x}_i;\mathbf{c})-y_i\right)^2.
\end{equation}
The objective is solved with Levenberg-Marquardt optimization from logarithmically spaced initializations that span multiple orders of magnitude. During local search, constants are refitted periodically after several structural edits.

\paragraph{Structural repair.}
The search paths run in parallel, each maintaining an independent current expression. A path iteratively modifies its expression by replacing an operator, variable, or small subtree. After several mutations, the constants are refitted and the candidate is evaluated on \(D_{\mathrm{train}}\) using NMSE normalized by the target variance. The candidate replaces the current expression only if it achieves a lower NMSE. Each path performs 300 mutation attempts.

\subsubsection{Selection and Simplification}
\label{subsec:search_portfolio}

Throughout local search, a shared Hall of Fame stores the best candidates found across all search paths. Once all paths have completed 300 mutation attempts, MOSAIC-SR selects the final expression from this candidate pool using \(D_{\mathrm{val}}\).
Candidates with validation NMSE below \(10^{-15}\) are classified as near exact fits. When one or more candidates satisfy this threshold, the selector chooses the expression with the fewest operators. Otherwise, it selects the candidate with the lowest validation NMSE. This criterion favors structurally simpler expressions once the validation error is near zero.

The selected expression is then simplified by replacing small constants with zero, and fitted values close to common physical constants are snapped to their symbolic forms. A physical constant snapping is retained only when its validation error remains within a small tolerance of the original
expression.

\section{Experiments}
\label{sec:experiments}

We conduct two evaluations. The first focuses on scientific equation recovery using SRSD-Feynman, including its dummy variable variant. The second tests whether the same method transfers to six established symbolic regression benchmarks. We also conduct ablation studies on the SRSD-Feynman Easy split to examine conditional MCTS, local mutation search, and search breadth and depth. The ablation results, detailed training parameters, and experimental settings are provided in the supplementary material.

\subsection{Evaluation Protocol}
\label{subsec:metrics}
We evaluate numerical accuracy and symbolic solution rate following the SRBench protocol. For every run conducted in this work, the test set is used only for final scoring and the returned expression is independently parsed and reevaluated outside the method.
Invalid expressions, non-finite predictions and timeouts remain failures in the full benchmark denominator. Each evaluation reports a numerical recovery rate and a symbolic solution rate. For a set of problems \(\mathcal{P}\), symbolic solution rate is
\begin{equation}
\mathrm{Sym}=
\frac{1}{|\mathcal{P}|}
\sum_{p\in\mathcal{P}}
\mathbf{1}\left[
\mathrm{SymVerify}(\hat{e}_p,e^\star_p)=1
\right],
\end{equation}
where \(\hat{e}_p\) is the returned expression, \(e^\star_p\) is the ground truth expression, and \(\mathrm{SymVerify}\) is the fixed symbolic verifier.
We use the SRBench symbolic solution verifier \citep{lacava2021contemporary}.
The numerical criterion is specified below in each experiment. 

\subsection{SRSD-Feynman Evaluation}
\label{subsec:main_results}

\paragraph{Benchmark.}
SRSD-Feynman contains 120 physics inspired equations with known ground truth and physically motivated sampling ranges \citep{matsubara2022rethinking}.
The benchmark is divided into Easy, Medium, and Hard splits with 30, 40, and 50 problems, respectively.
Each problem provides 8000 training, 1000 validation, and 1000 test samples.
We also evaluate the dummy variable variant, which adds between one and three irrelevant input columns at random positions.
Results for both the original and dummy variable variants are reported below.
We additionally include EPLEX \citep{lacava2016epsilon}, DGSR \citep{holt2023deep}, TaylorGP \citep{he2022taylor}, DySymNet \citep{li2024dysymnet}, FFX \citep{mcconaghy2011ffx}, and BSR \citep{jin2020bayesian} in the SRSD-Feynman comparison.

For this evaluation, predictive quality is measured by
\begin{equation}
R^2_p=1-
\frac{\sum_i\left(y_i-f_{\hat{e}_p}(\mathbf{x}_i)\right)^2}
{\sum_i\left(y_i-\bar{y}\right)^2},
\end{equation}
and \(\mathrm{Acc}\) is the percentage of test problems with \(R^2>0.999\).

\begin{table}[t]
\centering
\scriptsize
\setlength{\tabcolsep}{1pt}
\caption{Results on SRSD-Feynman. Easy, Medium, and Hard contain 30, 40, and 50 equations. Results cited from SRSD-Feynman are taken directly from the original paper. Results evaluated in this work are means over five seeds. Sym is the symbolic solution rate in percent, and Acc is the percentage of equations with \(R^2>0.999\). The top three values per column, including ties, are bolded. The best value in each column is also italicized.}
\label{tab:main_results}
\begin{tabular*}{\columnwidth}{@{\extracolsep{\fill}}llcccccc@{}}
\toprule
& & \multicolumn{2}{c}{Easy (30)} & \multicolumn{2}{c}{Medium (40)} & \multicolumn{2}{c}{Hard (50)} \\
\cmidrule(lr){3-4}
\cmidrule(lr){5-6}
\cmidrule(lr){7-8}
Method & Type & Sym & Acc & Sym & Acc & Sym & Acc \\
\midrule
\multicolumn{8}{l}{\textit{Cited from SRSD-Feynman}} \\
uDSR & RL & 50.0 & \textbf{\textit{100.0}} & 17.5 & 75.0 & 4.0 & 20.0 \\
PySR & GP & \textbf{60.0} & 66.7 & 30.0 & 45.0 & 4.0 & 38.0 \\
DSR & RL & 46.7 & 63.3 & 10.0 & 45.0 & 2.0 & 28.0 \\
AI Feynman & Rules & 30.0 & 33.3 & 2.5 & 5.0 & 2.0 & 6.0 \\
E2E & Neural & 0.0 & 26.7 & 0.0 & 17.5 & 0.0 & 14.0 \\
AFP-FE & GP & 23.3 & 26.7 & 2.5 & 2.5 & 0.0 & 4.0 \\
AFP & GP & 20.0 & 20.0 & 2.5 & 2.5 & 0.0 & 4.0 \\
gplearn & GP & 6.7 & 6.7 & 0.0 & 7.5 & 0.0 & 2.0 \\
\midrule
\multicolumn{8}{l}{\textit{Evaluated in this work}} \\
SBP-GP & GP & 8.0 & \textbf{87.3} & 4.0 & \textbf{82.0} & 0.0 & \textbf{\textit{56.0}} \\
SR4MDL & Hybrid & \textbf{63.3} & 86.7 & \textbf{43.0} & 66.0 & \textbf{21.2} & 48.4 \\
TPSR & Hybrid & 47.3 & 86.0 & \textbf{44.0} & \textbf{81.0} & \textbf{20.0} & \textbf{52.0} \\
NGGP & Hybrid & 36.7 & 65.3 & 9.0 & 52.0 & 0.4 & 34.8 \\
RSRM & Hybrid & 0.0 & 64.0 & 0.0 & 47.0 & 0.0 & 22.8 \\
GP-GOMEA & GP & 24.7 & 63.3 & 19.5 & 65.0 & 10.0 & 44.0 \\
EPLEX & GP & 6.7 & 62.0 & 10.0 & 55.0 & 4.4 & 28.0 \\
DGSR & Neural & 36.7 & 58.7 & 11.5 & 49.0 & 0.0 & 22.4 \\
TaylorGP & GP & 23.3 & 35.3 & 18.5 & 24.0 & 6.4 & 10.4 \\
DySymNet & Neural & 7.3 & 29.3 & 0.0 & 17.5 & 0.0 & 4.8 \\
FFX & Linear & 14.0 & 26.7 & 17.5 & 31.0 & 11.6 & 16.0 \\
Operon & GP & 2.0 & 23.3 & 7.0 & 32.0 & 6.0 & 14.0 \\
NeSymReS & Neural & 16.7 & 21.3 & 2.5 & 14.0 & 3.6 & 4.8 \\
BSR & Bayes & 0.0 & 0.7 & 0.0 & 0.0 & 0.0 & 0.0 \\
\midrule
MOSAIC-SR & Hybrid & \textbf{\textit{84.7}} & \textbf{96.7} & \textbf{\textit{55.0}} & \textbf{\textit{90.0}} & \textbf{\textit{34.8}} & \textbf{53.2} \\
\bottomrule
\end{tabular*}
\end{table}

\begin{table}[t]
\centering
\scriptsize
\setlength{\tabcolsep}{1pt}
\caption{Results on SRSD-Feynman with dummy variables. Results cited from SRSD-Feynman are taken directly from the original paper. Results evaluated in this work are means over five seeds. Sym is the symbolic solution rate in percent, and Acc is the percentage of problems with test \(R^2>0.999\). The top three values in each column, including ties, are bolded. The best value in each column is also italicized.}
\label{tab:dummy_results}
\begin{tabular*}{\columnwidth}{@{\extracolsep{\fill}}llcccccc@{}}
\toprule
& & \multicolumn{2}{c}{Easy (30)} & \multicolumn{2}{c}{Medium (40)} & \multicolumn{2}{c}{Hard (50)} \\
\cmidrule(lr){3-4}
\cmidrule(lr){5-6}
\cmidrule(lr){7-8}
Method & Type & Sym & Acc & Sym & Acc & Sym & Acc \\
\midrule
\multicolumn{8}{l}{\textit{Cited from SRSD-Feynman}} \\
uDSR & RL & 10.0 & 53.3 & 7.5 & 37.5 & 0.0 & 12.0 \\
PySR & GP & 20.0 & 20.0 & 5.0 & 10.0 & 0.0 & 2.0 \\
DSR & RL & 10.0 & 76.7 & 0.0 & 45.0 & 2.0 & 22.0 \\
AI Feynman & Rules & 0.0 & 6.7 & 0.0 & 0.0 & 0.0 & 0.0 \\
E2E & Neural & 0.0 & 16.7 & 0.0 & 12.5 & 0.0 & 10.0 \\
AFP-FE & GP & 16.7 & 16.7 & 0.0 & 0.0 & 0.0 & 4.0 \\
AFP & GP & 16.7 & 20.0 & 0.0 & 5.0 & 0.0 & 4.0 \\
gplearn & GP & 0.0 & 0.0 & 0.0 & 0.0 & 0.0 & 0.0 \\
\midrule
\multicolumn{8}{l}{\textit{Evaluated in this work}} \\
SBP-GP & GP & 0.0 & \textbf{86.7} & 0.0 & \textbf{\textit{82.5}} & 0.0 & \textbf{\textit{53.2}} \\
SR4MDL & Hybrid & \textbf{40.0} & 80.0 & \textbf{14.0} & 28.5 & 4.8 & 24.8 \\
TPSR & Hybrid & \textbf{32.0} & \textbf{84.0} & \textbf{36.0} & \textbf{82.0} & \textbf{17.2} & \textbf{48.0} \\
NGGP & Hybrid & 8.7 & 60.0 & 6.0 & 50.5 & 0.0 & 33.2 \\
RSRM & Hybrid & 0.0 & 60.7 & 0.0 & 45.0 & 0.0 & 22.0 \\
GP-GOMEA & GP & 4.7 & 60.0 & 9.5 & 62.5 & \textbf{8.8} & 42.0 \\
EPLEX & GP & 2.7 & 61.3 & 7.0 & 54.5 & 2.8 & 24.4 \\
DGSR & Neural & 3.3 & 30.7 & 1.5 & 31.5 & 0.0 & 14.0 \\
TaylorGP & GP & 6.7 & 26.0 & 10.5 & 33.0 & 6.8 & 14.0 \\
DySymNet & Neural & 4.7 & 5.3 & 0.0 & 4.5 & 0.0 & 2.0 \\
FFX & Linear & 11.3 & 26.7 & 12.5 & 31.5 & 8.8 & 16.0 \\
Operon & GP & 2.7 & 23.3 & 6.0 & 32.0 & 6.4 & 14.0 \\
NeSymReS & Neural & 4.0 & 8.0 & 0.0 & 6.0 & 0.8 & 2.0 \\
BSR & Bayes & 0.0 & 0.0 & 0.0 & 0.0 & 0.0 & 0.0 \\
\midrule
MOSAIC-SR & Hybrid & \textbf{\textit{74.0}} & \textbf{\textit{94.0}} & \textbf{\textit{54.5}} & \textbf{\textit{82.5}} & \textbf{\textit{31.2}} & \textbf{52.4} \\
\bottomrule
\end{tabular*}
\end{table}

\begin{table*}[t]
\centering
\scriptsize
\setlength{\tabcolsep}{1pt}
\caption{Results and runtime on six symbolic regression benchmarks. Sym and Acc are percentages. Time is the average runtime in seconds per equation. Acc denotes test NMSE below \(10^{-6}\). For each dataset, the top three Sym and Acc values are bolded.}
\label{tab:cross_benchmark}
\begin{tabular*}{\textwidth}{@{\extracolsep{\fill}}l*{6}{rrr}@{}}
\toprule
Method
& \multicolumn{3}{c}{Nguyen}
& \multicolumn{3}{c}{Korns}
& \multicolumn{3}{c}{Keijzer}
& \multicolumn{3}{c}{Vladislavleva}
& \multicolumn{3}{c}{Strogatz}
& \multicolumn{3}{c}{Livermore} \\
\cmidrule(lr){2-4}
\cmidrule(lr){5-7}
\cmidrule(lr){8-10}
\cmidrule(lr){11-13}
\cmidrule(lr){14-16}
\cmidrule(lr){17-19}
& \multicolumn{1}{c}{Sym} & \multicolumn{1}{c}{Acc} & \multicolumn{1}{c}{Time}
& \multicolumn{1}{c}{Sym} & \multicolumn{1}{c}{Acc} & \multicolumn{1}{c}{Time}
& \multicolumn{1}{c}{Sym} & \multicolumn{1}{c}{Acc} & \multicolumn{1}{c}{Time}
& \multicolumn{1}{c}{Sym} & \multicolumn{1}{c}{Acc} & \multicolumn{1}{c}{Time}
& \multicolumn{1}{c}{Sym} & \multicolumn{1}{c}{Acc} & \multicolumn{1}{c}{Time}
& \multicolumn{1}{c}{Sym} & \multicolumn{1}{c}{Acc} & \multicolumn{1}{c}{Time} \\
\midrule
PySR & \textbf{8.3} & \textbf{\textit{91.7}} & 99.6 & \textbf{20.0} & \textbf{\textit{60.0}} & 619.6 & \textbf{26.7} & \textbf{\textit{73.3}} & 100.8 & \textbf{12.5} & \textbf{25.0} & 126.9 & \textbf{42.9} & \textbf{\textit{71.4}} & 150.6 & \textbf{31.8} & \textbf{72.7} & 146.2 \\
Operon & 0.0 & 0.0 & 4.6 & 6.7 & \textbf{53.3} & 20.4 & 0.0 & 20.0 & 6.5 & 0.0 & 12.5 & 11.7 & 7.1 & 21.4 & 9.7 & 0.0 & 0.0 & 6.5 \\
GP-GOMEA & \textbf{8.3} & 16.7 & 29.2 & 0.0 & 26.7 & 88.9 & 6.7 & 13.3 & 33.3 & \textbf{12.5} & 12.5 & 53.2 & 14.3 & \textbf{64.3} & 45.9 & 18.2 & 18.2 & 31.4 \\
E2E-SR & 0.0 & 0.0 & 9.0 & 0.0 & 6.7 & 21.9 & 0.0 & 0.0 & 13.4 & 0.0 & 0.0 & 18.2 & 0.0 & 0.0 & 13.6 & 0.0 & 0.0 & 12.4 \\
SNIP-SR & 0.0 & 0.0 & 26.3 & 6.7 & 6.7 & 171.9 & 0.0 & 0.0 & 25.1 & 0.0 & 0.0 & 101.9 & 0.0 & 7.1 & 25.4 & 0.0 & 0.0 & 24.6 \\
GenSR & 0.0 & 0.0 & 16.0 & 0.0 & 40.0 & 234.4 & 0.0 & 6.7 & 53.7 & 0.0 & 12.5 & 357.3 & 0.0 & 21.4 & 82.6 & 0.0 & 4.5 & 23.8 \\
TPSR & 0.0 & 33.3 & 60.3 & 0.0 & 20.0 & 78.3 & 0.0 & \textbf{33.3} & 85.5 & 0.0 & \textbf{25.0} & 170.9 & 0.0 & 28.6 & 124.4 & 0.0 & 18.2 & 46.9 \\
SR4MDL & \textbf{41.7} & \textbf{41.7} & 873.5 & \textbf{20.0} & 26.7 & 1,966.4 & \textbf{20.0} & 26.7 & 1,198.3 & \textbf{12.5} & 12.5 & 2,218.6 & \textbf{50.0} & \textbf{64.3} & 874.9 & \textbf{22.7} & \textbf{27.3} & 1,181.8 \\
\midrule
MOSAIC-SR & \textbf{\textit{75.0}} & \textbf{\textit{91.7}} & 75.6 & \textbf{\textit{40.0}} & \textbf{53.3} & 288.1 & \textbf{\textit{53.3}} & \textbf{66.7} & 71.3 & \textbf{\textit{25.0}} & \textbf{\textit{37.5}} & 91.3 & \textbf{\textit{64.3}} & \textbf{\textit{71.4}} & 99.1 & \textbf{\textit{45.5}} & \textbf{\textit{90.9}} & 83.8 \\
\bottomrule
\end{tabular*}
\end{table*}

\paragraph{Compared methods.}
We compare MOSAIC-SR with search-based, neural, and hybrid symbolic regression methods.
For methods reported in the original SRSD-Feynman study, we show the published values from Matsubara et al. \citep{matsubara2022rethinking}.
The remaining methods are evaluated under our pipeline and are shown separately in the table.
These evaluations include the GP methods SBP-GP and GP-GOMEA \citep{virgolin2019semantic,virgolin2021improving}, the Transformer-MCTS hybrid TPSR \citep{shojaee2023transformer}, the reinforcement-learning and MCTS method RSRM \citep{xu2024rsrm}, and the pretrained Transformer model NeSymReS \citep{biggio2021neural}.

\paragraph{Implementation.}
MOSAIC-SR uses 200 sampled input-output pairs to generate 48 candidate sketches over three nucleus sampling passes.
MCTS explores variable assignments for these sketches, and its top 24 complete expressions initialize 24 local search trajectories with 300 mutation steps each. Constants and local edits are optimized on the training data. Final selection uses validation error and expression complexity. All methods used a budget of 3,600 seconds per problem and the settings recommended by their original implementations. Results for MOSAIC-SR and all baselines evaluated in this work are reported as means over five seeds. Cited results are taken directly from the original SRSD-Feynman paper. Runtime was not reported in the original paper, so we do not compare runtime on the SRSD-Feynman dataset. We record and analyze mean wall-clock runtime in the six cross-benchmark evaluations presented later. Experiments were run on a cluster with four 80 GB H100 GPUs, four 80 GB A100 GPUs, and CPU nodes. Each training run used a single GPU without distributed training.

\paragraph{Results.}
Table~\ref{tab:main_results} reports the SRSD-Feynman results. Numerical accuracy and symbolic recovery can differ substantially.
Some baselines obtain high \(R^2\) based accuracy but low symbolic solution rates. For example, SBP-GP reaches \(87.3\%\) accuracy on the Easy split but has an \(8.0\%\) symbolic solution rate. MOSAIC-SR reduces this gap, achieving the highest symbolic solution rate on every difficulty split and ranking among the top two methods in predictive accuracy.

We next evaluate robustness to irrelevant dummy variables. Table~\ref{tab:dummy_results} shows that dummy variables make symbolic recovery substantially harder for many baselines. For example, the Easy symbolic solution rate of PySR falls from \(60.0\%\) in the clean setting to \(20.0\%\), and that of DGSR falls from \(36.7\%\) to \(3.3\%\). MOSAIC-SR retains symbolic solution rates of \(74.0\%\), \(54.5\%\), and \(31.2\%\) on the Easy, Medium, and Hard splits, respectively. These rates exceed the strongest baseline by \(34.0\), \(18.5\), and \(14.0\) percentage points. Compared with the clean setting, they decrease by only \(10.7\), \(0.5\), and \(3.6\) percentage points. Predictive accuracy is also best on Easy, tied for best on Medium, and within \(0.8\) percentage points of the best result on Hard. These results indicate that multi-sketch initialization and structural repair preserve recovery performance when the observations contain irrelevant inputs. Appendix~\ref{app:srsd_results} provides additional error bar plots showing mean performance and variation across random seeds, together with rank comparisons for both SRSD-Feynman variants.

\subsection{Cross-Benchmark Evaluation}
\label{subsec:cross_benchmark}

We further evaluate MOSAIC-SR on six widely used symbolic regression datasets: Nguyen-12 \citep{uy2011semantic}, Korns-15 \citep{korns2011accuracy}, Keijzer-15 \citep{keijzer2003improving}, Vladislavleva-8 \citep{vladislavleva2009order}, Strogatz-14 \citep{strogatz2014nonlinear,lacava2021contemporary}, and Livermore-22 \citep{mundhenk2021symbolic}.
We compare against three groups of baselines. The search-based baselines are PySR, Operon, and GP-GOMEA \citep{cranmer2023pysr,burlacu2020operon,virgolin2021improving}. The neural baselines are E2E-SR, SNIP-SR, and GenSR \citep{kamienny2022end,meidani2024snip,li2026gensr}. The hybrid baselines are TPSR and SR4MDL \citep{shojaee2023transformer,yu2025sr4mdl}.

\paragraph{Settings.}
Numerical recovery requires test NMSE below \(10^{-6}\). We follow the standard setting for each dataset. Nguyen uses independent uniform samples from the equation domains.
Korns uses independent fit and test samples of 10,000 points each over five inputs in \([-50,50]\).
Keijzer and Vladislavleva use their equation specific sampling methods and sample sizes.
Livermore follows the equation specific domains and sample counts distributed with DSO \citep{petersen2021deep}. Strogatz uses the fixed 75/25 SRBench split of the PMLB ODE datasets \citep{lacava2021contemporary}.
For benchmarks without a predefined validation set, we select the final candidate using the training data. MOSAIC-SR uses the same 24-by-300 search configuration on every dataset. All baselines use their default budgets.

\paragraph{Results.}
Table~\ref{tab:cross_benchmark} summarizes the multi-benchmark results. It reports symbolic solution rate, predictive accuracy, and mean wall-clock time per equation. MOSAIC-SR obtains the highest symbolic solution rate on every dataset. Improvements over the strongest baseline range from \(12.5\) to \(33.3\) percentage points.
These gains cannot be explained by predictive accuracy alone. MOSAIC-SR and PySR have identical numerical recovery on Nguyen and Strogatz, but MOSAIC-SR increases symbolic solution rates from 8.3\% to 75.0\% and from 42.9\% to 64.3\%. Although PySR has slightly higher numerical recovery on Korns and Keijzer, MOSAIC-SR approximately doubles symbolic recovery on both datasets. Moreover, the neural baselines reach at most 6.7\% symbolic recovery. Operon, GP-GOMEA, and direct neural generators are generally faster, but their symbolic recovery is substantially lower. In contrast, MOSAIC-SR transfers across benchmarks and more effectively converts accurate predictions into exact equations. Among competitive recovery baselines, MOSAIC-SR is 1.3-2.2$\times$ faster than PySR and 6.8-24.3$\times$ faster than SR4MDL across all six datasets. This highlights the importance of search and repair after neural generation.

\section{Conclusion}
\label{sec:conclusion}

We presented MOSAIC-SR, a hybrid symbolic regression method that combines learned equation priors with explicit recovery search. A pretrained Transformer identifies promising expression structures instead of relying on random initialization. Conditional MCTS completes partially specified sketches. Within each search path, local tree edits alternate with constant refitting throughout the search budget. This repeated interaction corrects errors in both symbolic structure and numerical scale.

Across various benchmarks, MOSAIC-SR achieves the highest symbolic solution rate in every reported dataset while retaining competitive predictive accuracy. Its advantage persists when irrelevant variables are introduced. This supports the use of learned sketches as search priors rather than one-shot predictions. The runtime results further suggest that a learned sketch prior can reduce search time while supporting higher symbolic solution rates. Despite these gains, recovery becomes more difficult for high dimensional equations because the corresponding search space grows rapidly. Targets far outside the pretraining distribution may also require a larger search budget. Future work will use broader pretraining distributions to improve recovery for high-dimensional equations with complex symbolic structure. We will also study recovery from sparse observations, which better reflect data availability in real world physical systems.

\clearpage
\begingroup
\hbadness=10000
\vbadness=10000
\bibliographystyle{AuthorKit27/aaai2027}
\bibliography{AuthorKit27/aaai2027}
\endgroup

\clearpage
\appendix
\renewcommand{\thesection}{\Alph{section}}

\setcounter{topnumber}{3}
\setcounter{bottomnumber}{2}
\setcounter{totalnumber}{5}
\renewcommand{\topfraction}{0.9}
\renewcommand{\bottomfraction}{0.8}
\renewcommand{\textfraction}{0.08}
\renewcommand{\floatpagefraction}{0.8}

\section{Visualization of SRSD-Feynman Results}
\label{app:srsd_results}

We provide predictive accuracy and symbolic recovery visualizations for SRSD-Feynman \citep{matsubara2022rethinking}, together with rank comparisons involving expression complexity. The quantitative results are reported in the paper.

\subsection{Results without Dummy Variables}

\begin{figure}[!htbp]
\centering
\begin{minipage}[t]{0.48\textwidth}
\centering
\includegraphics[width=\linewidth,trim=0bp 0bp 390.49bp 45.76bp,clip]{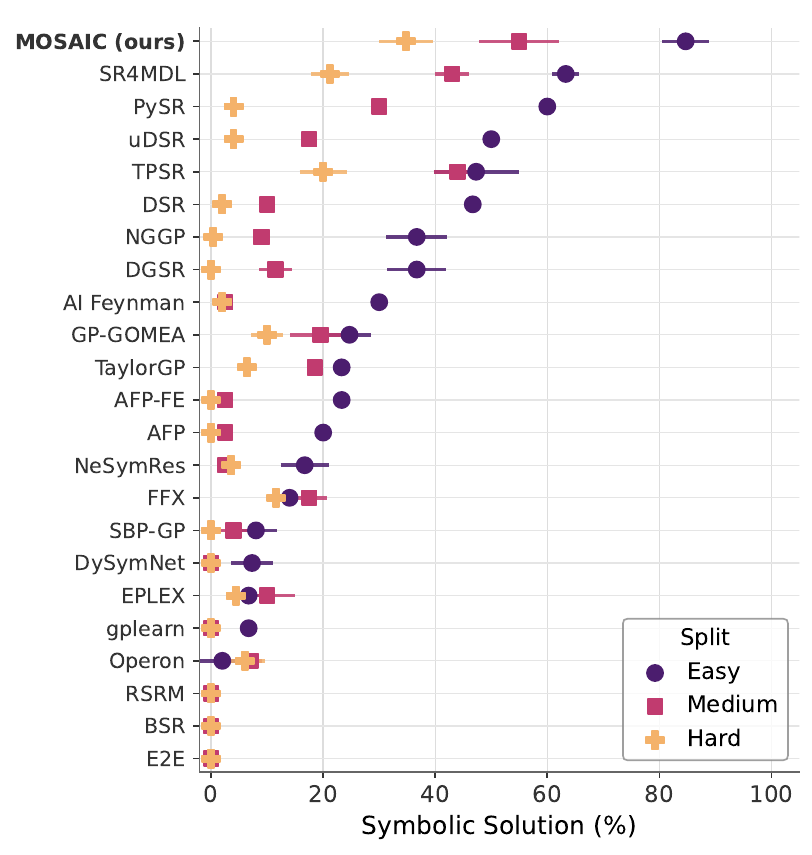}
\end{minipage}
\hfill
\begin{minipage}[t]{0.48\textwidth}
\centering
\includegraphics[width=\linewidth,trim=390bp 0bp 0bp 45.76bp,clip]{Figures/srsd/srsd_nondummy.pdf}
\end{minipage}
\caption{Predictive accuracy and symbolic solution rate on SRSD-Feynman without dummy variables. Points show mean performance on the Easy, Medium, and Hard splits. Horizontal bars show the reported variation across seeds.}
\label{fig:srsd_nondummy_overview}
\end{figure}

Figure~\ref{fig:srsd_nondummy_overview} compares predictive accuracy and symbolic solution rate across the Easy, Medium, and Hard splits without dummy variables.

MOSAIC-SR achieves the highest symbolic solution rate on all three splits while remaining among the most accurate methods. Its symbolic advantage is especially clear on the Easy split and remains substantial as equation difficulty increases.

\begin{figure}[!htbp]
\centering
\begin{minipage}[t]{0.44\textwidth}
\centering
\includegraphics[width=\linewidth]{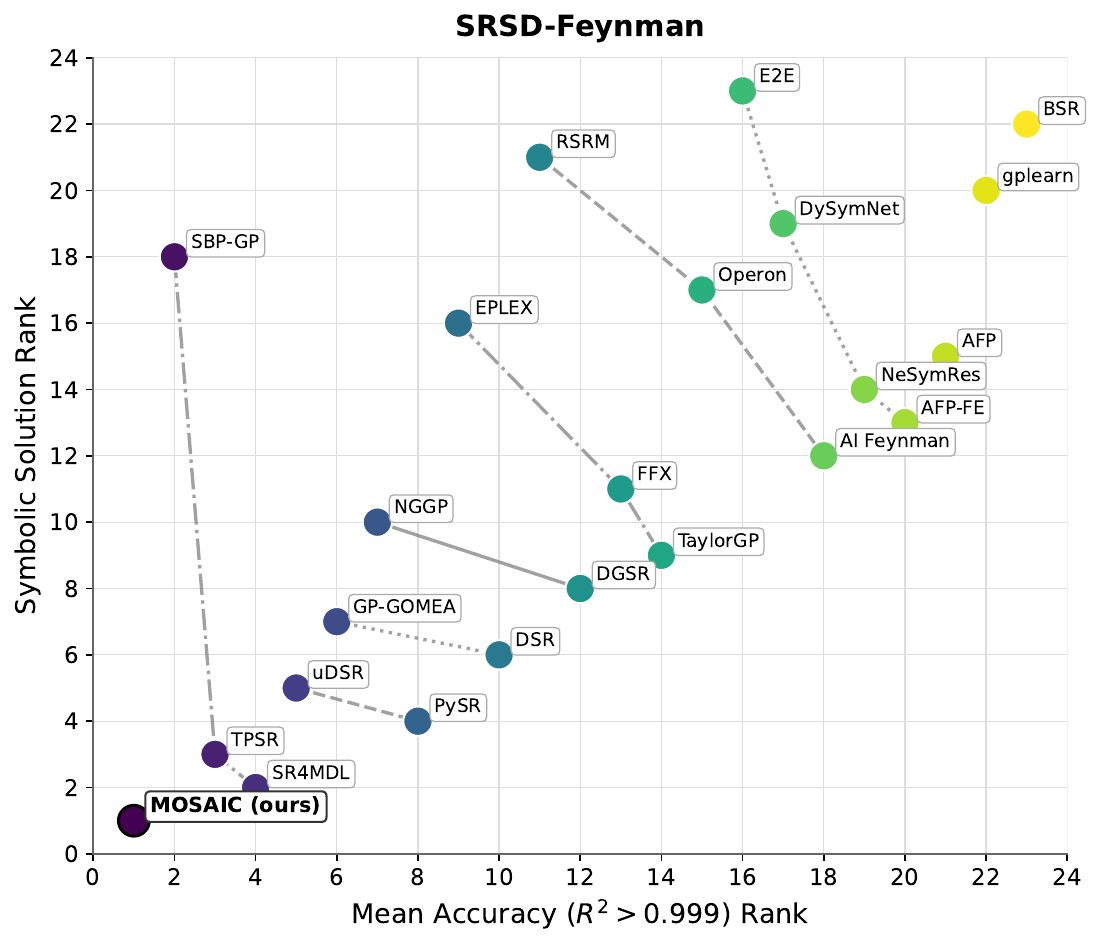}
\textbf{(a) Accuracy versus symbolic recovery}
\end{minipage}
\hfill
\begin{minipage}[t]{0.44\textwidth}
\centering
\includegraphics[width=\linewidth]{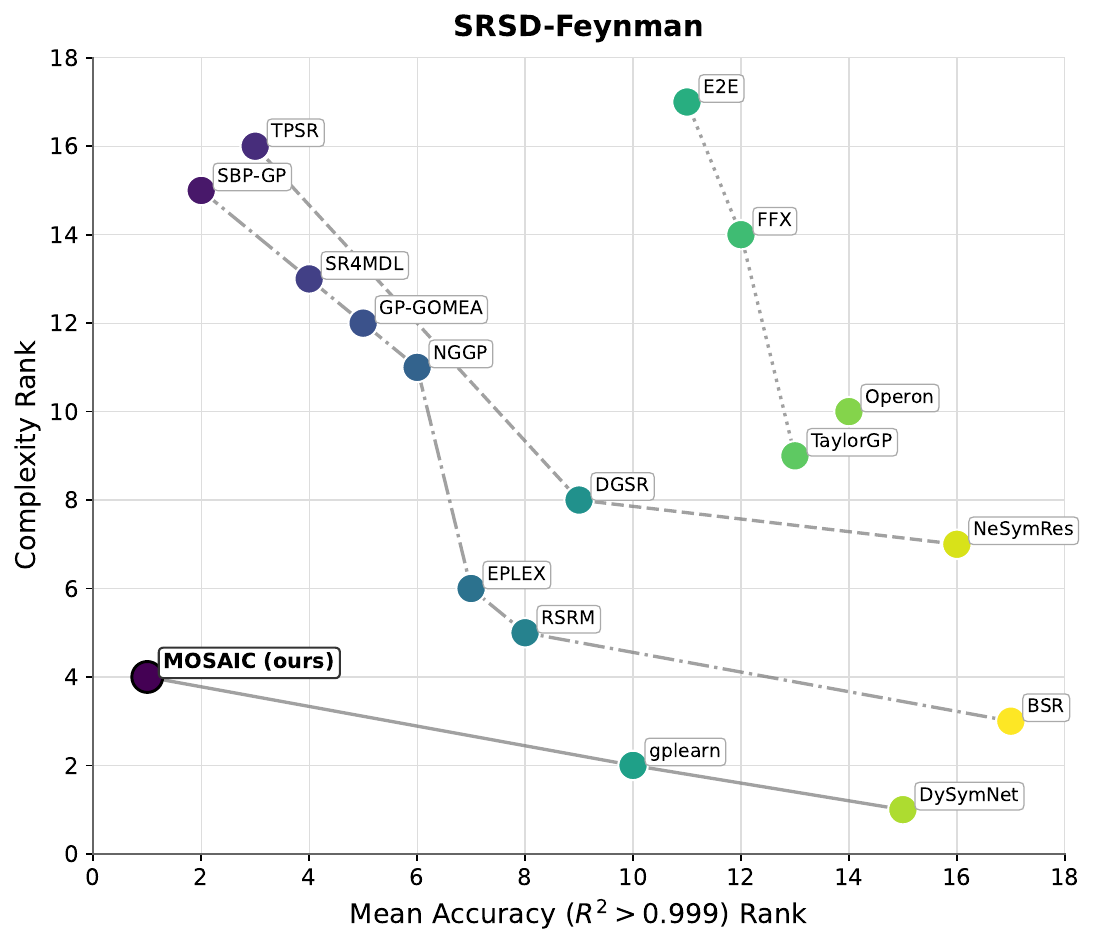}
\textbf{(b) Accuracy versus complexity}
\end{minipage}
\caption{Rank-based comparisons on SRSD-Feynman without dummy variables. The panels compare mean predictive accuracy rank with symbolic solution rank and expression complexity rank, respectively. Methods closer to the lower left corner achieve better tradeoffs.}
\label{fig:srsd_nondummy_rank_tradeoffs}
\end{figure}
Figure~\ref{fig:srsd_nondummy_rank_tradeoffs} summarizes the rank based tradeoffs without dummy variables. Lower ranks are better in both panels. MOSAIC-SR ranks first in both predictive accuracy and symbolic recovery, giving the strongest joint result in Figure~\ref{fig:srsd_nondummy_rank_tradeoffs}(a). It also achieves top four complexity rank in Figure~\ref{fig:srsd_nondummy_rank_tradeoffs}(b). Methods with slightly simpler expressions have lower accuracy ranks.

\subsection{Results with Dummy Variables}

\begin{figure}[!htbp]
\centering
\begin{minipage}[t]{0.48\textwidth}
\centering
\includegraphics[width=\linewidth,trim=0bp 0bp 390.49bp 45.76bp,clip]{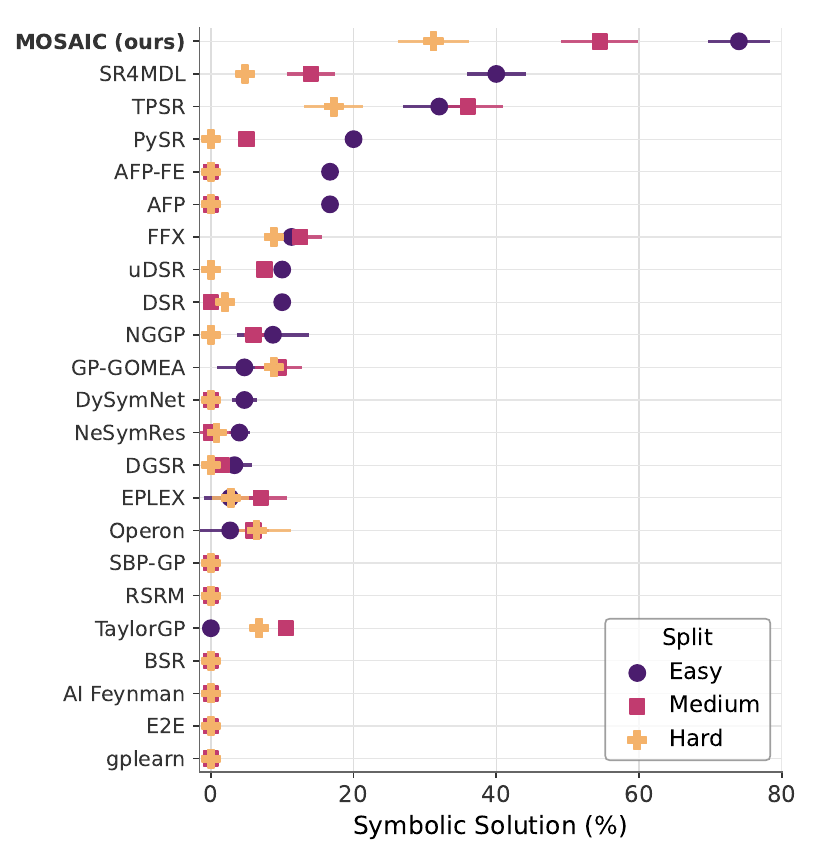}
\end{minipage}
\hfill
\begin{minipage}[t]{0.48\textwidth}
\centering
\includegraphics[width=\linewidth,trim=390bp 0bp 0bp 45.76bp,clip]{Figures/srsd/srsd_dummy.pdf}
\end{minipage}
\caption{Predictive accuracy and symbolic solution rate on SRSD-Feynman with dummy variables. Points show mean performance on the Easy, Medium, and Hard splits. Horizontal bars show the reported variation across seeds.}
\label{fig:srsd_dummy_overview}
\end{figure}

The dummy-variable setting adds irrelevant input columns and therefore tests robustness to variable selection. Figure~\ref{fig:srsd_dummy_overview} compares predictive and symbolic performance under this perturbation.

\begin{figure}[H]
\centering
\begin{minipage}[t]{0.44\textwidth}
\centering
\includegraphics[width=\linewidth]{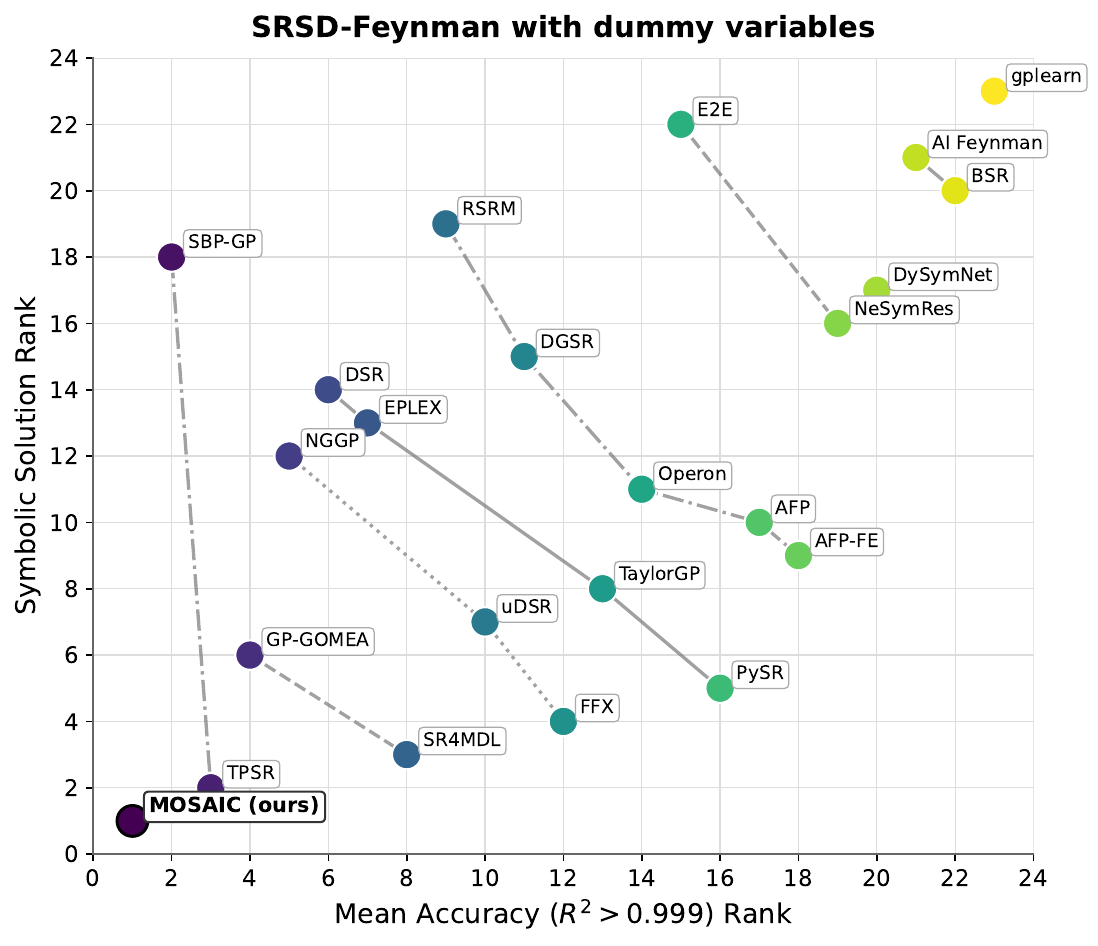}
\textbf{(a) Accuracy versus symbolic recovery}
\end{minipage}
\hfill
\begin{minipage}[t]{0.44\textwidth}
\centering
\includegraphics[width=\linewidth]{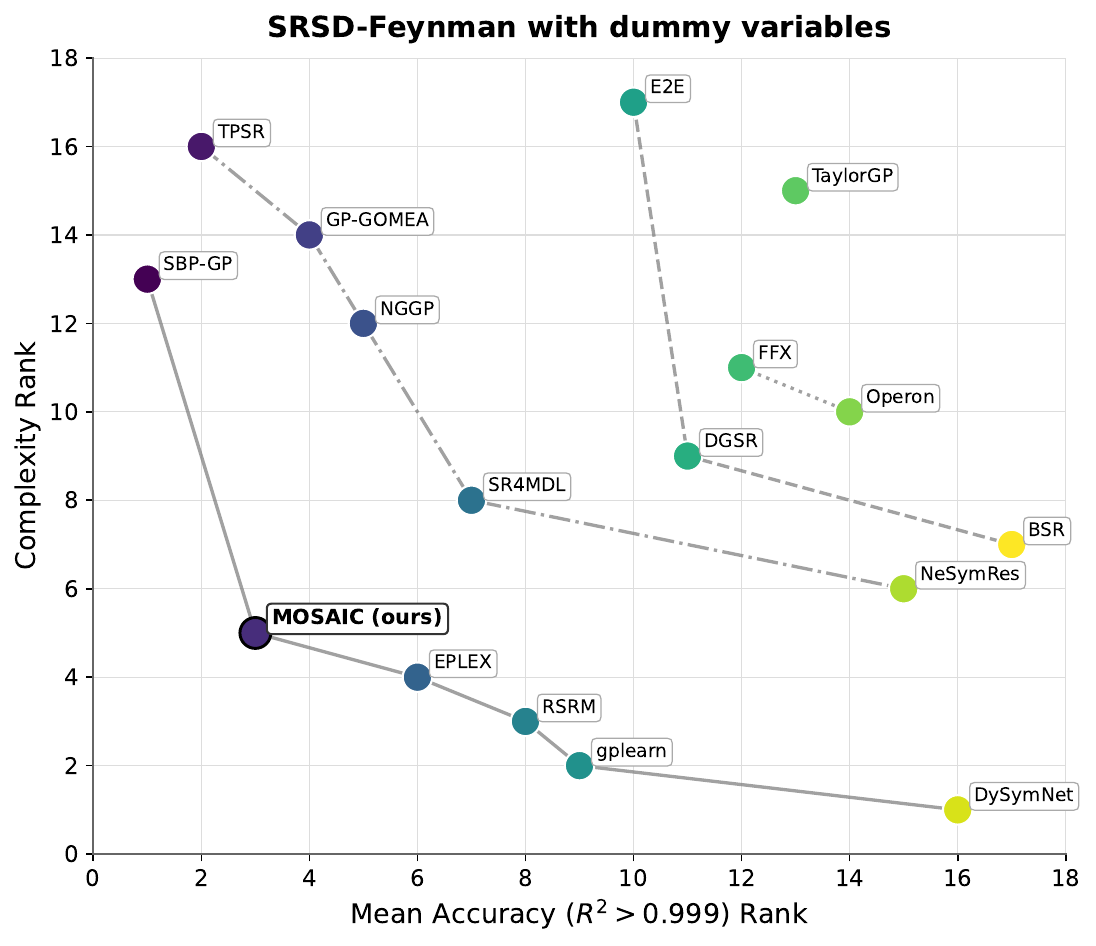}
\textbf{(b) Accuracy versus complexity}
\end{minipage}
\caption{Rank based comparisons on SRSD-Feynman with dummy variables. The panels compare mean predictive accuracy rank with symbolic solution rank and expression complexity rank, respectively. Methods closer to the lower left corner achieve better tradeoffs.}
\label{fig:srsd_dummy_rank_tradeoffs}
\end{figure}

MOSAIC-SR continues to achieve the highest symbolic solution rate across Easy, Medium, and Hard. Although irrelevant variables reduce absolute recovery on some splits, the ordering of the leading methods is preserved. Predictive accuracy also remains highest or close to highest across the three difficulty levels. Figure~\ref{fig:srsd_dummy_rank_tradeoffs} presents the corresponding rank based comparisons.

With dummy variables, MOSAIC-SR remains first in the joint accuracy-symbolic comparison. Its complexity rank is fifth while its accuracy rank remains near the top. This shows that robustness to irrelevant inputs does not require a large increase in recovered expression complexity.

\FloatBarrier

\section{Validation-NMSE Trajectories during Local Search}
\label{app:val_nmse_trajectories}

Figures~\ref{fig:val_nmse_easy_true_ops}--\ref{fig:val_nmse_hard_decoded_ops} track the best validation NMSE achieved so far throughout each MOSAIC-SR local-search run. Lower values are better and the vertical axis is logarithmic. Thin gray step curves show individual runs, the colored curve is the median, and the shaded region is the interquartile range. The dashed horizontal line marks NMSE \(=10^{-3}\), equivalent to \(R^2>0.999\), which is the accuracy threshold used for SRSD-Feynman. To distinguish target difficulty from proposal quality, each split is analyzed first by ground-truth operator count and then by the operator count of the top-1 decoded sketch. Operator count is used as a proxy for structural size. It is positively related to expression length and complexity but is not identical to either. Easy and Medium trajectories are shown for 50 seconds, while Hard trajectories are shown for 100 seconds. All three splits use five seeds.

\subsection{SRSD-Feynman Easy}

\begin{figure}[!htbp]
\centering
\includegraphics[width=\textwidth]{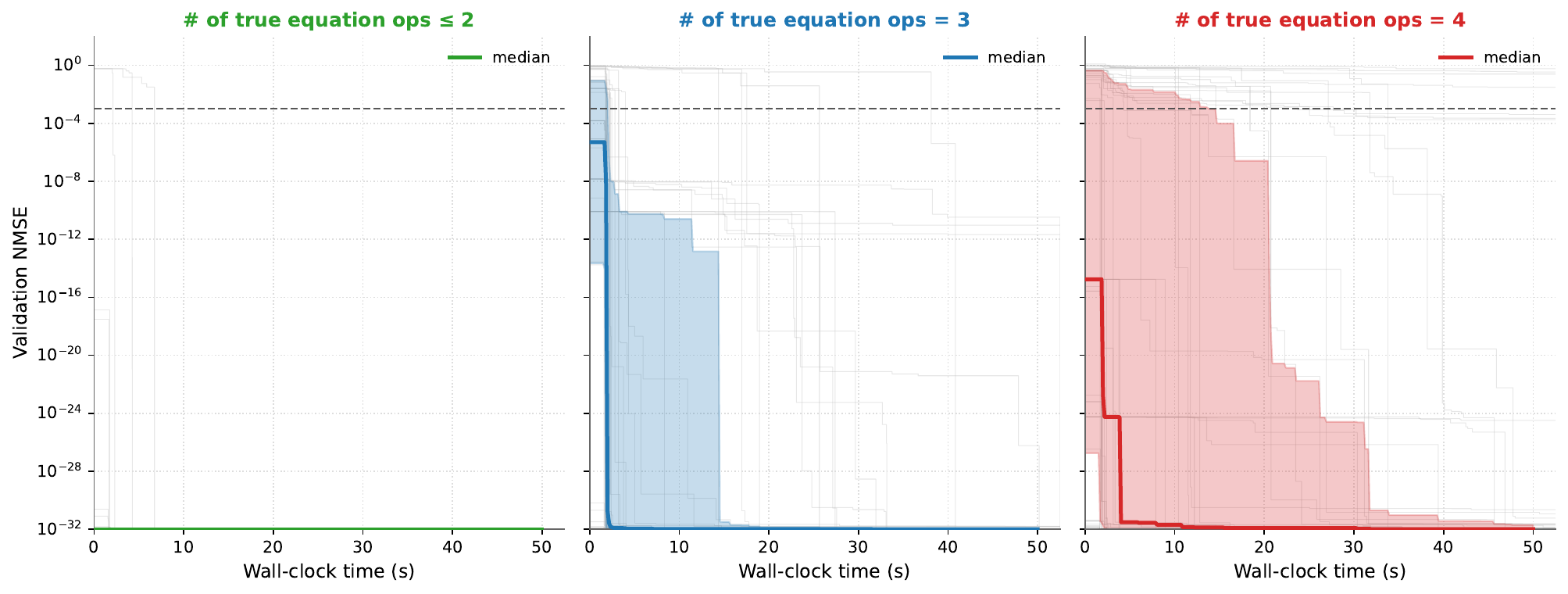}
\caption{Best validation NMSE versus wall-clock time on SRSD-Feynman Easy, grouped by ground-truth operator count. The 150 runs comprise 30 equations evaluated over five seeds. The groups contain one or two, three, and four true operators, with 40, 55, and 55.}
\label{fig:val_nmse_easy_true_ops}
\end{figure}

\paragraph{Ground-truth operator count.}
Figure~\ref{fig:val_nmse_easy_true_ops} first groups runs by the number of operators in the true equation. Equations with fewer operators reach low NMSE earlier. The three and four operator groups nevertheless cross the accuracy threshold soon afterward and converge to the same low error regime.

\begin{figure}[!htbp]
\centering
\includegraphics[width=\textwidth]{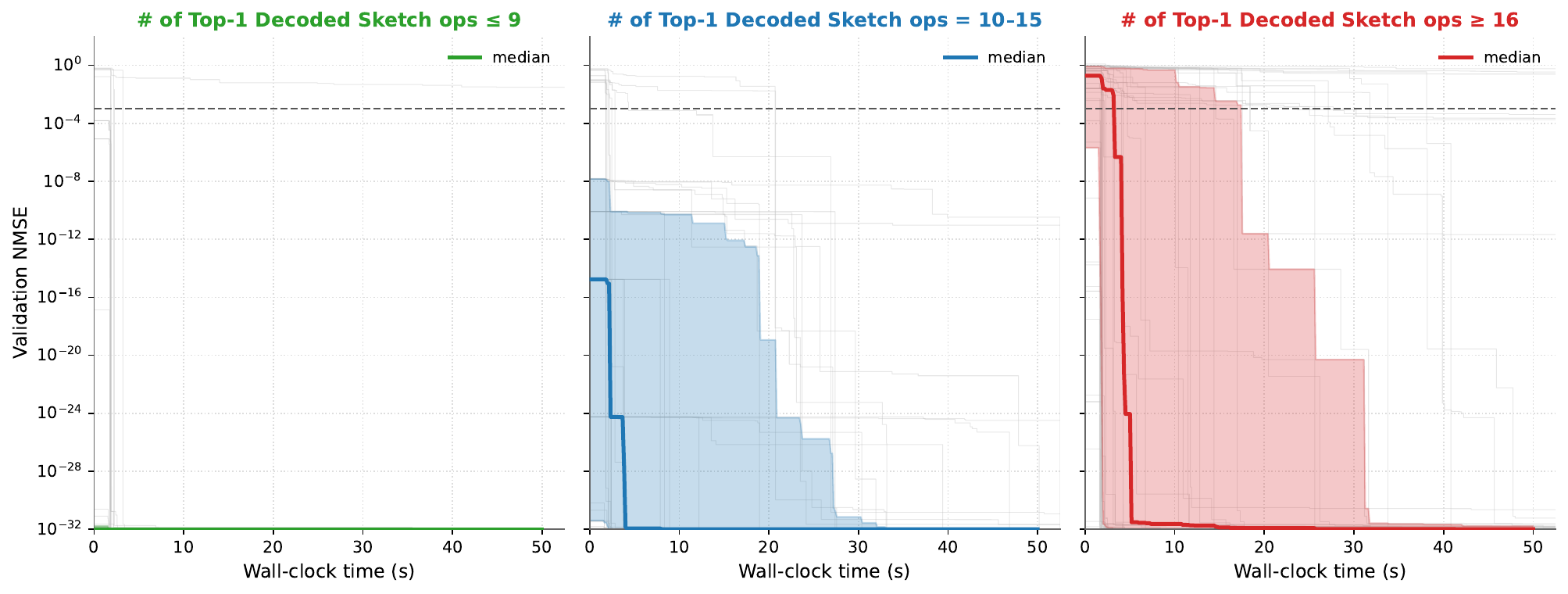}
\caption{Best validation NMSE versus wall-clock time on SRSD-Feynman Easy, grouped by top-1 decoded sketch operator count. The runs are divided into sketches with fewer than 10, between 10 and 15, and more than 15 operators, with 57, 44, and 49 runs respectively.}
\label{fig:val_nmse_easy_decoded_ops}
\end{figure}

\paragraph{Decoded sketch operator count.}
Figure~\ref{fig:val_nmse_easy_decoded_ops} shows a much stronger separation by decoded sketch operator count. Sketches with fewer than 10 operators cross the threshold within a few seconds, while those with 10-15 operators converge more slowly and some runs with more than 15 operators remain above it. Because Easy targets contain at most four operators, high decoded operator counts appear to constrain local search more than ground truth operator count.

\FloatBarrier
\subsection{SRSD-Feynman Medium}

\begin{figure}[!htbp]
\centering
\includegraphics[width=\textwidth]{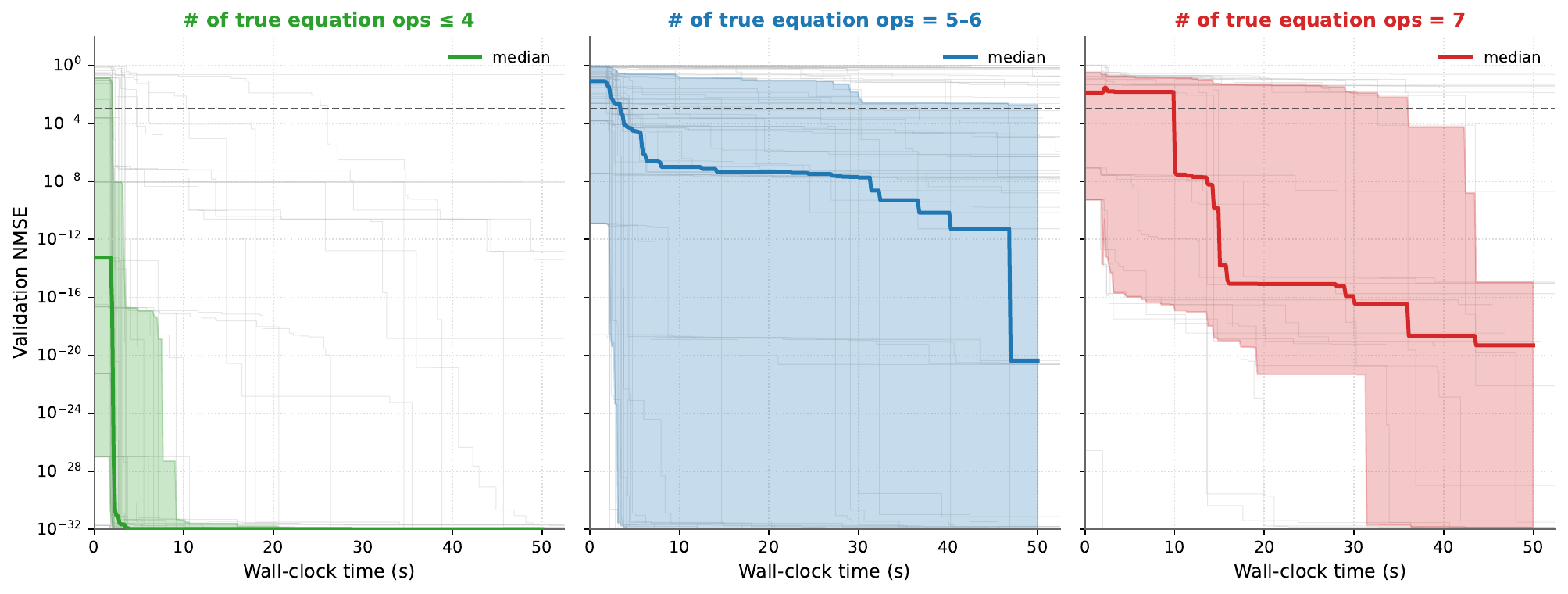}
\caption{Best validation NMSE versus wall-clock time on SRSD-Feynman Medium, grouped by ground-truth operator count. The 200 runs comprise 40 equations evaluated over five seeds. The groups contain fewer than five, five or six, and more than six true operators, with 75, 95, and 30 runs, respectively.}
\label{fig:val_nmse_medium_true_ops}
\end{figure}

Figure~\ref{fig:val_nmse_medium_true_ops} groups runs by the number of operators in the true equation. Equations with fewer than five operators reach low NMSE earlier, while the two higher-count groups improve more slowly and show wider variation. Their medians are not strictly ordered by operator count, indicating that operator count captures only one component of target difficulty. The remaining variation depends on the specific algebraic structure of each equation.

\begin{figure}[!htbp]
\centering
\includegraphics[width=\textwidth]{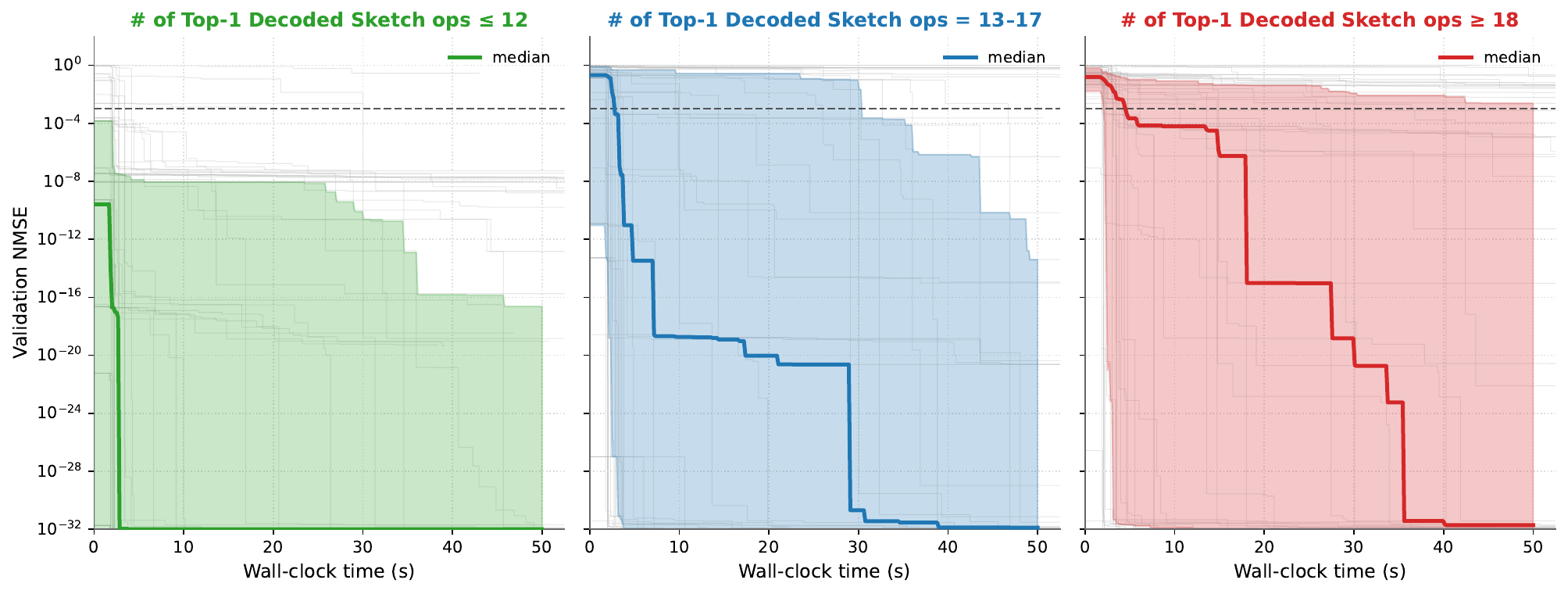}
\caption{Best validation NMSE versus wall-clock time on SRSD-Feynman Medium, grouped by top-1 decoded sketch operator count. The groups contain fewer than 13, between 13 and 17, and more than 17 decoded operators, with 82, 53, and 65 runs respectively.}
\label{fig:val_nmse_medium_decoded_ops}
\end{figure}

\paragraph{Decoded sketch operator count.}
Figure~\ref{fig:val_nmse_medium_decoded_ops} shows a clearer separation by decoded sketch operator count. Sketches with fewer than 13 operators cross the accuracy threshold in about 10 seconds, those with 13--17 operators require closer to 30 seconds, and the median for sketches with more than 17 operators remains above it. Decoded sketch operator count therefore provides a stronger indication of local-search progress.

\FloatBarrier
\subsection{SRSD-Feynman Hard}

\begin{figure}[!htbp]
\centering
\includegraphics[width=\textwidth]{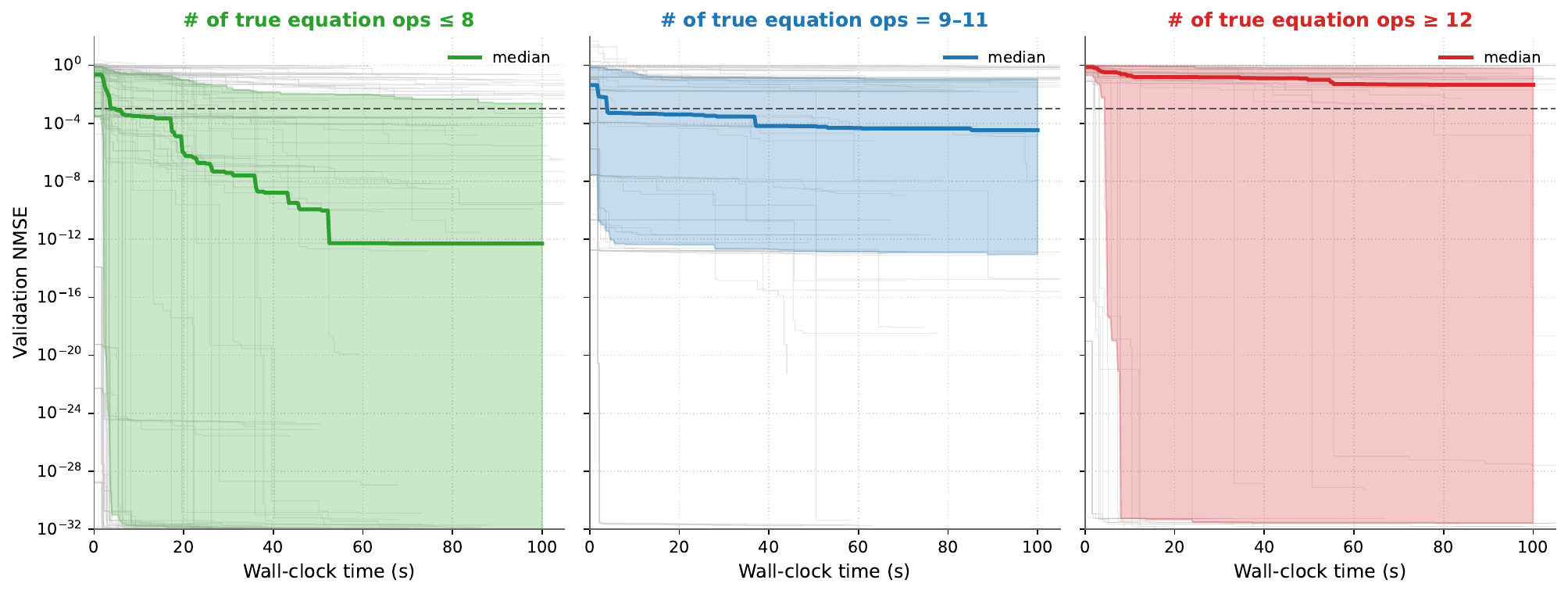}
\caption{Best validation NMSE versus wall-clock time during the 100-second search window on SRSD-Feynman Hard, grouped by ground-truth operator count. The 250 runs comprise 50 equations evaluated over five seeds. The groups contain fewer than nine, between nine and 11, and more than 11 true operators, with 115, 75, and 60 runs, respectively.}
\label{fig:val_nmse_hard_true_ops}
\end{figure}

\paragraph{Ground-truth operator count.}
Figure~\ref{fig:val_nmse_hard_true_ops} groups runs by the number of operators in the true equation. The medians for equations with fewer than nine operators and those with nine to 11 operators both cross the accuracy threshold early. The lower-count group subsequently reaches much lower NMSE. The median for equations with more than 11 operators improves gradually but remains above the threshold after 100 seconds. Ground-truth operator count therefore captures an important component of local-search difficulty on Hard.

\begin{figure}[!htbp]
\centering
\includegraphics[width=\textwidth]{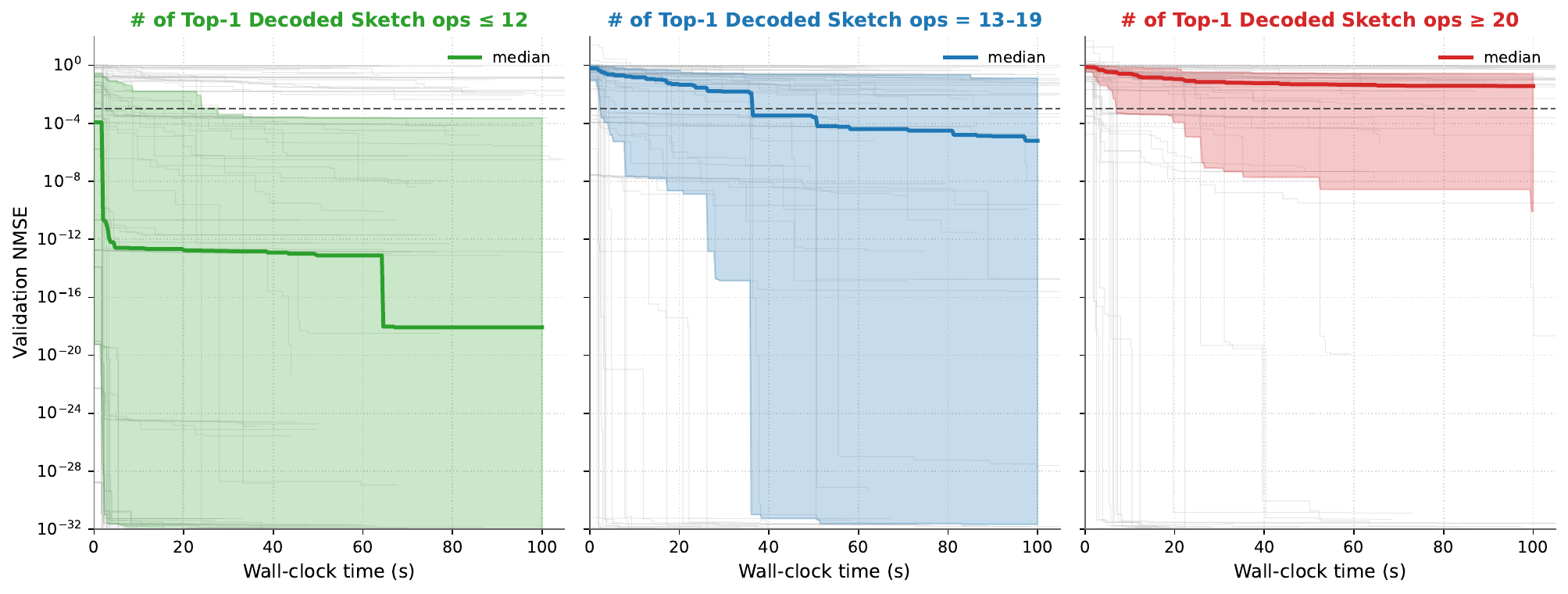}
\caption{Best validation NMSE versus wall-clock time on SRSD-Feynman Hard, grouped by top-1 decoded sketch operator count. The x-axis spans 0--100 seconds. The 250 runs comprise 50 equations evaluated over five seeds. The groups contain fewer than 13, between 13 and 19, and more than 19 decoded operators, with 107, 73, and 70 runs, respectively.}
\label{fig:val_nmse_hard_decoded_ops}
\end{figure}

\paragraph{Decoded sketch operator count.}
Figure~\ref{fig:val_nmse_hard_decoded_ops} shows a stronger separation by decoded sketch operator count. The median for sketches with fewer than 13 operators crosses the accuracy threshold almost immediately. The median for sketches with 13-19 operators crosses after about 40 seconds. The median for sketches with more than 19 operators remains above the threshold until the end of the 100 second window. A high decoded operator count therefore slows local search and may require the full available budget to reach the threshold.

\FloatBarrier
\paragraph{Overall implication.}
Across all three splits, higher ground truth operator counts generally slow convergence. In comparison, decoded sketch operator count produces a stronger and more consistent separation of local-search outcomes. This finding is practically useful. Ground truth operator count is unknown at inference time. However, decoded sketch operator count is directly observable. These results motivate operator count aware decoding and resampling of sketches with unusually high operator counts. They also support adaptive allocation of the local-search budget based on proposal size.

\FloatBarrier

\section{Ablation Studies}
\label{app:ablations}

We conduct the ablations on the Easy split of SRSD-Feynman, which contains 30 equations. Each result reports symbolic recovery under the same verification protocol as the main SRSD-Feynman experiment, together with numerical recovery. The component ablation is evaluated with three method seeds: 0, 1, and 2.

\paragraph{Contribution of each recovery stage.}
Table~\ref{tab:ablation_stages} compares sketch decoding alone, decoding followed by conditional MCTS, and the complete recovery procedure. Decoder only disables both MCTS and mutation search. The second stage adds conditional MCTS, while Full MOSAIC further applies 24 independent search paths with 300 mutations per path. Conditional MCTS improves numerical recovery for all three seeds and increases symbolic recovery for two of them. The complete procedure achieves 83.3--90.0\% symbolic recovery, compared with 30.0--43.3\% before mutation search, showing that the learned prior provides useful starting structures but does not replace equation repair.

\begin{table}[H]
\centering
\scriptsize
\setlength{\tabcolsep}{3pt}
\caption{Component ablation on 30 clean SRSD-Feynman Easy equations. Sym and Acc report symbolic and numerical recovery. All values are percentages. Decoder, Decoder + MCTS, and Full MOSAIC are evaluated with the same three method seeds.}
\label{tab:ablation_stages}
\begin{tabular}{lcccccc}
\toprule
& \multicolumn{2}{c}{\makebox[0.14\textwidth][c]{Decoder}}
& \multicolumn{2}{c}{\makebox[0.14\textwidth][c]{Decoder + MCTS}}
& \multicolumn{2}{c}{\makebox[0.14\textwidth][c]{Full MOSAIC}} \\
\cmidrule(lr){2-3}\cmidrule(lr){4-5}\cmidrule(lr){6-7}
Seed & Sym & Acc & Sym & Acc & Sym & Acc \\
\midrule
0 & 36.7 & 60.0 & 43.3 & 66.7 & 83.3 & 96.7 \\
1 & 30.0 & 46.7 & 30.0 & 60.0 & 83.3 & 96.7 \\
2 & 30.0 & 46.7 & 40.0 & 66.7 & \textbf{90.0} & \textbf{100.0} \\
\bottomrule
\end{tabular}
\end{table}

\paragraph{Search breadth and depth.}
Table~\ref{tab:ablation_path_budget} varies the number of independent search paths and the number of mutation steps per path. The reported recovery rates are averaged over three method seeds. Paths are independent, and the columns are snapshots from continuous trajectories.
Recovery generally improves with both search breadth and trajectory depth. At 500 steps, increasing the number of paths from one to 48 raises numerical recovery from 80.0\% to 98.9\% and symbolic recovery from 56.7\% to 87.8\%. The gains from additional steps become smaller at larger search budgets, with numerical recovery at 48 paths reaching 98.9\% after 300 steps and remaining unchanged at 500 steps, while symbolic recovery increases from 85.6\% to 87.8\%. The gains from increasing the number of paths from 24 to 48 are also modest, improving numerical recovery by 1.1 percentage points and symbolic recovery by 2.2 percentage points at 500 steps.

\begin{table}[H]
\centering
\captionsetup{font=footnotesize,skip=3pt}
\scriptsize
\renewcommand{\arraystretch}{0.9}
\setlength{\tabcolsep}{2.5pt}
\caption{Search budget ablation on 30 SRSD-Feynman Easy equations. Sym and Acc denote symbolic and numerical recovery percentages averaged over three method seeds.}
\label{tab:ablation_path_budget}
\begin{tabular}{lcccccc}
\toprule
& \multicolumn{2}{c}{\makebox[0.14\textwidth][c]{100 steps}}
& \multicolumn{2}{c}{\makebox[0.14\textwidth][c]{300 steps}}
& \multicolumn{2}{c}{\makebox[0.14\textwidth][c]{500 steps}} \\
\cmidrule(lr){2-3}\cmidrule(lr){4-5}\cmidrule(lr){6-7}
Paths & Sym & Acc & Sym & Acc & Sym & Acc \\
\midrule
1  & 47.8 & 70.0 & 53.3 & 76.7 & 56.7 & 80.0 \\
6  & 57.8 & 77.8 & 70.0 & 87.8 & 75.6 & 90.0 \\
12 & 66.7 & 83.3 & 76.7 & 93.3 & 82.2 & 94.4 \\
24 & 70.0 & 84.4 & 85.6 & 97.8 & 85.6 & 97.8 \\
48 & 73.3 & 87.8 & 85.6 & 98.9 & \textbf{87.8} & \textbf{98.9} \\
\bottomrule
\end{tabular}
\end{table}

\FloatBarrier

\section{Implementation Details}
\label{app:implementation}

The key implementation settings used in the reported experiments are summarized below.

\begingroup
\setlength{\LTpre}{6pt}
\setlength{\LTpost}{8pt}
\renewcommand{\arraystretch}{1.02}

\scriptsize
\setlength{\tabcolsep}{5pt}
\begin{longtable}{@{}lcc@{}}
\caption{Core optimization settings used in the two pretraining stages.}
\label{tab:training_config} \\
\toprule
Setting & Stage 1 & Stage 2 \\
\midrule
Batch size & \(128\) & \(128\) \\
Training steps & \(400{,}000\) & \(250{,}000\) \\
Peak learning rate & \(10^{-4}\) & \(10^{-5}\) \\
Encoder learning rate & \(10^{-4}\) & \(5\times10^{-6}\) \\
Warmup steps & \(16{,}000\) & \(2{,}000\) \\
Learning-rate schedule & Cosine decay & Constant \\
AdamW \(\beta_2\) & \(0.999\) & \(0.95\) \\
Weight decay & \(0.01\) & \(0.01\) \\
Gradient clipping & \(1.0\) & \(1.0\) \\
\bottomrule
\end{longtable}

\scriptsize
\setlength{\tabcolsep}{6pt}
\begin{longtable}{@{}lc@{}}
\caption{Detailed Transformer architecture used by MOSAIC-SR, including the numeric encoder, symbolic encoder, and symbolic decoder. The complete architecture has approximately \(17.4\) million parameters and is relatively lightweight.}
\label{tab:cross_benchmark_architecture} \\
\toprule
Parameter & Value \\
\midrule
\endfirsthead
\toprule
Parameter & Value \\
\midrule
\endhead
\multicolumn{2}{@{}l}{\textit{Set-Transformer numeric encoder}} \\
Architecture & Set Transformer \\
Input dimension & 11 (10 variables + output) \\
ISAB blocks & 4 \\
Hidden dimension & 256 \\
Attention heads & 8 \\
Inducing points per ISAB & 32 \\
PMA seed vectors & 16 \\
Residual normalization & Pre-LayerNorm \\
PMA output dimension & \(16\times256=4{,}096\), then projected \\
\midrule
\multicolumn{2}{@{}l}{\textit{Symbolic Transformer encoder}} \\
Encoder layers & 6 \\
Hidden dimension & 256 \\
Attention heads & 8 \\
Feed-forward dimension & \(1{,}024\) \\
Dropout & 0.1 \\
Activation & GELU \\
\midrule
\multicolumn{2}{@{}l}{\textit{Symbolic Transformer decoder}} \\
Decoder layers & 4 \\
Hidden dimension & 256 \\
Attention heads & 8 \\
Feed-forward dimension & \(1{,}024\) \\
Dropout & 0.1 \\
Maximum output length & 128 tokens \\
\midrule
\multicolumn{2}{@{}l}{\textit{Shared settings}} \\
Vocabulary size & 45 \\
Maximum variables & 10 \\
Total parameters & Approximately \(17.4\) M \\
\bottomrule
\end{longtable}

\scriptsize
\setlength{\tabcolsep}{6pt}
\begin{longtable}{@{}lc@{}}
\caption{MOSAIC-SR inference time decoder, MCTS, and local-search budget for the cross-benchmark experiments.}
\label{tab:cross_benchmark_search} \\
\toprule
Parameter & Value \\
\midrule
\endfirsthead
\toprule
Parameter & Value \\
\midrule
\endhead
Independent search paths & 24 \\
Mutations per path & 300 \\
Decoder/search passes & 6 \\
Decoder sketches retained & 24 \\
Workers & 12 \\
Conditional MCTS & Enabled \\
MCTS trigger / expansions & Validation NMSE \(10^{-4}\) / \(1{,}500\) \\
Maximum evolved AST size & 32 \\
Constant-fitting starts & 16 \\
\bottomrule
\end{longtable}

\scriptsize
\setlength{\tabcolsep}{4pt}
\begin{longtable}{@{}lcccc@{}}
\caption{SRSD-Feynman benchmark groups, variable count ranges, dummy variable distribution, data split sizes, and input domain extrema \citep{matsubara2022rethinking}. The extrema summarize equation specific sampling distributions rather than a shared interval. Dummy counts give the numbers of problems containing one, two, or three irrelevant variables.}
\label{tab:srsd_benchmark_groups} \\
\toprule
Group & Problems & Variables & 1/2/3 dummy & Train/Val/Test \\
\midrule
\multicolumn{2}{@{}l}{\textit{Without dummy variables}} & \multicolumn{3}{r@{}}{Input domain extrema: \(10^{-30}\)--\(10^{4}\)} \\
\addlinespace[2pt]
Easy & 30 & 1--5 & 0/0/0 & \(8{,}000/1{,}000/1{,}000\) \\
Medium & 40 & 1--6 & 0/0/0 & \(8{,}000/1{,}000/1{,}000\) \\
Hard & 50 & 1--8 & 0/0/0 & \(8{,}000/1{,}000/1{,}000\) \\
\midrule
\multicolumn{2}{@{}l}{\textit{With dummy variables}} & \multicolumn{3}{r@{}}{Input domain extrema: \(10^{-33}\)--\(10^{33}\)} \\
\addlinespace[2pt]
Easy + Dummy & 30 & 3--8 & 6/11/13 & \(8{,}000/1{,}000/1{,}000\) \\
Medium + Dummy & 40 & 3--8 & 12/18/10 & \(8{,}000/1{,}000/1{,}000\) \\
Hard + Dummy & 50 & 3--10 & 13/18/19 & \(8{,}000/1{,}000/1{,}000\) \\
\bottomrule
\end{longtable}

\scriptsize
\setlength{\tabcolsep}{5pt}
\begin{longtable}{@{}lcccc@{}}
\caption{Numbers of problems, variable count ranges, training and test sample counts for the Nguyen \citep{uy2011semantic}, Korns \citep{korns2011accuracy}, Keijzer \citep{keijzer2003improving}, Vladislavleva \citep{vladislavleva2009order}, Strogatz \citep{strogatz2014nonlinear,lacava2021contemporary}, and Livermore \citep{mundhenk2021symbolic} benchmarks.}
\label{tab:cross_benchmark_sizes} \\
\toprule
Dataset & Problems & Variables & Training rows & Test rows \\
\midrule
Nguyen & 12 & 1--2 & 20 & 20 \\
Korns & 15 & 5 & \(10{,}000\) & \(10{,}000\) \\
Keijzer & 15 & 1--3 & 20--\(1{,}000\) & 120--\(361{,}201\) \\
Vladislavleva & 8 & 1--5 & 30--\(1{,}024\) & 221--\(93{,}636\) \\
Strogatz & 14 & 2 & 300 & 100 \\
Livermore & 22 & 1--2 & 20--\(1{,}000\) & 20--\(1{,}000\) \\
\bottomrule
\end{longtable}

\endgroup
\newpage
\section{Qualitative Recovery Across Taylor Orders}
\label{subsec:taylor_qualitative}

These experiments examine how Taylor approximation order affects recovery of the underlying analytic structure. Here, \(K\) is the truncation order used to generate the observations. For each base function, we construct four datasets from Taylor polynomials truncated at \(K\in\{2,5,7,10\}\) about the benchmark-specific expansion point and evaluate every method independently on each dataset. A prediction labeled \(K\) is therefore recovered from data generated at that order. Each table lists the data generating polynomials followed by the corresponding predictions. Ellipses abbreviate long expressions.

\begingroup
\scriptsize
\setlength{\tabcolsep}{3pt}
\setlength{\doublerulesep}{0.4pt}
\renewcommand{\arraystretch}{1.07}

\paragraph{Logarithm.}
Ground truth: \(\displaystyle \log{(x_{0} \allowbreak+ 1 )}\).

\begin{longtable}{@{}>{\raggedright\arraybackslash}p{0.16\textwidth}>{\centering\arraybackslash}p{0.06\textwidth}>{\raggedright\arraybackslash}p{0.74\textwidth}@{}}
\toprule
\midrule
\textbf{Taylor data} & 2 & \(\displaystyle x_{0} \allowbreak+ \frac{(-1) \allowbreak\cdot x_{0}^{2}}{2}\) \\
 & 5 & \(\displaystyle \frac{x_{0}^{5}}{5} \allowbreak- \frac{x_{0}^{4}}{4} \allowbreak+ \frac{x_{0}^{3}}{3} \allowbreak- \frac{x_{0}^{2}}{2} \allowbreak+ x_{0}\) \\
 & 7 & \(\displaystyle \frac{x_{0}^{7}}{7} \allowbreak- \frac{x_{0}^{6}}{6} \allowbreak+ \frac{x_{0}^{5}}{5} \allowbreak- \frac{x_{0}^{4}}{4} \allowbreak+ \frac{x_{0}^{3}}{3} \allowbreak- \frac{x_{0}^{2}}{2} \allowbreak+ x_{0}\) \\
 & 10 & \(\displaystyle \frac{x_{0}^{9}}{9} \allowbreak- \frac{x_{0}^{8}}{8} \allowbreak+ \frac{x_{0}^{7}}{7} \allowbreak- \frac{x_{0}^{6}}{6} \allowbreak+ \frac{x_{0}^{5}}{5} \allowbreak- \frac{x_{0}^{4}}{4} \allowbreak+ \frac{x_{0}^{3}}{3} \allowbreak- \frac{x_{0}^{2}}{2} \allowbreak+ x_{0} \allowbreak+ \frac{(-1) \allowbreak\cdot x_{0}^{10}}{10}\) \\
\midrule
\midrule
\textbf{MOSAIC-SR} & 2 & \(\displaystyle - 0.5 \allowbreak\cdot x_{0}^{2} \allowbreak+ x_{0}\) \\
 & 5 & \(\displaystyle \frac{x_{0} \allowbreak\cdot (0.6 \allowbreak\cdot x_{0}^{4} \allowbreak- 0.75 \allowbreak\cdot x_{0}^{3} \allowbreak+ x_{0}^{2} \allowbreak- 1.5 \allowbreak\cdot x_{0} \allowbreak+ 3.0)}{3}\) \\
 & 7 & \(\displaystyle \log{(x_{0} \allowbreak+ 1.00006603079797 )}\) \\
 & 10 & \(\displaystyle \log{(0.999982548715611 \allowbreak\cdot x_{0} \allowbreak+ 1.0 )}\) \\
\midrule
PySR & 2 & \(\displaystyle x_{0} \allowbreak\cdot (-0.50000006) \allowbreak\cdot x_{0} \allowbreak+ x_{0}\) \\
 & 5 & \(\displaystyle \log{(x_{0} \allowbreak+ e^{0.16825567 \allowbreak\cdot x_{0}^{4} \allowbreak\cdot (0.1677055 \allowbreak\cdot x_{0}^{2} \allowbreak+ x_{0}) \allowbreak\cdot \sin{(x_{0} )}} )}\) \\
 & 7 & \(\displaystyle \log{(\frac{0.9312916 \allowbreak\cdot x_{0}^{2}}{(8495.134 \allowbreak\cdot x_{0} \allowbreak+ 4841.4176432432) \allowbreak\cdot \cos{(x_{0} \allowbreak+ \sin{(x_{0} )} \allowbreak+ 0.55235815 )}} \allowbreak+ x_{0} \allowbreak+ 0.9999974 )}\) \\
 & 10 & \(\displaystyle \frac{x_{0}}{\sqrt{x_{0} \allowbreak+ e^{0.08364746 \allowbreak\cdot \sin{(\sin{(x_{0} \allowbreak\cdot (- x_{0} \allowbreak\cdot (0.007794200238976 \allowbreak- 0.01248848 \allowbreak\cdot x_{0}) \allowbreak+ x_{0}) )} )}}}}\) \\
\midrule
TPSR & 2 & \(\displaystyle (0.001919221591558286 \allowbreak- 0.06896537602514179 \allowbreak\cdot (3.264972382184671 \allowbreak\cdot x_{0} \allowbreak+ 0.027804329418150822)) \allowbreak\cdot (\ldots \allowbreak- 4.45999942314764)\) \\
 & 5 & \(\displaystyle (\ldots \allowbreak+ 0.1477134780683493) \allowbreak\cdot (0.4994556780444385 \allowbreak\cdot (3.5857739179080634 \allowbreak\cdot x_{0} \allowbreak+ 0.016444471966369778) \allowbreak- 0.008310635538500689)\) \\
 & 7 & \(\displaystyle (0.0004599611012800237 \allowbreak- 0.007574319367220268 \allowbreak\cdot (3.675921308958977 \allowbreak\cdot x_{0} \allowbreak+ 0.060970365116553515)) \allowbreak\cdot (\ldots \allowbreak- 9.800723047144695)\) \\
 & 10 & \(\displaystyle (0.00048668399774344864 \allowbreak- 0.008860476303170692 \allowbreak\cdot (3.547005732829724 \allowbreak\cdot x_{0} \allowbreak+ 0.05530797638861583)) \allowbreak\cdot (\ldots \allowbreak- 8.998199250504804)\) \\
\midrule
E2E-SR & 2 & \(\displaystyle \left(-6.33 \allowbreak+ \frac{7.87}{\ldots}\right) \allowbreak\cdot (\ldots)\) \\
 & 5 & \(\displaystyle (-9.9 \allowbreak- \frac{3.0}{0.055220918335784178 \allowbreak\cdot x_{0} \allowbreak+ 0.0842532448682821}) \allowbreak\cdot (- 0.021873220899239185 \allowbreak\cdot x_{0} \allowbreak- 3.1127899485564804 \allowbreak\cdot 10^{-7})\) \\
 & 7 & \(\displaystyle (-0.00015 \allowbreak+ \frac{0.0565}{3.675921308958977 \allowbreak\cdot x_{0} \allowbreak+ 6.6709703651165538}) \allowbreak\cdot (\ldots \allowbreak+ (0.16798960381942527 \allowbreak\cdot x_{0} \allowbreak+ 6.0327863456858267))\) \\
 & 10 & \(\displaystyle (0.00791 \allowbreak- \frac{0.396}{- 5.536875948947199 \allowbreak\cdot x_{0} \allowbreak- 9.1173357511426303}) \allowbreak\cdot (19.54400158789178 \allowbreak\cdot x_{0} \allowbreak+ 0.00174694990127322)\) \\
\midrule
Operon & 2 & \(\displaystyle 1.0 \allowbreak\cdot ((1.6375 \allowbreak- 2.1795 \allowbreak\cdot x_{0}) \allowbreak\cdot (0.2294 \allowbreak\cdot x_{0} \allowbreak- 0.2865) \allowbreak+ 0.4691) \allowbreak+ 0.0\) \\
 & 5 & \(\displaystyle 1.0 \allowbreak\cdot (- (2.7168 \allowbreak- 2.0625 \allowbreak\cdot x_{0}) \allowbreak- (\ldots \allowbreak+ 1.3181) \allowbreak+ e^{\cos{(0.2009 \allowbreak\cdot x_{0} )}} \allowbreak+ \sin{(- 0.6926 \allowbreak\cdot x_{0} )}) \allowbreak+ 0.0\) \\
 & 7 & \(\displaystyle 1.0 \allowbreak\cdot (\cos{(- 0.3504 \allowbreak\cdot x_{0} )} \allowbreak\cdot 0.3038 \allowbreak\cdot x_{0} \allowbreak- 1.3606 \allowbreak\cdot x_{0} \allowbreak\cdot (-0.4265) \allowbreak\cdot \frac{1}{0.5983 \allowbreak\cdot x_{0} \allowbreak+ 0.8338}) \allowbreak+ 0.0\) \\
 & 10 & \(\displaystyle 1.0001 \allowbreak\cdot \log{(0.9999 \allowbreak\cdot x_{0} \allowbreak+ e^{\sin{(\cos{(1.5708 )} )}} )} \allowbreak+ 0.0\) \\
\bottomrule
\end{longtable}
\newpage
\paragraph{Sine.}
Ground truth: \(\displaystyle \sin{(x_{0} )}\).

\begin{longtable}{@{}>{\raggedright\arraybackslash}p{0.16\textwidth}>{\centering\arraybackslash}p{0.06\textwidth}>{\raggedright\arraybackslash}p{0.74\textwidth}@{}}
\toprule
\midrule
\textbf{Taylor data} & 2 & \(\displaystyle x_{0}\) \\
 & 5 & \(\displaystyle \frac{x_{0}^{5}}{120} \allowbreak- \frac{x_{0}^{3}}{6} \allowbreak+ x_{0}\) \\
 & 7 & \(\displaystyle \frac{x_{0}^{5}}{120} \allowbreak- \frac{x_{0}^{3}}{6} \allowbreak+ x_{0} \allowbreak+ \frac{(-1) \allowbreak\cdot x_{0}^{7}}{5040}\) \\
 & 10 & \(\displaystyle \frac{x_{0}^{9}}{362880} \allowbreak- \frac{x_{0}^{7}}{5040} \allowbreak+ \frac{x_{0}^{5}}{120} \allowbreak- \frac{x_{0}^{3}}{6} \allowbreak+ x_{0}\) \\
\midrule
\midrule
\textbf{MOSAIC-SR} & 2 & \(\displaystyle x_{0}\) \\
 & 5 & \(\displaystyle - 0.15 \allowbreak\cdot x_{0}^{3} \allowbreak- 0.00833333333333333 \allowbreak\cdot x_{0} \allowbreak\cdot (- (x_{0}^{2} \allowbreak- 1)^{2} \allowbreak- 0.949359216081706) \allowbreak+ 0.983755339865986 \allowbreak\cdot x_{0}\) \\
 & 7 & \(\displaystyle 0.995212385036962 \allowbreak\cdot \sin{(x_{0} )}\) \\
 & 10 & \(\displaystyle \sin{(0.999224487053661 \allowbreak\cdot x_{0} )}\) \\
\midrule
PySR & 2 & \(\displaystyle x_{0}\) \\
 & 5 & \(\displaystyle \sin{(x_{0} \allowbreak\cdot (\cos{(\sin{(x_{0} \allowbreak\cdot x_{0} \allowbreak\cdot 0.099885456 \allowbreak\cdot \cos{(\sin{(x_{0} \allowbreak\cdot (-0.6655867) )} )} )} )} \allowbreak- 1 \allowbreak\cdot 0.0023687228) )} \allowbreak\cdot 1.0040035\) \\
 & 7 & \(\displaystyle \sin{(\frac{x_{0}}{\cos{(x_{0} \allowbreak\cdot (-0.03543025) \allowbreak\cdot (- 0.61757547 \allowbreak\cdot x_{0} \allowbreak+ \sin{(\sin{(\sin{(\sin{(\sin{(\sin{(\sin{(\sin{(\sin{(x_{0} )} )} )} )} )} )} )} )} )}) )}} )}\) \\
 & 10 & \(\displaystyle \sin{(x_{0} \allowbreak\cdot \cos{(x_{0} \allowbreak\cdot 0.0016573971 \allowbreak\cdot \frac{1}{(-1) \allowbreak\cdot 2.5042472 \allowbreak\cdot \frac{1}{- x_{0} \allowbreak\cdot x_{0} \allowbreak\cdot x_{0} \allowbreak+ \sin{((x_{0} \allowbreak+ x_{0}) \allowbreak\cdot (-2.066677) )}}} )} )}\) \\
\midrule
TPSR & 2 & \(\displaystyle 1.6952640543497626 \allowbreak\cdot (0.5898786076624276 \allowbreak\cdot x_{0} \allowbreak- 0.1503609060667909) \allowbreak+ 0.2549014392344917\) \\
 & 5 & \(\displaystyle - (\ldots \allowbreak+ 0.008829876210206134) \allowbreak+ 0.5311733478647308 \allowbreak\cdot (0.5442460160982873 \allowbreak\cdot x_{0} \allowbreak+ 0.03327453788862279) \allowbreak- 0.008829876210206504\) \\
 & 7 & \(\displaystyle 0.0037014237986746907 \allowbreak- 1.1106297415087139 \allowbreak\cdot \sin{(6.201124685239097 \allowbreak- 1.7856942534231586 \allowbreak\cdot (\ldots \allowbreak- 0.04590158661878021) )}\) \\
 & 10 & \(\displaystyle 0.9998616696023864 \allowbreak\cdot \sin{(1.7327776217284636 \allowbreak\cdot (\ldots \allowbreak- 0.06370211051208034) \allowbreak+ 0.11012377319587009 )} \allowbreak+ 6.858283367093453 \allowbreak\cdot 10^{-5}\) \\
\midrule
E2E-SR & 2 & \(\displaystyle (5.0538457150294822 \allowbreak\cdot 10^{-5} \allowbreak- 0.017224455343742888 \allowbreak\cdot x_{0}) \allowbreak\cdot (- 0.0060226605842333864 \allowbreak\cdot x_{0} \allowbreak- 58.016864815149059)\) \\
 & 5 & \(\displaystyle 0.00832 \allowbreak- 0.9720000000000001 \allowbreak\cdot \sin{(0.02411009851315413 \allowbreak\cdot x_{0} \allowbreak+ 0.043594062028465994 \allowbreak- \frac{0.807}{\ldots \allowbreak- 9.38 \allowbreak+ 0.0027099999999999997} )}\) \\
 & 7 & \(\displaystyle 1.0 \allowbreak\cdot \sin{(1.00998324807091 \allowbreak\cdot x_{0} \allowbreak- 0.00077925815314189615 )} \allowbreak+ 0.0021\) \\
 & 10 & \(\displaystyle 1.0 \allowbreak\cdot \sin{(0.9856914566698666 \allowbreak\cdot x_{0} \allowbreak+ 0.0016099999999999999 \allowbreak\cdot \arctan{(\ldots)} \allowbreak- 6.2710054014902136 )} \allowbreak- 0.0\) \\
\midrule
Operon & 2 & \(\displaystyle 1.0 \allowbreak\cdot 1.0 \allowbreak\cdot x_{0} \allowbreak+ 0.0\) \\
 & 5 & \(\displaystyle 1.001 \allowbreak\cdot (\cos{(- 1.0037 \allowbreak\cdot x_{0} )} \allowbreak\cdot 1.2879 \allowbreak\cdot x_{0} \allowbreak- \sin{(- (\ldots \allowbreak- 1.4461 \allowbreak\cdot x_{0}) \allowbreak- 0.1272 \allowbreak\cdot 0.0001 )}) \allowbreak+ 0.0\) \\
 & 7 & \(\displaystyle 1.0 \allowbreak\cdot (\cos{(- 0.4227 \allowbreak\cdot x_{0} )} \allowbreak\cdot 1.2701 \allowbreak\cdot x_{0} \allowbreak- (\ldots \allowbreak+ 0.0 \allowbreak\cdot x_{0} \allowbreak- 0.2272) \allowbreak\cdot \sin{(1.5034 \allowbreak\cdot x_{0} )}) \allowbreak+ 0.0\) \\
 & 10 & \(\displaystyle 1.001 \allowbreak\cdot (\cos{(0.5492 \allowbreak\cdot x_{0} )} \allowbreak\cdot (-0.2811) \allowbreak\cdot x_{0} \allowbreak- \sin{(- 0.9814 \allowbreak\cdot x_{0} )} \allowbreak+ \cos{(\ldots \allowbreak+ 0.626 \allowbreak\cdot x_{0} )}) \allowbreak+ 0.0\) \\
\bottomrule
\end{longtable}

\paragraph{Feynman I.18.12.}
Ground truth: \(\displaystyle x_{0} \allowbreak\cdot x_{1} \allowbreak\cdot \sin{(x_{2} )}\).

\begin{longtable}{@{}>{\raggedright\arraybackslash}p{0.16\textwidth}>{\centering\arraybackslash}p{0.06\textwidth}>{\raggedright\arraybackslash}p{0.74\textwidth}@{}}
\toprule
\midrule
\textbf{Taylor data} & 2 & \(\displaystyle - x_{0} \allowbreak\cdot x_{1} \allowbreak\cdot x_{2} \allowbreak+ \pi \allowbreak\cdot x_{0} \allowbreak\cdot x_{1}\) \\
 & 5 & \(\displaystyle \ldots \allowbreak+ \frac{- x_{0} \allowbreak\cdot x_{1} \allowbreak\cdot x_{2}^{5}}{120} \allowbreak+ \frac{\pi \allowbreak\cdot x_{0} \allowbreak\cdot x_{1} \allowbreak\cdot x_{2}^{4}}{24} \allowbreak- \frac{\pi^{2} \allowbreak\cdot x_{0} \allowbreak\cdot x_{1} \allowbreak\cdot x_{2}^{3}}{12} \allowbreak+ \frac{x_{0} \allowbreak\cdot x_{1} \allowbreak\cdot x_{2}^{3}}{6} \allowbreak- \frac{\pi \allowbreak\cdot x_{0} \allowbreak\cdot x_{1} \allowbreak\cdot x_{2}^{2}}{2} \allowbreak+ \frac{\pi^{3} \allowbreak\cdot x_{0} \allowbreak\cdot x_{1} \allowbreak\cdot x_{2}^{2}}{12}\) \\
 & 7 & \(\displaystyle \ldots \allowbreak+ \frac{x_{0} \allowbreak\cdot x_{1} \allowbreak\cdot x_{2}^{7}}{5040} \allowbreak- \frac{\pi \allowbreak\cdot x_{0} \allowbreak\cdot x_{1} \allowbreak\cdot x_{2}^{6}}{720} \allowbreak- \frac{x_{0} \allowbreak\cdot x_{1} \allowbreak\cdot x_{2}^{5}}{120} \allowbreak+ \frac{\pi^{2} \allowbreak\cdot x_{0} \allowbreak\cdot x_{1} \allowbreak\cdot x_{2}^{5}}{240} \allowbreak- \frac{\pi^{3} \allowbreak\cdot x_{0} \allowbreak\cdot x_{1} \allowbreak\cdot x_{2}^{4}}{144} \allowbreak+ \pi \allowbreak\cdot x_{0} \allowbreak\cdot x_{1} \allowbreak\cdot x_{2}^{4}\) \\
 & 10 & \(\displaystyle \ldots \allowbreak+ \frac{- x_{0} \allowbreak\cdot x_{1} \allowbreak\cdot x_{2}^{9}}{362880} \allowbreak+ \frac{\pi \allowbreak\cdot x_{0} \allowbreak\cdot x_{1} \allowbreak\cdot x_{2}^{8}}{40320} \allowbreak- \frac{\pi^{2} \allowbreak\cdot x_{0} \allowbreak\cdot x_{1} \allowbreak\cdot x_{2}^{7}}{10080} \allowbreak+ \frac{x_{0} \allowbreak\cdot x_{1} \allowbreak\cdot x_{2}^{7}}{5040} \allowbreak- \frac{\pi \allowbreak\cdot x_{0} \allowbreak\cdot x_{1} \allowbreak\cdot x_{2}^{6}}{720} \allowbreak+ \pi^{3} \allowbreak\cdot x_{0} \allowbreak\cdot x_{1}\) \\
\midrule
\midrule
\textbf{MOSAIC-SR} & 2 & \(\displaystyle x_{0} \allowbreak\cdot x_{1} \allowbreak\cdot (\pi \allowbreak- x_{2})\) \\
 & 5 & \(\displaystyle x_{0} \allowbreak\cdot x_{1} \allowbreak\cdot \sin{(x_{2} )}\) \\
 & 7 & \(\displaystyle x_{0} \allowbreak\cdot x_{1} \allowbreak\cdot \sin{(1.002337079977 \allowbreak\cdot x_{2} )}\) \\
 & 10 & \(\displaystyle 0.564336516744925 \allowbreak\cdot \sqrt{\pi} \allowbreak\cdot x_{0} \allowbreak\cdot x_{1} \allowbreak\cdot \sin{(x_{2} )}\) \\
\midrule
PySR & 2 & \(\displaystyle x_{0} \allowbreak\cdot x_{1} \allowbreak\cdot (3.1415925 \allowbreak- x_{2})\) \\
 & 5 & \(\displaystyle \frac{x_{0} \allowbreak\cdot (0.33221138 \allowbreak\cdot x_{1} \allowbreak\cdot \sin{(\cos{(x_{2} )} )} \allowbreak+ x_{1}) \allowbreak\cdot \sin{(\sin{(x_{2} \allowbreak\cdot 0.4749027 \allowbreak+ 1.6496296 )} \allowbreak\cdot 2.7744522 )}}{0.9437794}\) \\
 & 7 & \(\displaystyle x_{0} \allowbreak\cdot x_{1} \allowbreak\cdot \sin{(x_{2} \allowbreak- - 0.019365529 \allowbreak\cdot (x_{2} \allowbreak- 3.146233) \allowbreak\cdot \cos{(\sin{(\sin{(\sin{(x_{2} )} \allowbreak\cdot (-1.9357585) )} )} \allowbreak\cdot (-1.7473638) )} )}\) \\
 & 10 & \(\displaystyle x_{0} \allowbreak\cdot (x_{1} \allowbreak\cdot e^{\cos{(x_{2} )}} \allowbreak\cdot (-0.0030647963) \allowbreak\cdot \cos{(x_{2} )} \allowbreak+ x_{1}) \allowbreak\cdot \sin{(x_{2} \allowbreak\cdot 0.9981049 \allowbreak- 1 \allowbreak\cdot (-0.0058534783) )}\) \\
\midrule
TPSR & 2 & \(\displaystyle (\ldots \allowbreak+ 0.28331615049921344 \allowbreak+ 0.4956534371075346) \allowbreak\cdot (0.34386125723713024 \allowbreak\cdot x_{0} \allowbreak- 1.8420104004975268 \allowbreak+ 1.8420104004834439)\) \\
 & 5 & \(\displaystyle 1.276389115335405 \allowbreak+ 1 \allowbreak\cdot \frac{1}{\ldots \allowbreak- 0.07925775134645766 \allowbreak+ 0.2495263264800342 \allowbreak+ 0.2580259436964331 \allowbreak+ 0.2746491984632519}\) \\
 & 7 & \(\displaystyle \sin{(- 2.215140796977261 \allowbreak\cdot |{\ldots \allowbreak+ 0.5758522222337376 \allowbreak+ 1.7552518942051996}| \allowbreak- 0.8214622877077248 )} \allowbreak+ 0.610334905734776\) \\
 & 10 & \(\displaystyle (\ldots \allowbreak+ 0.0850328212232805 \allowbreak+ 0.1470358948108155) \allowbreak\cdot (0.36255539574241513 \allowbreak\cdot x_{0} \allowbreak- 1.8127229580805884 \allowbreak+ 1.812829094663855)\) \\
\midrule
E2E-SR & 2 & \(\displaystyle (- 0.002704484223201092 \allowbreak\cdot x_{1} \allowbreak- 0.60237161959079474) \allowbreak\cdot (\ldots \allowbreak+ 0.0011760054997509854 \allowbreak\cdot x_{0} \allowbreak- 0.0427947554976017 \allowbreak\cdot x_{2} \allowbreak+ 0.13465458451346508)\) \\
 & 5 & \(\displaystyle (0.76120861641426228 \allowbreak\cdot x_{0} \allowbreak+ 0.022982771994114469) \allowbreak\cdot (0.37355026941984976 \allowbreak\cdot x_{1} \allowbreak+ 0.17673765134826056) \allowbreak\cdot (\ldots \allowbreak- 0.0016428320073234725 \allowbreak\cdot x_{0} \allowbreak- 3.12)\) \\
 & 7 & \(\displaystyle (169.62829714499832 \allowbreak- 54.02295102376346 \allowbreak\cdot x_{2}) \allowbreak\cdot (\ldots \allowbreak+ 1.8694742958078208 \allowbreak\cdot x_{1}) \allowbreak\cdot (0.001513002143198883 \allowbreak\cdot x_{0} \allowbreak+ 4.3499570545021802 \allowbreak\cdot 10^{-5})\) \\
 & 10 & \(\displaystyle (0.0018476647025342624 \allowbreak\cdot x_{0} \allowbreak+ 7.4163269681210592 \allowbreak\cdot 10^{-5}) \allowbreak\cdot (- 18.721308938254057 \allowbreak\cdot x_{1} \allowbreak- 3.1062206918614961) \allowbreak\cdot 28.2 \allowbreak\cdot \sin{(\ldots \allowbreak+ 0.9922408707994587 )}\) \\
\midrule
Operon & 2 & \(\displaystyle 1.0 \allowbreak\cdot 12.6214 \allowbreak\cdot x_{0} \allowbreak\cdot (83.0285 \allowbreak- 26.4288 \allowbreak\cdot x_{2}) \allowbreak\cdot (0.0 \allowbreak\cdot x_{0} \allowbreak+ 0.003 \allowbreak\cdot x_{1}) \allowbreak+ 0.0\) \\
 & 5 & \(\displaystyle 1.0 \allowbreak\cdot (- 0.9508 \allowbreak\cdot x_{2} \allowbreak- -2.9873) \allowbreak\cdot (\ldots \allowbreak- 97.4074 \allowbreak\cdot x_{0} \allowbreak- 76.7503 \allowbreak\cdot x_{0}) \allowbreak- 0.0011\) \\
 & 7 & \(\displaystyle 1.0 \allowbreak\cdot (- 0.1636 \allowbreak\cdot x_{0} \allowbreak\cdot \sin{(- 0.539 \allowbreak\cdot x_{2} \allowbreak- 1.4483 )} \allowbreak- 0.7675 \allowbreak\cdot x_{2} \allowbreak- (\ldots \allowbreak- 2.4113)) \allowbreak- 0.0007\) \\
 & 10 & \(\displaystyle 1.0 \allowbreak\cdot (0.1933 \allowbreak\cdot x_{0} \allowbreak\cdot 5.0729 \allowbreak\cdot x_{1} \allowbreak\cdot \sin{(1.0105 \allowbreak\cdot x_{2} )} \allowbreak- 0.006 \allowbreak\cdot x_{0} \allowbreak\cdot (\ldots \allowbreak+ 5.3587 \allowbreak\cdot x_{1})) \allowbreak+ 0.0005\) \\
\bottomrule
\end{longtable}

\paragraph{Feynman I.12.11.}
Ground truth: \(\displaystyle x_{0} \allowbreak\cdot (x_{1} \allowbreak+ x_{2} \allowbreak\cdot x_{3} \allowbreak\cdot \sin{(x_{4} )})\).

\begin{longtable}{@{}>{\raggedright\arraybackslash}p{0.16\textwidth}>{\centering\arraybackslash}p{0.06\textwidth}>{\raggedright\arraybackslash}p{0.74\textwidth}@{}}
\toprule
\midrule
\textbf{Taylor data} & 2 & \(\displaystyle x_{0} \allowbreak\cdot x_{1} \allowbreak- \frac{\sqrt{2} \allowbreak\cdot x_{0} \allowbreak\cdot x_{2} \allowbreak\cdot x_{3} \allowbreak\cdot x_{4}^{2}}{4} \allowbreak+ \frac{\sqrt{2} \allowbreak\cdot \pi \allowbreak\cdot x_{0} \allowbreak\cdot x_{2} \allowbreak\cdot x_{3} \allowbreak\cdot x_{4}}{8} \allowbreak+ \frac{\sqrt{2} \allowbreak\cdot x_{0} \allowbreak\cdot x_{2} \allowbreak\cdot x_{3} \allowbreak\cdot x_{4}}{2} \allowbreak- \frac{\sqrt{2} \allowbreak\cdot \pi \allowbreak\cdot x_{0} \allowbreak\cdot x_{2} \allowbreak\cdot x_{3}}{8} \allowbreak- \frac{\sqrt{2} \allowbreak\cdot \pi^{2} \allowbreak\cdot x_{0} \allowbreak\cdot x_{2} \allowbreak\cdot x_{3}}{64} \allowbreak+ \frac{\sqrt{2} \allowbreak\cdot x_{0} \allowbreak\cdot x_{2} \allowbreak\cdot x_{3}}{2}\) \\
 & 5 & \(\displaystyle \ldots \allowbreak+ x_{0} \allowbreak\cdot x_{1} \allowbreak+ \frac{\sqrt{2} \allowbreak\cdot x_{0} \allowbreak\cdot x_{2} \allowbreak\cdot x_{3} \allowbreak\cdot x_{4}^{5}}{240} \allowbreak- \frac{\sqrt{2} \allowbreak\cdot \pi \allowbreak\cdot x_{0} \allowbreak\cdot x_{2} \allowbreak\cdot x_{3} \allowbreak\cdot x_{4}^{4}}{192} \allowbreak+ \frac{\sqrt{2} \allowbreak\cdot x_{0} \allowbreak\cdot x_{2} \allowbreak\cdot x_{3} \allowbreak\cdot x_{4}^{4}}{48} \allowbreak- \sqrt{2} \allowbreak\cdot x_{0} \allowbreak\cdot x_{2} \allowbreak\cdot x_{3} \allowbreak\cdot x_{4}\) \\
 & 7 & \(\displaystyle \ldots \allowbreak+ x_{0} \allowbreak\cdot x_{1} \allowbreak- \frac{\sqrt{2} \allowbreak\cdot x_{0} \allowbreak\cdot x_{2} \allowbreak\cdot x_{3} \allowbreak\cdot x_{4}^{7}}{10080} \allowbreak- \frac{\sqrt{2} \allowbreak\cdot x_{0} \allowbreak\cdot x_{2} \allowbreak\cdot x_{3} \allowbreak\cdot x_{4}^{6}}{1440} \allowbreak+ \frac{\sqrt{2} \allowbreak\cdot \pi \allowbreak\cdot x_{0} \allowbreak\cdot x_{2} \allowbreak\cdot x_{3} \allowbreak\cdot x_{4}^{6}}{5760} \allowbreak- \sqrt{2} \allowbreak\cdot \pi^{2} \allowbreak\cdot x_{0}\) \\
 & 10 & \(\displaystyle \ldots \allowbreak+ x_{0} \allowbreak\cdot x_{1} \allowbreak- \frac{\sqrt{2} \allowbreak\cdot x_{0} \allowbreak\cdot x_{2} \allowbreak\cdot x_{3} \allowbreak\cdot x_{4}^{10}}{7257600} \allowbreak+ \frac{\sqrt{2} \allowbreak\cdot \pi \allowbreak\cdot x_{0} \allowbreak\cdot x_{2} \allowbreak\cdot x_{3} \allowbreak\cdot x_{4}^{9}}{2903040} \allowbreak+ \frac{\sqrt{2} \allowbreak\cdot x_{0} \allowbreak\cdot x_{2} \allowbreak\cdot x_{3} \allowbreak\cdot x_{4}^{9}}{725760} \allowbreak- \sqrt{2} \allowbreak\cdot \pi \allowbreak\cdot x_{0}\) \\
\midrule
\midrule
\textbf{MOSAIC-SR} & 2 & \(\displaystyle x_{0} \allowbreak\cdot (x_{1} \allowbreak+ 0.712793490396519 \allowbreak\cdot x_{2} \allowbreak\cdot x_{3} \allowbreak\cdot x_{4} \allowbreak+ 0.0669510918992685 \allowbreak\cdot x_{2} \allowbreak\cdot x_{3})\) \\
 & 5 & \(\displaystyle x_{0} \allowbreak\cdot (x_{1} \allowbreak+ x_{2} \allowbreak\cdot x_{3} \allowbreak\cdot \sin{(x_{4} )})\) \\
 & 7 & \(\displaystyle x_{0} \allowbreak\cdot (x_{1} \allowbreak+ 1.0 \allowbreak\cdot x_{2} \allowbreak\cdot x_{3} \allowbreak\cdot \sin{(x_{4} )})\) \\
 & 10 & \(\displaystyle x_{0} \allowbreak\cdot (x_{1} \allowbreak+ x_{2} \allowbreak\cdot x_{3} \allowbreak\cdot \sin{(x_{4} )})\) \\
\midrule
PySR & 2 & \(\displaystyle x_{0} \allowbreak\cdot (x_{1} \allowbreak+ x_{2} \allowbreak\cdot x_{3} \allowbreak\cdot (x_{4} \allowbreak\cdot (1.262467 \allowbreak+ x_{4} \allowbreak\cdot (-0.35355335)) \allowbreak- 1 \allowbreak\cdot 0.06634307))\) \\
 & 5 & \(\displaystyle x_{0} \allowbreak\cdot (x_{1} \allowbreak+ x_{3} \allowbreak\cdot 0.99937403 \allowbreak\cdot x_{2} \allowbreak\cdot (0.0004353141 \allowbreak\cdot x_{4} \allowbreak+ \sin{(x_{4} )} \allowbreak+ 8.375039 \allowbreak\cdot 10^{-5}))\) \\
 & 7 & \(\displaystyle x_{0} \allowbreak\cdot (x_{1} \allowbreak+ x_{2} \allowbreak\cdot x_{3} \allowbreak\cdot (\sin{(x_{4} \allowbreak- \frac{5.59005 \allowbreak\cdot 10^{-8}}{e^{x_{4}} \allowbreak- 0.98920304} )} \allowbreak- \frac{6.089482 \allowbreak\cdot 10^{-8}}{\cos{(x_{4} )}}))\) \\
 & 10 & \(\displaystyle x_{0} \allowbreak\cdot (x_{1} \allowbreak+ x_{2} \allowbreak\cdot x_{3} \allowbreak\cdot \sin{(x_{4} )})\) \\
\midrule
TPSR & 2 & \(\displaystyle - (\ldots \allowbreak- 0.41703132683721883) \allowbreak+ 0.20866626498961413 \allowbreak\cdot (0.3604272691006456 \allowbreak\cdot x_{1} \allowbreak- 1.744712561451487) \allowbreak- 0.05296867316273429\) \\
 & 5 & \(\displaystyle \ldots \allowbreak+ 0.34294855858724976 \allowbreak\cdot x_{1} \allowbreak- 4.723488471036021 \allowbreak- 1.7988515848521762 \allowbreak+ 1.7989003805824426 \allowbreak+ 2.495998943387973\) \\
 & 7 & \(\displaystyle (\ldots \allowbreak+ 61.51639518156274) \allowbreak\cdot (0.5088734890611518 \allowbreak\cdot (0.17031588558061792 \allowbreak\cdot x_{0} \allowbreak- 0.14978961891848902) \allowbreak+ 0.07771255202217539)\) \\
 & 10 & \(\displaystyle (\ldots \allowbreak+ 9.407194864169705) \allowbreak\cdot (0.7224790948921128 \allowbreak\cdot (0.17230290182470367 \allowbreak\cdot x_{0} \allowbreak- 0.07264073340143025) \allowbreak+ 0.05248141132993334)\) \\
\midrule
E2E-SR & 2 & \(\displaystyle (0.1629358158362928 \allowbreak\cdot x_{0} \allowbreak- 0.0023126882037319991) \allowbreak\cdot (\ldots \allowbreak+ 6.16330630162104 \allowbreak\cdot x_{1} \allowbreak- (- 8.0523542289226607 \allowbreak\cdot x_{3} \allowbreak- 1.4868480523885724))\) \\
 & 5 & \(\displaystyle (- 0.31289792906740601 \allowbreak\cdot x_{0} \allowbreak- 0.0002607466038638947) \allowbreak\cdot (- 3.2888766768517251 \allowbreak\cdot x_{1} \allowbreak+ 0.3503022089874898 \allowbreak\cdot x_{2} \allowbreak- (\ldots \allowbreak- 0.06099736784503796 \allowbreak\cdot x_{4}))\) \\
 & 7 & \(\displaystyle (0.093673737069339864 \allowbreak\cdot x_{0} \allowbreak- 0.0023842904051689659) \allowbreak\cdot (\ldots \allowbreak+ 10.452156425909193 \allowbreak\cdot x_{1} \allowbreak- (2.038660868691718 \allowbreak- 0.5685205786851859 \allowbreak\cdot x_{4}))\) \\
 & 10 & \(\displaystyle (\ldots \allowbreak+ 0.0014473443753275107 \allowbreak\cdot x_{0}) \allowbreak\cdot (0.17230290182470367 \allowbreak\cdot x_{0} \allowbreak- 0.004740733401430247) \allowbreak\cdot (0.27324955376110477 \allowbreak\cdot x_{4} \allowbreak- 0.97426820283141966)\) \\
\midrule
Operon & 2 & \(\displaystyle 1.0 \allowbreak\cdot (\ldots \allowbreak- 2.3002 \allowbreak\cdot x_{0} \allowbreak\cdot 0.0 \allowbreak\cdot x_{3} \allowbreak+ 0.0102 \allowbreak\cdot x_{0} \allowbreak\cdot (97.7415 \allowbreak\cdot x_{1} \allowbreak+ 0.0 \allowbreak\cdot x_{2})) \allowbreak+ 0.0\) \\
 & 5 & \(\displaystyle 1.0 \allowbreak\cdot (2.3026 \allowbreak\cdot x_{1} \allowbreak\cdot 0.4342 \allowbreak\cdot x_{0} \allowbreak+ 0.0133 \allowbreak\cdot x_{0} \allowbreak\cdot 23.3392 \allowbreak\cdot x_{4} \allowbreak\cdot (\ldots \allowbreak+ 0.0236 \allowbreak\cdot x_{2})) \allowbreak- 0.0025\) \\
 & 7 & \(\displaystyle 1.0021 \allowbreak\cdot (4.6741 \allowbreak\cdot x_{0} \allowbreak- (0.6931 \allowbreak\cdot x_{4} \allowbreak\cdot (-0.0536) \allowbreak\cdot x_{4} \allowbreak\cdot 102.1899 \allowbreak\cdot x_{3} \allowbreak- (\ldots \allowbreak- 29.3384 \allowbreak\cdot x_{3}))) \allowbreak- 0.4812\) \\
 & 10 & \(\displaystyle 1.0 \allowbreak\cdot (-0.0658) \allowbreak\cdot x_{4} \allowbreak\cdot 1.1554 \allowbreak\cdot x_{3} \allowbreak\cdot (\ldots \allowbreak- 0.1744 \allowbreak\cdot x_{0} \allowbreak\cdot (-5.8297) \allowbreak\cdot x_{4}) \allowbreak- 0.0127\) \\
\bottomrule
\end{longtable}

\endgroup
\section{SRSD-Feynman Predictions}
\label{app:srsd_prediction_examples}

We list predictions for twelve selected SRSD-Feynman equations, with four structurally varied examples from each difficulty split.
% The first MOSAIC-SR row in each table reports the initial formula, and the second reports the search-recovered formula.
The MOSAIC-SR row in each table reports the search-recovered formula.
Ellipses indicate long baseline predictions abbreviated for readability.

\begingroup
\scriptsize
\setlength{\tabcolsep}{2pt}
\renewcommand{\arraystretch}{1.08}
\def\UrlFont{\ttfamily\scriptsize}

\subsubsection{Easy}

\paragraph{feynman-i.18.16.}
Ground truth: \(\displaystyle x_{0} \allowbreak\cdot x_{1} \allowbreak\cdot x_{2} \allowbreak\cdot \sin{(x_{3} )}\).

\begin{longtable}{@{}>{\raggedright\arraybackslash}p{0.14\textwidth}>{\raggedright\arraybackslash}p{0.82\textwidth}@{}}
\toprule
Method & Prediction \\
\midrule
\endfirsthead
\toprule
Method & Prediction \\
\midrule
\endhead
% \textbf{MOSAIC-SR}\newline\textit{Initial Formula} & \(x_{0} \allowbreak\cdot (x_{1} \allowbreak\cdot x_{2}) \allowbreak\cdot \sin(x_{3})\) \\
% \textit{Search Recovered}
\textbf{MOSAIC-SR} & \(x_{0} \allowbreak\cdot x_{1} \allowbreak\cdot x_{2} \allowbreak\cdot \sin(x_{3})\) \\
\midrule
PySR & \(x_{0} \allowbreak\cdot x_{1} \allowbreak\cdot x_{2} \allowbreak\cdot ((x_{2} \allowbreak- (x_{3} \allowbreak+ (x_{3} \allowbreak+ \sin{(x_{2} )} \allowbreak+ \cos{(x_{3} )}) \allowbreak\cdot \cos{(x_{1} )})) \allowbreak\cdot (-1.090031 \allowbreak\cdot 10^{-8}) \allowbreak+ \sin{(x_{3} )})\) \\
Operon & \(1.0 \allowbreak\cdot \sin{(1.0 \allowbreak\cdot x_{3} )} \allowbreak\cdot (-0.0027) \allowbreak\cdot x_{1} \allowbreak\cdot (0.0 \allowbreak\cdot x_{2} \allowbreak\cdot e^{\sin{(\cos{(- 2.0643 \allowbreak\cdot x_{3} )} )}} \allowbreak- 2.8718 \allowbreak\cdot x_{2}) \allowbreak\cdot (- 1.1801 \allowbreak\cdot \ldots \allowbreak\cdot x_{1} \allowbreak+ 127.7102 \allowbreak\cdot x_{0}) \allowbreak+ 0.0\) \\
GP-GOMEA & \(0.560146 \allowbreak\cdot x_{0} \allowbreak\cdot x_{1} \allowbreak\cdot x_{2} \allowbreak\cdot e^{\cos{(\sin{(45.244 )} )}} \allowbreak\cdot \sin{(x_{3} )} \allowbreak+ 0.0\) \\
E2E-SR & \((- 0.00047451032853731212 \allowbreak\cdot x_{1} \allowbreak- 19.400000000000002 \allowbreak\cdot (\ldots \allowbreak- 0.0034) \allowbreak\cdot (30.38912823447775 \allowbreak\cdot x_{0} \allowbreak+ 28.70121528936986)^{2}) \allowbreak\cdot (\sin{(0.5475682868652938 \allowbreak\cdot x_{3} \allowbreak- 1.7094145799484706 )} \allowbreak- 0.0010999999999999998)\) \\
SNIP-SR & \((2.9242714060991837 \allowbreak\cdot x_{0} \allowbreak+ 2.8138086776705413 \allowbreak\cdot x_{2}) \allowbreak\cdot (0.058021734319863794 \allowbreak\cdot x_{1} \allowbreak- (-0.11031702200615587) \allowbreak\cdot 1 \allowbreak\cdot \frac{1}{\ldots \allowbreak- 1.980797909 \allowbreak- 0.6172368400935637}) \allowbreak- 0.3353196824463989\) \\
\bottomrule
\end{longtable}

\paragraph{feynman-i.47.23.}
Ground truth: \(\displaystyle \sqrt{\frac{x_{0} \allowbreak\cdot x_{1}}{x_{2}}}\).

\begin{longtable}{@{}>{\raggedright\arraybackslash}p{0.14\textwidth}>{\raggedright\arraybackslash}p{0.82\textwidth}@{}}
\toprule
Method & Prediction \\
\midrule
\endfirsthead
\toprule
Method & Prediction \\
\midrule
\endhead
% \textbf{MOSAIC-SR}\newline\textit{Initial Formula} & \(\frac{\exp\!\left(1.678 \allowbreak\cdot \exp(-3.95 \allowbreak\cdot 10^{4} \allowbreak\cdot x_{1})\right) \allowbreak\cdot x_{0} \allowbreak}{x_{2}}\) \\
% \textit{Search Recovered}
\textbf{MOSAIC-SR} & \(\sqrt{\frac{x_{0} \allowbreak\cdot x_{1}}{x_{2}}}\) \\
\midrule
PySR & \(\sqrt{\frac{x_{0} \allowbreak\cdot x_{1}}{x_{2}}}\) \\
Operon & \(0.0 \allowbreak+ 1.0 \allowbreak\cdot (- 0.3697 \allowbreak\cdot x_{0} \allowbreak+ 37267.543 \allowbreak\cdot x_{1} \allowbreak- 0.3693 \allowbreak\cdot x_{2} \allowbreak+ \log{(- (-1) \allowbreak\cdot 405434.875 \allowbreak\cdot x_{1} \allowbreak+ 0.0036 \allowbreak\cdot x_{2} \allowbreak+ 1.3023 )} \allowbreak+ 0.2423) \allowbreak\cdot (-1.5741) \allowbreak\cdot \frac{1}{\ldots \allowbreak- 1.316}\) \\
GP-GOMEA & \(0.088311 \allowbreak\cdot \sqrt{x_{0} \allowbreak\cdot x_{1}} \allowbreak\cdot (\frac{x_{2}}{451.691759865387} \allowbreak+ \sin{(x_{2} )} \allowbreak- 3.832) \allowbreak\cdot (- x_{1} \allowbreak+ x_{2} \allowbreak- 8.635 \allowbreak+ 3.845) \allowbreak- 2.6 \allowbreak\cdot 10^{-5}\) \\
E2E-SR & \((0.3 \allowbreak+ \frac{0.3}{0.35927349724435038 \allowbreak\cdot x_{2} \allowbreak+ 0.1556208603753693}) \allowbreak\cdot (\ldots \allowbreak+ 0.00019999999999999998 \allowbreak\cdot \sqrt{(77.563144512331082 \allowbreak\cdot x_{0} \allowbreak- 25.889254950623197) \allowbreak\cdot (369968.1560669523 \allowbreak\cdot x_{1} \allowbreak- 0.683865090989673)})\) \\
SNIP-SR & \(0.0008867575457021422 \allowbreak\cdot x_{0} \allowbreak- (0.007603756123524418 \allowbreak\cdot x_{1} \allowbreak- 0.013302360696511688 \allowbreak\cdot x_{2} \allowbreak- (\ldots \allowbreak- 0.0017) \allowbreak+ 0.0617609181319026 \allowbreak+ 0.07176091798886611) \allowbreak+ 0.0042390819407224665\) \\
\bottomrule
\end{longtable}

\paragraph{feynman-iii.15.27.}
Ground truth: \(\displaystyle \frac{2 \allowbreak\cdot \pi \allowbreak\cdot x_{0}}{x_{1} \allowbreak\cdot x_{2}}\).

\begin{longtable}{@{}>{\raggedright\arraybackslash}p{0.14\textwidth}>{\raggedright\arraybackslash}p{0.82\textwidth}@{}}
\toprule
Method & Prediction \\
\midrule
\endfirsthead
\toprule
Method & Prediction \\
\midrule
\endhead
% \textbf{MOSAIC-SR}\newline\textit{Initial Formula} & \(\exp(6.565 \allowbreak\cdot 10^{13} \allowbreak\cdot \pi^{2}) \allowbreak\cdot \frac{\left(x_{0} \allowbreak+ x_{2} \allowbreak\cdot \exp(\pi)\right)}{x_{1} \allowbreak\cdot x_{2}}\) \\
% \textit{Search Recovered}
\textbf{MOSAIC-SR} & \(\frac{2 \allowbreak\cdot \pi \allowbreak\cdot x_{0}}{x_{1} \allowbreak\cdot x_{2}}\) \\
\midrule
PySR & \(x_{0} \allowbreak\cdot 6.2831855 \allowbreak\cdot \frac{1}{x_{1} \allowbreak\cdot x_{2}}\) \\
Operon & \(1.0 \allowbreak\cdot (-1.72) \allowbreak\cdot x_{0} \allowbreak\cdot \frac{1}{(-1) \allowbreak\cdot 0.5005 \allowbreak\cdot x_{1}} \allowbreak\cdot \frac{1}{0.5469 \allowbreak\cdot x_{2}} \allowbreak+ 354.8659\) \\
GP-GOMEA & \(10326321737.098848 \allowbreak\cdot (x_{0} \allowbreak+ x_{0} \allowbreak- (x_{0} \allowbreak+ x_{1}) \allowbreak\cdot \sin{(x_{0} )} \allowbreak- (- x_{1} \allowbreak+ x_{2})) \allowbreak\cdot e^{- \sqrt{x_{1}} \allowbreak+ \cos{(x_{1} )}} \allowbreak- 715320458.443097\) \\
E2E-SR & \(- 0.00046093254802904855 \allowbreak\cdot x_{0} \allowbreak+ (23.8 \allowbreak- 0.03485279359938367 \allowbreak\cdot e^{0.03610003915333401 \allowbreak\cdot x_{1}}) \allowbreak\cdot (0.016461876715323162 \allowbreak\cdot x_{0} \allowbreak+ 0.004137440901804627) \allowbreak\cdot (\ldots \allowbreak- 3165117.571990305 \allowbreak\cdot x_{2} \allowbreak- 0.03078025829268)\) \\
SNIP-SR & \(x_{0} \allowbreak\cdot e^{17.656406591853308 \allowbreak- 0.29635039365231586 \allowbreak\cdot x_{1}} \allowbreak\cdot 1 \allowbreak\cdot \frac{1}{5611389.50330818 \allowbreak\cdot x_{2} \allowbreak+ 6.00837704285586 \allowbreak\cdot 10^{-5}} \allowbreak- 607824857.6181431\) \\
\bottomrule
\end{longtable}

\paragraph{feynman-ii.2.42.}
Ground truth: \(\displaystyle \frac{x_{0} \allowbreak\cdot x_{3} \allowbreak\cdot (x_{1} \allowbreak- x_{2})}{x_{4}}\).

\begin{longtable}{@{}>{\raggedright\arraybackslash}p{0.14\textwidth}>{\raggedright\arraybackslash}p{0.82\textwidth}@{}}
\toprule
Method & Prediction \\
\midrule
\endfirsthead
\toprule
Method & Prediction \\
\midrule
\endhead
% \textbf{MOSAIC-SR}\newline\textit{Initial Formula} & \(\log(x_{0}) \allowbreak\cdot \frac{x_{1} \allowbreak\cdot (-x_{3})}{-1 \allowbreak+ x_{1}^{2} \allowbreak- x_{2} \allowbreak- 4.506 \allowbreak\cdot 10^{5} \allowbreak\cdot x_{1}}\) \\
% \textit{Search Recovered}
\textbf{MOSAIC-SR} & \(\frac{x_{0} \allowbreak\cdot x_{3} \allowbreak\cdot (x_{1} \allowbreak- x_{2})}{x_{4}}\) \\
\midrule
PySR & \(\displaystyle \frac{-1.1901392 \allowbreak\cdot 0.84023786 \allowbreak\cdot x_{0} \allowbreak\cdot x_{3} \allowbreak\cdot (-x_{1} \allowbreak+ x_{2})}{x_{4}}\) \\
Operon & \(0.0 \allowbreak+ 1.0 \allowbreak\cdot (0.0 \allowbreak\cdot x_{1} \allowbreak+ \frac{(-0.0018 \allowbreak\cdot x_{1} \allowbreak+ 0.0018 \allowbreak\cdot x_{2}) \allowbreak\cdot (-1.1077) \allowbreak\cdot x_{0} \allowbreak\cdot (30998.9414 \allowbreak\cdot x_{3} \allowbreak+ \ldots)}{63.3968 \allowbreak\cdot x_{4}})\) \\
GP-GOMEA & \(-0.338812 \allowbreak- 0.204571 \allowbreak\cdot \sqrt{x_{3}} \allowbreak\cdot x_{0}^{2} \allowbreak\cdot (x_{4} \allowbreak\cdot x_{1} \allowbreak- (x_{1} \allowbreak- x_{2})) \allowbreak\cdot e^{\sin{x_{2}} \allowbreak- x_{4} \allowbreak\cdot x_{0}}\) \\
E2E-SR & \((\ldots \allowbreak+ 0.006059999999999999) \allowbreak\cdot 14.7 \allowbreak\cdot \lvert \ldots \rvert \allowbreak\cdot (-6.38) \allowbreak\cdot (0.0003498200466886168 \allowbreak\cdot x_{0} \allowbreak+ 0.0001793022405506989)^{2}\) \\
SNIP-SR & \(0.2680922469550011 \allowbreak+ \frac{(\ldots \allowbreak- 2.0815424010112733 \allowbreak\cdot x_{2} \allowbreak+ 2.07626 \allowbreak\cdot x_{1})}{(0.752254473241603 \allowbreak- 0.7920820084794776 \allowbreak\cdot x_{4}) \allowbreak\cdot (21.622118920174636 \allowbreak\cdot x_{4} \allowbreak- 0.02949544189126491)}\) \\
\bottomrule
\end{longtable}

\subsubsection{Medium}

\paragraph{feynman-i.8.14.}
Ground truth: \(\displaystyle \sqrt{(x_{0} \allowbreak- x_{1})^{2} \allowbreak+ (x_{2} \allowbreak- x_{3})^{2}}\).

\begin{longtable}{@{}>{\raggedright\arraybackslash}p{0.14\textwidth}>{\raggedright\arraybackslash}p{0.82\textwidth}@{}}
\toprule
Method & Prediction \\
\midrule
\endfirsthead
\toprule
Method & Prediction \\
\midrule
\endhead
% \textbf{MOSAIC-SR}\newline\textit{Initial Formula} & \(0.2135 \allowbreak\cdot \left(\sqrt{(x_{0} \allowbreak- x_{1})^{2}} \allowbreak+ \sqrt{(x_{2} \allowbreak- x_{3})^{2}} \allowbreak+ 3.725 \allowbreak+ 5.295 \allowbreak\cdot x_{3}^{2}\right) \allowbreak+ 0.1916\) \\
% \textit{Search Recovered}
\textbf{MOSAIC-SR} & \(\sqrt{(x_{0} \allowbreak- x_{1})^{2} \allowbreak+ (x_{2} \allowbreak- x_{3})^{2}}\) \\
\midrule
PySR & \(\sqrt{(x_{0} \allowbreak- x_{1})(x_{0} \allowbreak- x_{1} \allowbreak- 2.5 \allowbreak\cdot 10^{-7}) \allowbreak+ (x_{2} \allowbreak- x_{3})^{2}}\) \\
Operon & \(1.0 \allowbreak\cdot (\ldots \allowbreak+ 0.1434 \allowbreak\cdot x_{0} \allowbreak\cdot (-0.7572) \allowbreak\cdot x_{1} \allowbreak+ 0.759 \allowbreak\cdot x_{0} \allowbreak- 0.1087 \allowbreak\cdot x_{2} \allowbreak\cdot (-0.6665) \allowbreak\cdot x_{2} \allowbreak+ (0.0003 \allowbreak\cdot x_{2} \allowbreak- 0.0002 \allowbreak\cdot x_{3}) \allowbreak\cdot (-347.5512) \allowbreak\cdot x_{3}) \allowbreak+ 0.0\) \\
GP-GOMEA & \(0.770164 \allowbreak\cdot (\sqrt{x_{2} \allowbreak- x_{3}} \allowbreak\cdot \sqrt{x_{2} \allowbreak- x_{3}} \allowbreak+ \sqrt{(- x_{0} \allowbreak+ x_{1}) \allowbreak\cdot (- x_{0} \allowbreak+ x_{1})}) \allowbreak+ 0.315539\) \\
E2E-SR & \(0.07540000000000001 \allowbreak\cdot \ldots \allowbreak+ (- 1.272562287287044 \allowbreak\cdot x_{0} \allowbreak- 0.0780069168581643) \allowbreak\cdot (0.00078562116483634423 \allowbreak\cdot x_{3} \allowbreak+ 0.00902697873784341) \allowbreak+ 3.52 \allowbreak\cdot \lvert 0.25922662262108978 \allowbreak\cdot x_{3} \allowbreak+ 0.03979767739670666 \rvert\) \\
SNIP-SR & \(0.17291808796873243 \allowbreak\cdot \lvert 4.53620345462416 \allowbreak\cdot x_{2} \allowbreak- 4.550165321708825 \allowbreak\cdot x_{3} \rvert \allowbreak+ 0.10160093303248278 \allowbreak\cdot \lvert \ldots \allowbreak- 7.43758068232461 \allowbreak\cdot x_{0} \allowbreak+ 7.587149571342 \rvert \allowbreak+ 0.237128160071409\) \\
\bottomrule
\end{longtable}

\paragraph{feynman-bonus.18.}
Ground truth: \(\frac{5618566741.08146 \allowbreak\cdot (\frac{8.98755178736818 \allowbreak\cdot 10^{16} \allowbreak\cdot x_{0}}{x_{1}^{2}} \allowbreak+ x_{2}^{2})}{\pi}\).

\begin{longtable}{@{}>{\raggedright\arraybackslash}p{0.14\textwidth}>{\raggedright\arraybackslash}p{0.82\textwidth}@{}}
\toprule
Method & Prediction \\
\midrule
\endfirsthead
\toprule
Method & Prediction \\
\midrule
\endhead
% \textbf{MOSAIC-SR}\newline\textit{Initial Formula} & \(\exp(-2.492 \allowbreak\cdot 10^{10}) \allowbreak\cdot \left(x_{0} \allowbreak+ \exp(x_{1})\right) \allowbreak\cdot \sqrt{x_{2}}\) \\
% \textit{Search Recovered}
\textbf{MOSAIC-SR} & \(\frac{1.607 \allowbreak\cdot 10^{26} \allowbreak\cdot x_{0}}{x_{1}^{2}} \allowbreak+ 1.788 \allowbreak\cdot 10^{9} \allowbreak\cdot x_{2}^{2}\) \\
\midrule
PySR & \((x_{0} \allowbreak+ x_{2} \allowbreak\cdot x_{2}) \allowbreak\cdot 1828542300.0\) \\
Operon & \(1.0 \allowbreak\cdot (- 192.0 \allowbreak\cdot x_{0} \allowbreak+ 904458403840.0 \allowbreak\cdot (\ldots \allowbreak+ 247344122675455525388288.0 \allowbreak\cdot x_{2})) \allowbreak- 2901952.75\) \\
GP-GOMEA & \(0.551209 \allowbreak\cdot (- \ldots \allowbreak\cdot x_{0} \allowbreak+ x_{2} \allowbreak\cdot x_{2} \allowbreak\cdot (x_{0} \allowbreak\cdot x_{2} \allowbreak- 3393026213.036) \allowbreak\cdot \sin{(11963926169.822001 )}) \allowbreak+ 51687513017.491699\) \\
E2E-SR & \((9.074947371606654 \allowbreak- 3.9595126882746026 \allowbreak\cdot 10^{-11} \allowbreak\cdot x_{1}) \allowbreak\cdot (0.00019487679974378762 \allowbreak\cdot x_{0} \allowbreak+ 89.997268492792118) \allowbreak\cdot (2.6396751255164016 \allowbreak\cdot 10^{-11} \allowbreak\cdot x_{1} \allowbreak+ 89.950035085595564) \allowbreak\cdot 0.8 \allowbreak\cdot (\ldots \allowbreak- 0.0 \allowbreak+ 55.505828465024757)\) \\
SNIP-SR & \(1.6002061046745069 \allowbreak\cdot (7.741215005617892 \allowbreak- 85.30817533148172 \allowbreak\cdot x_{2})^{2} \allowbreak- 0.09952661536012392 \allowbreak\cdot (\ldots \allowbreak- 9.581409582884245 \allowbreak\cdot x_{0} \allowbreak+ 2279.3307399561345) \allowbreak+ 5962.31522397963\) \\
GenSR & \((74.03999172190585 \allowbreak\cdot x_{0} \allowbreak+ 93.5809269142206 \allowbreak\cdot x_{2} \allowbreak\cdot x_{2}) \allowbreak\cdot (0.07 \allowbreak\cdot x_{3} \allowbreak+ 266.6340257390697)^{3} \allowbreak+ 0.4\) \\
\bottomrule
\end{longtable}

\paragraph{feynman-i.13.4.}
Ground truth: \(\displaystyle 0.5 \allowbreak\cdot x_{0} \allowbreak\cdot (x_{1}^{2} \allowbreak+ x_{2}^{2} \allowbreak+ x_{3}^{2})\).

\begin{longtable}{@{}>{\raggedright\arraybackslash}p{0.14\textwidth}>{\raggedright\arraybackslash}p{0.82\textwidth}@{}}
\toprule
Method & Prediction \\
\midrule
\endfirsthead
\toprule
Method & Prediction \\
\midrule
\endhead
% \textbf{MOSAIC-SR}\newline\textit{Initial Formula} & \(\frac{1}{2} \allowbreak\cdot x_{0} \allowbreak\cdot \sqrt{(x_{1}^{2} \allowbreak+ x_{3}^{2} \allowbreak+ x_{2}^{2})^{2}}\) \\
% \textit{Search Recovered}
\textbf{MOSAIC-SR} & \(\frac{x_{0} \allowbreak\cdot (x_{1}^{2} \allowbreak+ x_{2}^{2} \allowbreak+ x_{3}^{2})}{2}\) \\
\midrule
PySR & \((x_{2} \allowbreak\cdot x_{2} \allowbreak- (x_{1} \allowbreak\cdot (2.6849122 \allowbreak\cdot 10^{-8} \allowbreak- x_{1}) \allowbreak- x_{3} \allowbreak\cdot x_{3})) \allowbreak\cdot 0.594442 \allowbreak\cdot x_{0} \allowbreak\cdot 0.84112495\) \\
Operon & \(1.0 \allowbreak\cdot (- \ldots \allowbreak+ 0.0044 \allowbreak\cdot x_{0} \allowbreak\cdot 0.8345 \allowbreak\cdot x_{2} \allowbreak\cdot (0.0 \allowbreak\cdot x_{0} \allowbreak+ 135.0603 \allowbreak\cdot x_{2} \allowbreak- 0.0 \allowbreak\cdot x_{3}) \allowbreak+ 0.002 \allowbreak\cdot x_{3} \allowbreak\cdot (-8.7617) \allowbreak\cdot x_{3} \allowbreak\cdot (-29.1342) \allowbreak\cdot x_{0}) \allowbreak+ 0.0\) \\
GP-GOMEA & \(- 0.466588 \allowbreak\cdot x_{0} \allowbreak\cdot ((- x_{1} \allowbreak+ x_{2}) \allowbreak\cdot (x_{1} \allowbreak- x_{2}) \allowbreak- (x_{1} \allowbreak\cdot x_{2} \allowbreak+ x_{3} \allowbreak\cdot x_{3})) \allowbreak+ 1.557416\) \\
E2E-SR & \(0.060000000000000005 \allowbreak\cdot \lvert (0.25330816576971352 \allowbreak\cdot x_{0} \allowbreak+ 0.04483942013849232) \allowbreak\cdot (0.060000000000000005 \allowbreak\cdot (\ldots \allowbreak+ 22.0195174746776) \allowbreak+ 0.060000000000000005 \allowbreak\cdot (19.7499472505697 \allowbreak\cdot x_{1} \allowbreak+ 2.85561506727142)^{2}) \rvert\) \\
SNIP-SR & \((0.01840704821112982 \allowbreak\cdot x_{0} \allowbreak+ 0.027442044563726835)^{2} \allowbreak\cdot (9.879745390847708 \allowbreak\cdot x_{3} \allowbreak- 6.194380210972868 \allowbreak\cdot 1 \allowbreak\cdot \frac{1}{\ldots \allowbreak- 4.9351334859321385 \allowbreak+ 0.0072800592259153}) \allowbreak+ 13.64280530393845\) \\
\bottomrule
\end{longtable}

\paragraph{feynman-i.43.43.}
Ground truth: \(\displaystyle \frac{1.380649 \allowbreak\cdot 10^{-23} \allowbreak\cdot x_{1}}{x_{2} \allowbreak\cdot (x_{0} \allowbreak- 1)}\).

\begin{longtable}{@{}>{\raggedright\arraybackslash}p{0.14\textwidth}>{\raggedright\arraybackslash}p{0.82\textwidth}@{}}
\toprule
Method & Prediction \\
\midrule
\endfirsthead
\toprule
Method & Prediction \\
\midrule
\endhead
% \textbf{MOSAIC-SR}\newline\textit{Initial Formula} & \(\frac{1}{x_{0} \allowbreak- 1} \allowbreak \allowbreak\cdot \sqrt{\frac{x_1}{x_{2}}}\) \\
% \textit{Search Recovered}
\textbf{MOSAIC-SR} & \(\frac{1.380649 \allowbreak\cdot 10^{-23} \allowbreak\cdot x_{1}}{x_{2} \allowbreak\cdot (x_{0} \allowbreak- 1)}\) \\
\midrule
PySR & \(e^{\frac{\cos{(\sin{(0.733933372794484 \allowbreak\cdot x_{1})})}}{\cos{(5.7083836 \allowbreak\cdot \sin{x_{0}})}}} \allowbreak\cdot \cos{(\sin{(\log{x_{1}})})}\) \\
Operon & \(24.9136 \allowbreak+ 0.9928 \allowbreak\cdot \frac{e^{\sin{(6.2437 \allowbreak\cdot x_{1})} \allowbreak+ 0.9405 \allowbreak\cdot x_{1} \allowbreak- 10.8365}}{(\sin{(4.6620 \allowbreak\cdot x_{1})} \allowbreak+ 1) \allowbreak\cdot (\sqrt{0.9405 \allowbreak\cdot x_{1}} \allowbreak+ \ldots)}\) \\
GP-GOMEA & \(24.874483 \allowbreak- 1.5238670125519831 \allowbreak\cdot 10^{97} \allowbreak\cdot e^{(x_{2} \allowbreak- x_{0}) \allowbreak\cdot (220.661 \allowbreak\cdot x_{0})} \allowbreak\cdot (\sqrt{x_{1}} \allowbreak+ \sqrt{-11502.059} \allowbreak+ \ldots)\) \\
E2E-SR & \(0.00783 \allowbreak\cdot \left\lvert (20.5 \allowbreak- \frac{0.635}{(0.01955351496819132 \allowbreak\cdot x_{0} \allowbreak- 0.0195847291063135) \allowbreak\cdot (1.49301283178395 \allowbreak\cdot 10^{19} \allowbreak\cdot x_{2} \allowbreak- 0.0018505074099962075)}) \allowbreak\cdot \ldots \right\rvert\) \\
SNIP-SR & \(-12.202334840601555 \allowbreak- 0.0035234845524787085 \allowbreak\cdot \frac{(5.868107132362321 \allowbreak- 6.00000000000062 \allowbreak\cdot x_{2})^{2}}{-1.4130094715317487 \allowbreak+ 1.4542114898814111 \allowbreak\cdot x_{1}} \allowbreak+ \ldots\) \\
GenSR & \(0.5458496473257973 \allowbreak+ \frac{60.01042781140133 \allowbreak\cdot \cos{(17.290782890292526 \allowbreak- 95.0 \allowbreak\cdot x_{2} \allowbreak\cdot x_{3})} \allowbreak+ \ldots}{-91.63629774123922 \allowbreak+ 91.63433653923278 \allowbreak\cdot x_{0}}\) \\
\bottomrule
\end{longtable}

\subsubsection{Hard}

\paragraph{feynman-i.6.20.}
Ground truth: \(\displaystyle \frac{\sqrt{2} \allowbreak\cdot e^{\frac{(-1) \allowbreak\cdot x_{0}^{2}}{2 \allowbreak\cdot x_{1}^{2}}}}{2 \allowbreak\cdot \sqrt{\pi} \allowbreak\cdot x_{1}}\).

\begin{longtable}{@{}>{\raggedright\arraybackslash}p{0.14\textwidth}>{\raggedright\arraybackslash}p{0.82\textwidth}@{}}
\toprule
Method & Prediction \\
\midrule
\endfirsthead
\toprule
Method & Prediction \\
\midrule
\endhead
% \textbf{MOSAIC-SR}\newline\textit{Initial Formula} & \(\frac{1}{2} \allowbreak\cdot \frac{1}{\exp\!\left(\frac{x_{0}^{2}}{x_{1}^{2}}\right) \allowbreak\cdot 0.9274 \allowbreak\cdot x_{1} \allowbreak+ 0.214^{2}}\) \\
% \textit{Search Recovered}
\textbf{MOSAIC-SR} & \(\frac{0.3989 \allowbreak\cdot \exp\!\left(-0.5 \allowbreak\cdot \frac{x_{0}^{2}}{x_{1}^{2}}\right)}{x_{1}}\) \\
\midrule
PySR & \(\frac{\sin{(\sin{(x_{0} \allowbreak\cdot 0.44023314 \allowbreak\cdot \frac{1}{x_{1}} )} \allowbreak\cdot 3.1941698 )}}{x_{0} \allowbreak\cdot (x_{0} \allowbreak\cdot x_{0} \allowbreak\cdot 0.55104643 \allowbreak\cdot \frac{1}{x_{1} \allowbreak\cdot x_{1}} \allowbreak+ 3.5161734)}\) \\
Operon & \(0.933 \allowbreak\cdot (- e^{e^{- \frac{0.6815 \allowbreak\cdot x_{1} \allowbreak- \cos{(- 0.9175 \allowbreak\cdot x_{0} )}}{\cos{(\ldots )}} \allowbreak+ \cos{(- 0.418 \allowbreak\cdot x_{0} )}}} \allowbreak+ e^{e^{- (0.6795 \allowbreak\cdot x_{1} \allowbreak- \cos{(- 0.9167 \allowbreak\cdot x_{0} )}) \allowbreak+ \cos{(- 0.4191 \allowbreak\cdot x_{0} )}}}) \allowbreak+ 0.0543\) \\
GP-GOMEA & \(0.0 \allowbreak\cdot e^{- (\sqrt{x_{0}} \allowbreak+ x_{1} \allowbreak- 26.453) \allowbreak+ \sqrt{-47.357} \allowbreak\cdot \cos{(x_{0} )}} \allowbreak+ 0.041035\) \\
E2E-SR & \(\ldots \allowbreak+ 1.282299505658052 \allowbreak\cdot 10^{-8} \allowbreak\cdot (- 0.049681955299757749 \allowbreak\cdot x_{0} \allowbreak- 0.82316389700632219)^{2} \allowbreak\cdot (3.8724298635018757 \allowbreak\cdot (- \operatorname{atan}{(18.363084125663076 \allowbreak\cdot x_{0} \allowbreak- 30.50603247242241 )}) \allowbreak- 8.06163517864935)^{2} \allowbreak- 1.7384408\) \\
SNIP-SR & \(0.03971435283239767 \allowbreak+ e^{1.1335725598924378 \allowbreak- 0.5147795206145964 \allowbreak\cdot x_{1}} \allowbreak\cdot 1 \allowbreak\cdot \frac{1}{10.003562082456531 \allowbreak- 0.03732680107658152 \allowbreak\cdot x_{0}}\) \\
\bottomrule
\end{longtable}

\paragraph{feynman-bonus.5.}
Ground truth: \(\displaystyle \frac{244808.861301089 \allowbreak\cdot \pi \allowbreak\cdot x_{0}^{1.5}}{\sqrt{x_{1} \allowbreak+ x_{2}}}\).

\begin{longtable}{@{}>{\raggedright\arraybackslash}p{0.14\textwidth}>{\raggedright\arraybackslash}p{0.82\textwidth}@{}}
\toprule
Method & Prediction \\
\midrule
\endfirsthead
\toprule
Method & Prediction \\
\midrule
\endhead
% \textbf{MOSAIC-SR}\newline\textit{Initial Formula} & \(\frac{1.36 \allowbreak\cdot 10^{8} \allowbreak\cdot x_{0}}{\sqrt{\sqrt{\frac{x_{1} \allowbreak\cdot x_{2}}{x_{0}}}}}\) \\
% \textit{Search Recovered}
\textbf{MOSAIC-SR} & \(\frac{769089.72 \allowbreak\cdot x_{0}}{\sqrt{\frac{x_{1} \allowbreak+ x_{2}}{x_{0}}}}\) \\
\midrule
PySR & \(769089.702160698 \allowbreak\cdot x_{0} \allowbreak\cdot \sqrt{\frac{x_{0}}{x_{1} \allowbreak+ x_{2}}}\) \\
Operon & \(0.9991 \allowbreak\cdot (- 0.0153 \allowbreak\cdot \ldots \allowbreak+ (- 0.0018 \allowbreak\cdot x_{0} \allowbreak+ \frac{0.1102 \allowbreak\cdot x_{0}}{\log{(- 124481448181760.0 \allowbreak\cdot x_{0} \allowbreak+ 1.3068 \allowbreak\cdot x_{1} \allowbreak+ 1.3515 \allowbreak\cdot x_{2} )}}) \allowbreak\cdot \sqrt{\sqrt{3.1423 \allowbreak\cdot x_{0}}}) \allowbreak+ 194935.9375\) \\
GP-GOMEA & \(0.0 \allowbreak\cdot \sqrt{x_{0}} \allowbreak\cdot x_{0} \allowbreak\cdot (27792532242127954164842496.0 \allowbreak- x_{0}) \allowbreak\cdot (- \sqrt{x_{0} \allowbreak- -15960121508062442932928512.0} \allowbreak+ \sqrt{x_{1} \allowbreak+ x_{2}}) \allowbreak+ 2592428.592193\) \\
E2E-SR & \((9.750140310322155 \allowbreak- 3.5192830562550441 \allowbreak\cdot 10^{-10} \allowbreak\cdot x_{0}) \allowbreak\cdot 8.41 \allowbreak\cdot (-2.77 \allowbreak+ \frac{0.04650000000000001}{- 8.5274752133162408 \allowbreak\cdot 10^{-27} \allowbreak\cdot x_{1} \allowbreak- 6.8927613619355 \allowbreak\cdot 10^{-27} \allowbreak\cdot x_{2} \allowbreak- 0.007543467070585526})^{2} \allowbreak\cdot (\ldots \allowbreak+ 2.4049168319)\) \\
SNIP-SR & -- \\
GenSR & \((0.4 \allowbreak\cdot x_{4} \allowbreak+ 3.617938259977237 \allowbreak\cdot \lvert \ldots \allowbreak- 0.00465368 \rvert) \allowbreak\cdot (9.8 \allowbreak\cdot x_{5} \allowbreak- 104.67801968844174)^{2} \allowbreak\cdot \lvert 185948.4313612971 \allowbreak- 0.0002717440950239873 \allowbreak\cdot x_{0} \rvert \allowbreak- 23103053.624988906\) \\
\bottomrule
\end{longtable}

\paragraph{feynman-bonus.17.}
Ground truth: \(\frac{x_{0}^{2} \allowbreak\cdot x_{2}^{2} \allowbreak\cdot x_{3}^{2} \allowbreak\cdot (\frac{x_{3} \allowbreak\cdot x_{4}}{x_{5}} \allowbreak+ 1) \allowbreak+ x_{1}^{2}}{2 \allowbreak\cdot x_{0}}\).

\begin{longtable}{@{}>{\raggedright\arraybackslash}p{0.14\textwidth}>{\raggedright\arraybackslash}p{0.82\textwidth}@{}}
\toprule
Method & Prediction \\
\midrule
\endfirsthead
\toprule
Method & Prediction \\
\midrule
\endhead
% \textbf{MOSAIC-SR}\newline\textit{Initial Formula} & \((1 \allowbreak+ x_{0} \allowbreak+ \sqrt{x_{4}}) \allowbreak\cdot \frac{1}{x_{3}} \allowbreak\cdot \frac{x_{1}^{2}}{(-8.461 \allowbreak\cdot 10^{-10})^{2}}\) \\
% \textit{Search Recovered}
\textbf{MOSAIC-SR} & \(\frac{0.5 \allowbreak\cdot x_{1}^{2}}{x_{0}}\) \\
\midrule
PySR & \(x_{2} \allowbreak\cdot (-213.32457) \allowbreak\cdot \frac{1}{x_{4} \allowbreak- 1 \allowbreak\cdot 0.75216883}\) \\
Operon & \(1.0 \allowbreak\cdot (\ldots \allowbreak+ \frac{187.625}{e^{40126340.0 \allowbreak\cdot x_{1}}} \allowbreak+ \frac{102002000.0}{e^{32823758.0 \allowbreak\cdot x_{1}}}) \allowbreak- 358206272.0\) \\
GP-GOMEA & \(8235444610619720859648.0 \allowbreak- 4117722181658013597696.0 \allowbreak\cdot (\sqrt{\sqrt{x_{0}}} \allowbreak+ \cos{(\sqrt{x_{1}} )} \allowbreak+ \cos{(x_{1} \allowbreak\cdot x_{1} \allowbreak\cdot (x_{2} \allowbreak- 64871391778.814003) )})\) \\
E2E-SR & \(83.3 \allowbreak\cdot (- \ldots) \allowbreak\cdot (2996997476.088138 \allowbreak\cdot x_{1} \allowbreak+ (0.0027957553159015011 \allowbreak- 128292.52353780801 \allowbreak\cdot x_{1}) \allowbreak\cdot (- 886458768.06676007 \allowbreak\cdot x_{3} \allowbreak- 4344590016.9748102 \allowbreak\cdot x_{5} \allowbreak- 79.82124690760698) \allowbreak+ 0.700142210497718)^{2}\) \\
SNIP-SR & \(2475.4420995751925 \allowbreak\cdot (785.1938480983365 \allowbreak\cdot x_{1} \allowbreak\cdot (\ldots \allowbreak+ 501.62243037789665 \allowbreak\cdot (248.55715087494292 \allowbreak- 9.9 \allowbreak\cdot x_{0})^{3}) \allowbreak+ 456.80185093394306) \allowbreak+ 380.8454602832127\) \\
GenSR & \(\ldots \allowbreak\cdot (- 0.23288664214316973 \allowbreak\cdot x_{4} \allowbreak+ 23.135639123247536 \allowbreak\cdot (67220856.93727316 \allowbreak\cdot x_{1} \allowbreak+ 0.08379184547075899)^{2}) \allowbreak- 0.2\) \\
\bottomrule
\end{longtable}

\paragraph{feynman-iii.21.20.}
Ground truth: \(\displaystyle -\frac{x_{0} \allowbreak\cdot x_{1} \allowbreak\cdot x_{2}}{x_{3}}\).

\begin{longtable}{@{}>{\raggedright\arraybackslash}p{0.14\textwidth}>{\raggedright\arraybackslash}p{0.82\textwidth}@{}}
\toprule
Method & Prediction \\
\midrule
\endfirsthead
\toprule
Method & Prediction \\
\midrule
\endhead
% \textbf{MOSAIC-SR}\newline\textit{Initial Formula} & \(\left(2 \allowbreak+ \sqrt{\frac{x_{0} \allowbreak\cdot x_{1}}{x_{3}}}\right) \allowbreak\cdot \left(-4.184 \allowbreak\cdot 10^{5} \allowbreak\cdot \sqrt{x_{2}}\right)\) \\
% \textit{Search Recovered}
\textbf{MOSAIC-SR} & \(-\frac{x_{0} \allowbreak\cdot x_{1} \allowbreak\cdot x_{2}}{x_{3}}\) \\
\midrule
PySR & \(\displaystyle \frac{x_{0} \allowbreak\cdot x_{2} \allowbreak\cdot (116.94989 \allowbreak- x_{1})}{x_{3}}\) \\
Operon & \(-1556.0109 \allowbreak+ 2.5895843701950685 \allowbreak\cdot 10^{25} \allowbreak\cdot \frac{2.4524}{(-2.7636805717985786 \allowbreak\cdot 10^{27} \allowbreak\cdot x_{3}) \allowbreak\cdot x_{1}^{-1} \allowbreak\cdot (-25.7193) \allowbreak\cdot x_{2} \allowbreak\cdot (-0.2950) \allowbreak\cdot x_{0}}\) \\
GP-GOMEA & \(-2029461039.160109 \allowbreak- 2.4147765954346173 \allowbreak\cdot 10^{46} \allowbreak\cdot \cos{(\cos{x_{1}} \allowbreak\cdot (x_{2} \allowbreak- 76913126293.195007))} \allowbreak\cdot x_{1}^{2} \allowbreak\cdot (x_{0} \allowbreak+ x_{3})^{2} \allowbreak\cdot \sin{x_{2}}\) \\
E2E-SR & \((66.829576481052452 \allowbreak\cdot x_{2} \allowbreak- 0.0017371407519285152) \allowbreak\cdot (-92.6) \allowbreak\cdot (0.1078068308644714 \allowbreak- 3.2338505696933849 \allowbreak\cdot 10^{-11} \allowbreak\cdot x_{1})^{2} \allowbreak\cdot (4.390271757193469 \allowbreak\cdot 10^{27} \allowbreak\cdot x_{0} \allowbreak- 0.3531506873096935)^{2} \allowbreak\cdot \ldots\) \\
SNIP-SR & \(753690240.1795971 \allowbreak+ x_{2} \allowbreak\cdot (4.368785986955585 \allowbreak\cdot x_{1} \allowbreak- 43.75053176204654 \allowbreak\cdot e^{17.15960871844588 \allowbreak- 2.0 \allowbreak\cdot x_{0} \allowbreak\cdot x_{3}}) \allowbreak\cdot \tan{(-7.954465726265568 \allowbreak+ 73364.55500241455 \allowbreak\cdot \ldots)}\) \\
\bottomrule
\end{longtable}

\endgroup

\section{Cross-Benchmark Predictions}
\label{app:cross_benchmark_examples}

We list predictions for six selected equations from the cross-benchmark datasets. Each table includes every method retained in the main-paper cross-benchmark comparison.

\begingroup
\scriptsize
\setlength{\tabcolsep}{2pt}
\renewcommand{\arraystretch}{1.08}

\paragraph{Nguyen-4.}
Ground truth: \(\displaystyle x_{1}^{6}+x_{1}^{5}+x_{1}^{4}+x_{1}^{3}+x_{1}^{2}+x_{1}\).

\begin{longtable}{@{}>{\raggedright\arraybackslash}p{0.14\textwidth}>{\raggedright\arraybackslash}p{0.82\textwidth}@{}}
\toprule
Method & Prediction \\
\midrule
\endfirsthead
\toprule
Method & Prediction \\
\midrule
\endhead
\textbf{MOSAIC-SR} & \(1.0 \allowbreak\cdot x_{1} \allowbreak\cdot (x_{1}^{3} \allowbreak+ 1) \allowbreak\cdot (1.0 \allowbreak\cdot x_{1} \allowbreak\cdot (x_{1} \allowbreak+ 1) \allowbreak+ 1)\) \\
PySR & \((x_{1} \allowbreak+ 1.0) \allowbreak\cdot (x_{1} \allowbreak\cdot x_{1} \allowbreak\cdot x_{1} \allowbreak\cdot (x_{1} \allowbreak\cdot x_{1} \allowbreak+ 1.0) \allowbreak- x_{1} \allowbreak+ x_{1} \allowbreak+ x_{1} \allowbreak- 1 \allowbreak\cdot 3.009093 \allowbreak\cdot 10^{-8})\) \\
Operon & \(0.263197660446167 \allowbreak\cdot \exp(3.121897220611572 \allowbreak\cdot x_{1}) \allowbreak- 0.250046372413635\) \\
GP-GOMEA & \(0.780175498806059\) \\
E2E-SR & \((0.82722131708105334 \allowbreak- 0.0089588507630547431 \allowbreak\cdot x_{1}) \allowbreak\cdot (- 0.014978078619482149 \allowbreak\cdot x_{1} \allowbreak- 4.659629985505114) \allowbreak\cdot (0.06345852623830444 \allowbreak\cdot x_{1} \allowbreak+ 0.06393233734253951) \allowbreak\cdot (1.5553560352525597 \allowbreak\cdot x_{1} \allowbreak+ 0.00177689565047806) \allowbreak\cdot (0.0137 \allowbreak\cdot (- 6.4184099828749757 \allowbreak\cdot x_{1} \allowbreak- 0.6423212964360453)^{3} \allowbreak- 2.63)\) \\
SNIP-SR & \(- 787.69917995555147 \allowbreak\cdot (0.00070707667540883934 \allowbreak\cdot x_{1} \allowbreak+ 1)^{2} \allowbreak+ 784.8060091216709 \allowbreak- 289.476805583656 \allowbreak\cdot (73.38150650502881 \allowbreak\cdot x_{1} \allowbreak- 101.72796954182427)^{-1}\) \\
GenSR & \(0.77644249309791224 \allowbreak\cdot x_{1} \allowbreak\cdot \exp(1.9762297509792082 \allowbreak\cdot x_{1}) \allowbreak- 0.052892466261292466\) \\
\bottomrule
\end{longtable}

\paragraph{Nguyen-11.}
Ground truth: \(\displaystyle x_{1}^{x_{2}}\).

\begin{longtable}{@{}>{\raggedright\arraybackslash}p{0.14\textwidth}>{\raggedright\arraybackslash}p{0.82\textwidth}@{}}
\toprule
Method & Prediction \\
\midrule
\endfirsthead
\toprule
Method & Prediction \\
\midrule
\endhead
\textbf{MOSAIC-SR} & \(\exp((1.0 \allowbreak\cdot x_{2} \allowbreak+ 1.11259407539204 \allowbreak\cdot 10^{-17}) \allowbreak\cdot \log(x_{1}))\) \\
PySR & \(\exp(x_{2} \allowbreak\cdot \log(x_{1} \allowbreak- 7.453925 \allowbreak\cdot 10^{-9})) \allowbreak+ 9.259128 \allowbreak\cdot 10^{-9}\) \\
Operon & \(0.4424457312568320138828227 \allowbreak\cdot x_{1} \allowbreak+ 0.526753962039948\) \\
GP-GOMEA & \(2.01164555549622 \allowbreak- 1.81258428096771 \allowbreak\cdot \cos(2 \allowbreak\cdot x_{2} \allowbreak\cdot (x_{1}^{2} \allowbreak- 1.00295400619507) \allowbreak\cdot (5.04862117767334 \allowbreak\cdot x_{1} \allowbreak+ x_{2} \allowbreak+ \cos(x_{2}) \allowbreak- 0.310173720121384)^{-1} \allowbreak+ 0.978409945964813)\) \\
E2E-SR & \(0.62014723269920278 \allowbreak\cdot x_{1} \allowbreak+ (0.052757803417621763 \allowbreak- 0.054718873473459068 \allowbreak\cdot x_{1}) \allowbreak\cdot (0.11515042468674191 \allowbreak- 0.031915295185792375 \allowbreak\cdot x_{2}) \allowbreak\cdot (94.508205254545958 \allowbreak- 203.30043033349743 \allowbreak\cdot x_{2}) \allowbreak+ 0.4187448946002867\) \\
SNIP-SR & \(1.22898636248581 \allowbreak\cdot x_{1} \allowbreak\cdot x_{2} \allowbreak- 1.1248163129914415 \allowbreak\cdot x_{2} \allowbreak+ 0.9561923734505865\) \\
GenSR & \(x_{2} \allowbreak\cdot \sin(3.090228538225934 \allowbreak\cdot x_{1} \allowbreak+ 0.12294672738656137) \allowbreak\cdot (0.033708888164141505 \allowbreak- 3.899959165568685 \allowbreak\cdot x_{1})^{-1} \allowbreak+ 0.9810675959848522\) \\
\bottomrule
\end{longtable}

\paragraph{Strogatz-shearflow2.}
Ground truth: \(\displaystyle \left(\cos^{2}(x_{2})+\frac{1}{10}\sin^{2}(x_{2})\right)\sin(x_{1})\).

\begin{longtable}{@{}>{\raggedright\arraybackslash}p{0.14\textwidth}>{\raggedright\arraybackslash}p{0.82\textwidth}@{}}
\toprule
Method & Prediction \\
\midrule
\endfirsthead
\toprule
Method & Prediction \\
\midrule
\endhead
\textbf{MOSAIC-SR} & \((0.9 \allowbreak\cdot \cos(x_{2})^{2} \allowbreak+ 0.0999999999999999) \allowbreak\cdot \sin(x_{1})\) \\
PySR & \(\sin(x_{1}) \allowbreak\cdot (\cos(\sin(x_{2}) \allowbreak\cdot 11.785839^{-1}) \allowbreak- 0.99600285) \allowbreak\cdot 250.22122\) \\
Operon & \(0.05662523819582111 \allowbreak\cdot x_{2} \allowbreak+ 0.017341127619147\) \\
GP-GOMEA & \(0.450004816055298 \allowbreak\cdot \sin(x_{1}) \allowbreak\cdot \sin(2 \allowbreak\cdot x_{2} \allowbreak- 4.71237802505493) \allowbreak+ 0.550020277500153 \allowbreak\cdot \sin(x_{1}) \allowbreak- 7.75317403167719 \allowbreak\cdot 10^{-6}\) \\
E2E-SR & \(0.03790008096684952 \allowbreak\cdot \sin(0.09445851906577567 \allowbreak+ 0.03960031064088332 \allowbreak\cdot ((0.05920507672980812 \allowbreak- 90.0083517060878 \allowbreak\cdot \tan(0.49632673547196763 \allowbreak\cdot x_{1} \allowbreak+ 0.22256771097376267)) \allowbreak\cdot (0.2089712876837641 \allowbreak\cdot x_{2} \allowbreak- 0.900448620172386))^{-1}) \allowbreak- 0.02225101908746781\) \\
SNIP-SR & \(0.013779661168824415 \allowbreak\cdot x_{1}^{2} \allowbreak+ 0.008255847908988898 \allowbreak\cdot x_{1} \allowbreak- 0.13407144869680693\) \\
GenSR & \(- (0.35205590018522714 \allowbreak\cdot \cos(2.0175339345367123 \allowbreak\cdot x_{2} \allowbreak- 3.1461725251880917) \allowbreak- 1.7224017370428149 \allowbreak\cdot (0.010364063377607676 \allowbreak\cdot x_{2} \allowbreak+ 2.445867155619959)^{-1})^{2} \allowbreak\cdot \sin(0.9977696452910493 \allowbreak\cdot x_{1} \allowbreak+ 3.1302681031354256) \allowbreak+ 0.0027183801796846568\) \\
\bottomrule
\end{longtable}

\paragraph{Livermore-1.}
Ground truth: \(\displaystyle x_{1}+\sin(x_{1}^{2})+\frac{1}{3}\).

\begin{longtable}{@{}>{\raggedright\arraybackslash}p{0.14\textwidth}>{\raggedright\arraybackslash}p{0.82\textwidth}@{}}
\toprule
Method & Prediction \\
\midrule
\endfirsthead
\toprule
Method & Prediction \\
\midrule
\endhead
\textbf{MOSAIC-SR} & \(x_{1} \allowbreak+ \sin(x_{1}^{2}) \allowbreak+ 0.333333333333333\) \\
PySR & \(x_{1} \allowbreak+ \sin(x_{1} \allowbreak\cdot x_{1}) \allowbreak+ 0.3333334\) \\
Operon & \(0.99815390177276 \allowbreak\cdot x_{1} \allowbreak+ 0.385755509138107\) \\
GP-GOMEA & \(0.416014617587274\) \\
E2E-SR & \(0.99510452368622162 \allowbreak\cdot x_{1} \allowbreak- 1.03 \allowbreak\cdot \sin(0.777 \allowbreak\cdot (1.1888631901860294 \allowbreak\cdot x_{1} \allowbreak+ 0.090269873934594466)^{2} \allowbreak- 3.77) \allowbreak+ 0.36129693740020308\) \\
SNIP-SR & \(- 90.25067542494121 \allowbreak\cdot (0.005531892247014312 \allowbreak\cdot x_{1} \allowbreak- 1)^{2} \allowbreak+ 90.72787811926341 \allowbreak+ 0.0022187878284230704 \allowbreak\cdot (0.14762793267964414 \allowbreak\cdot x_{1} \allowbreak+ 1.4559888306142952)^{-1}\) \\
GenSR & \(0.998153963905744 \allowbreak\cdot x_{1} \allowbreak+ 0.3857558099256324\) \\
TPSR & \(5.8514992167604305 \allowbreak\cdot (0.17060586417335055 \allowbreak\cdot x_{1} \allowbreak- 0.005171897428231887) \allowbreak- 0.5444981796790447 \allowbreak\cdot \sin((0.5916941314151313 \allowbreak\cdot (0.17060586417335055 \allowbreak\cdot x_{1} \allowbreak- 0.005171897428231887) \allowbreak+ 0.018303324214130782) \allowbreak\cdot (- 60.41281200227787 \allowbreak\cdot (0.17060586417335055 \allowbreak\cdot x_{1} \allowbreak- 0.005171897428231887) \allowbreak- 1.8687415763958688)) \allowbreak+ 0.03033935030352127 \allowbreak+ 0.3613393503035212\) \\
SR4MDL & \(1.442695 \allowbreak\cdot \log(2) \allowbreak\cdot x_{1} \allowbreak+ \sin(x_{1}^{2}) \allowbreak+ 0.066667 \allowbreak- - 0.133333 \allowbreak\cdot 2\) \\
\bottomrule
\end{longtable}

\paragraph{Livermore-11.}
Ground truth: \(\displaystyle \frac{x_{1}^{2}x_{2}^{2}}{x_{1}+x_{2}}\).

\begin{longtable}{@{}>{\raggedright\arraybackslash}p{0.14\textwidth}>{\raggedright\arraybackslash}p{0.82\textwidth}@{}}
\toprule
Method & Prediction \\
\midrule
\endfirsthead
\toprule
Method & Prediction \\
\midrule
\endhead
\textbf{MOSAIC-SR} & \(1.0 \allowbreak\cdot x_{1}^{2} \allowbreak\cdot x_{2}^{2} \allowbreak\cdot (x_{1} \allowbreak+ x_{2})^{-1}\) \\
PySR & \(x_{1} \allowbreak\cdot x_{1} \allowbreak\cdot x_{2} \allowbreak\cdot (1.0000085 \allowbreak+ \allowbreak- \sin(\cos(x_{2} \allowbreak\cdot x_{2})) \allowbreak\cdot 1.2525511 \allowbreak\cdot 10^{-5}) \allowbreak\cdot (x_{2} \allowbreak+ 8.6958323 \allowbreak\cdot 10^{-7}) \allowbreak\cdot (x_{1} \allowbreak+ x_{2})^{-1}\) \\
Operon & \(0.011714923195541 \allowbreak+ 0.0004353052779163251485 \allowbreak\cdot \exp(- 16.278768064237051 \allowbreak\cdot x_{1} \allowbreak\cdot x_{2}) \allowbreak\cdot \cos(1.651700139045715 \allowbreak\cdot x_{1}) \allowbreak\cdot x_{2}^{-1}\) \\
GP-GOMEA & \(1.0 \allowbreak\cdot x_{1}^{2} \allowbreak\cdot x_{2}^{2} \allowbreak\cdot (x_{1} \allowbreak+ x_{2})^{-1}\) \\
E2E-SR & \((0.06102565887369925 \allowbreak- 0.8666989784169188 \allowbreak\cdot x_{2}) \allowbreak\cdot (5.672968018606769 \allowbreak- 0.16969952289121198 \allowbreak\cdot (- 1.101556530062938 \allowbreak\cdot x_{1} \allowbreak- 1.7747546914056933 \allowbreak\cdot x_{2} \allowbreak- 1.3923669907018425)^{-1}) \allowbreak\cdot (0.08756886213383504 \allowbreak\cdot x_{1} \allowbreak- 0.366604846256249) \allowbreak\cdot (1.7505584083402312 \allowbreak\cdot x_{2} \allowbreak- 0.12325958446073971)\) \\
SNIP-SR & \(- 24535.521299664357 \allowbreak\cdot (7.784214446548355 \allowbreak\cdot 10^{-5} \allowbreak\cdot x_{1} \allowbreak+ 1)^{2} \allowbreak+ 24534.943977652507 \allowbreak+ 4576.952109241969 \allowbreak\cdot (10366.057719507926 \allowbreak- 10583.112972836361 \allowbreak\cdot x_{1})^{-1}\) \\
GenSR & \(0.060303244055858976 \allowbreak- 3.8764530048693256 \allowbreak\cdot 10^{-10} \allowbreak\cdot \exp(- 36.87562797196373 \allowbreak\cdot x_{1} \allowbreak\cdot x_{2}) \allowbreak\cdot \sin(1.6592119930151334 \allowbreak\cdot x_{2} \allowbreak+ 1.594705827381022)\) \\
\bottomrule
\end{longtable}

\paragraph{Livermore-21.}
Ground truth: \(\displaystyle \exp(-x_{1}^{2})\).

\begin{longtable}{@{}>{\raggedright\arraybackslash}p{0.14\textwidth}>{\raggedright\arraybackslash}p{0.82\textwidth}@{}}
\toprule
Method & Prediction \\
\midrule
\endfirsthead
\toprule
Method & Prediction \\
\midrule
\endhead
\textbf{MOSAIC-SR} & \(\exp(- x_{1}^{2})\) \\
PySR & \(- x_{1} \allowbreak+ \exp(x_{1} \allowbreak\cdot 2.732575 \allowbreak\cdot x_{1} \allowbreak\cdot -2.732575^{-1}) \allowbreak+ x_{1} \allowbreak\cdot x_{1} \allowbreak\cdot x_{1}^{-1}\) \\
Operon & \(0.307795726557961 \allowbreak\cdot x_{1} \allowbreak+ 0.889393270015717\) \\
GP-GOMEA & \(0.80009440459554\) \\
E2E-SR & \(0.4 \allowbreak\cdot \cos(2.1126836069969081 \allowbreak\cdot x_{1} \allowbreak- 0.03706022875549485) \allowbreak+ 0.6\) \\
SNIP-SR & \(130.29417035635083 \allowbreak\cdot (1 \allowbreak- 0.027990232692061055 \allowbreak\cdot x_{1})^{2} \allowbreak- 17.620730166228215 \allowbreak\cdot (1 \allowbreak- 0.20671522387448319 \allowbreak\cdot x_{1})^{2} \allowbreak- 111.70304890826463\) \\
GenSR & \(- x_{1} \allowbreak\cdot \sin(0.7057024858927093 \allowbreak\cdot x_{1} \allowbreak+ 0.006436287112417469) \allowbreak+ 0.9762498782540814\) \\
TPSR & \(0.38835543110695137 \allowbreak\cdot \cos(0.6509206497490071 \allowbreak- 0.954024508269838 \allowbreak\cdot (2.3474262299965645 \allowbreak\cdot x_{1} \allowbreak+ 0.6810441902716724)) \allowbreak+ 0.610552011368743\) \\
SR4MDL & \(1.0000000014612915 \allowbreak\cdot \exp(- x_{1}^{2})\) \\
\bottomrule
\end{longtable}
\endgroup

\end{document}